\documentclass{article}
\usepackage{iclr2027_conference,times}

\usepackage{amsmath,amsfonts,bm}

\newcommand{\figleft}{{\em (Left)}}

\newcommand{\figright}{{\em (Right)}}

\def\eqref#1{equation~\ref{#1}}

\def\1{\bm{1}}

\DeclareMathAlphabet{\mathsfit}{\encodingdefault}{\sfdefault}{m}{sl}
\SetMathAlphabet{\mathsfit}{bold}{\encodingdefault}{\sfdefault}{bx}{n}

\usepackage{algorithm}
\usepackage{algpseudocode}
\usepackage{hyperref}
\usepackage{url}
\usepackage{booktabs}
\usepackage{multirow}
\usepackage{graphicx}
\usepackage{capt-of}
\usepackage{tikz}
\usepackage{pgfplots}
\usepackage{wrapfig}
\usepackage{colortbl}
\usepackage{arydshln}
\usepackage{pifont}
\usepackage{needspace}
\newcommand{\cmark}{\ding{51}}
\definecolor{unifiedbg}{RGB}{222,235,250}
\definecolor{unitebg}{RGB}{226,243,222}
\newcommand{\dashedrule}{\noalign{\vskip\aboverulesep}\hdashline\noalign{\vskip\belowrulesep}}
\newcommand{\nocol}{\cellcolor{white}}
\usetikzlibrary{arrows.meta,calc,fit,positioning,shapes.geometric}
\pgfplotsset{compat=1.18}

\title{Single-Transformer Latent Diffusion}

\author{Hao-Yu Hsu, Shenlong Wang \& Alexander G.~Schwing \\
{\normalfont University of Illinois Urbana-Champaign} \\
{\normalfont\texttt{\{haoyuyh3, shenlong, aschwing\}@illinois.edu}} \\
\href{https://haoyuhsu.github.io/Unified-LDM/}{\textcolor{magenta}{\texttt{https://haoyuhsu.github.io/Unified-LDM/}}}
}

\iclrfinalcopy
\begin{document}

\maketitle
\lhead{Preprint}

\begin{abstract}
Latent generative models typically rely on three separate networks: an encoder that tokenizes pixels, a generator that models the latent distribution, and a decoder that maps latents back to pixels. 
UNITE recently merged the first two into a single generative encoder trained in one stage, but its decoder remains a separate network. We ask whether this last network can be folded in too, so that one diffusion transformer encodes, generates, and decodes. Simply letting the shared backbone decode does not help. What makes it work is to cast all three modes as position-aligned sequence maps over one fully shared backbone, distinguished only by a mode token and the timestep. The resulting model, UNIFIED, is simple: one network trained from scratch in a single stage with only reconstruction and flow-matching losses, without adversarial losses or pretrained encoders (e.g., DINO). On ImageNet-1K $256\times256$ at matched parameters, UNIFIED improves gFID over UNITE at three sizes (${\sim}60$M, ${\sim}115$M, and ${\sim}215$M) and reaches $2.08$ with $215$M parameters, the lowest among compared models under $250$M parameters. On text-to-image generation, UNIFIED-B ($218$M) reaches a GenEval score of $0.87$ at $512\times512$, exceeding single-stage pixel diffusion models with around one billion parameters.
Full unification of latent diffusion is thus not merely feasible but beneficial.
\end{abstract}

\section{Introduction}
\suppressfloats[t]
\begin{figure}[t]
    \centering
    \includegraphics[height=0.43\linewidth]{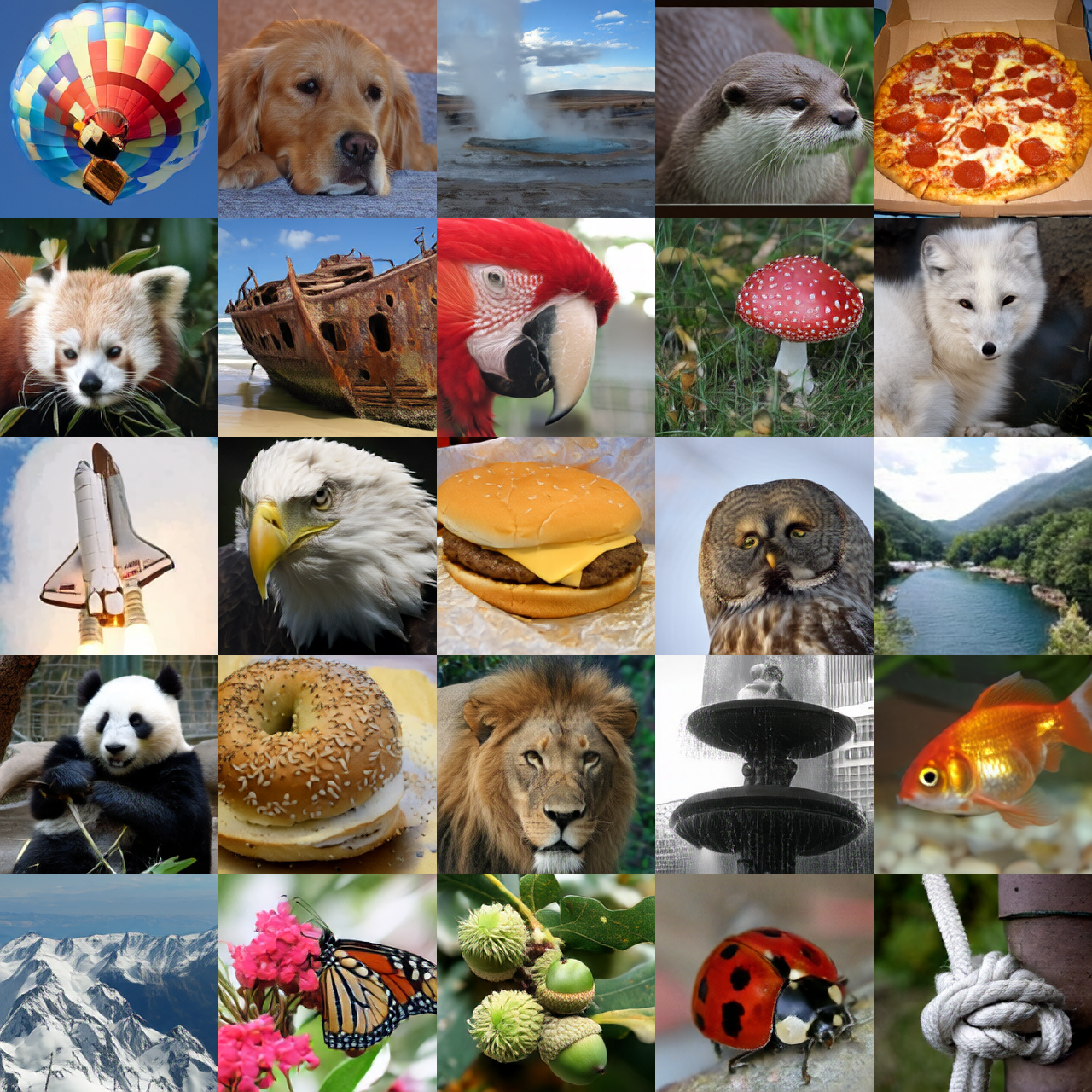}\hfill
    \includegraphics[height=0.43\linewidth]{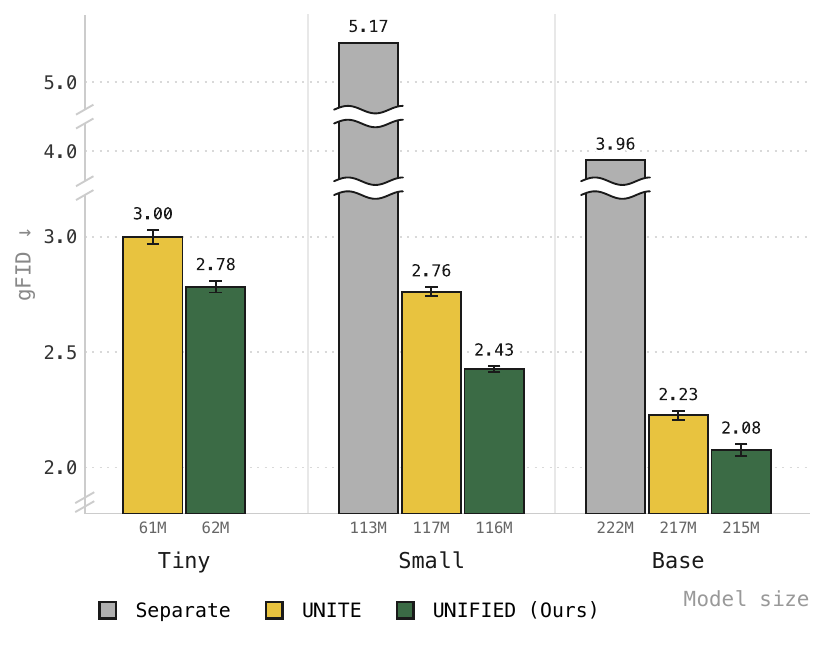}
    \caption{\figleft{} \textbf{Selected samples from UNIFIED} on ImageNet-1K $256\times256$ (CFG scale $4.0$). \figright{} \textbf{Full unification is not merely feasible but beneficial.} gFID ($\downarrow$) at three sizes; bar labels give the total trainable parameters. UNIFIED improves on UNITE at every size, and Tiny UNIFIED nearly matches Small UNITE with about half the parameters ($2.78$ at $62$M vs.\ $2.76$ at $117$M). Separate is a standard three-network pipeline (Figure~\ref{fig:architectures}a), trained at Small and Base. UNITE and UNIFIED: mean over five evaluation seeds (Table~\ref{tab:seeds}).}
    \label{fig:pareto}
\end{figure}

Latent diffusion models~\citep{rombach2022high,peebles2023scalable,ma2024sit}
rely on three trainable networks: an encoder that maps pixels to latents, a
generator that models the latent distribution, and a decoder that maps latents
back to pixels. They are usually trained in stages: the autoencoder is first
trained for reconstruction, and then frozen while the generator is fit to
its latents. This recipe fixes the representation before the model that must
sample from it is trained. It also forces a choice of how much capacity each
network receives, and the effect of that choice on generation is hard to
predict~\citep{hansen2025learnings}. 

We ask whether this pipeline can collapse into a single diffusion transformer
(DiT) whose every block is shared by encoding, denoising, and decoding, trained
in a single stage with only a reconstruction loss and a flow-matching loss, with no
pretrained encoders such as
DINO~\citep{oquab2023dinov2}. Most single-stage training methods keep the
tokenizer and generator as separate
networks~\citep{vahdat2021score,leng2025repa,zhao2026simflow,zheng2026genfirst,heek2026unified}, while
the few that use a single network still reserve parts of it for individual
modes: LDM-is-AE~\citep{zhang2026ldmisae} splits one DiT by depth into encoding
and decoding blocks, which turns the allocation choice above into a choice of
split depth, and DSD~\citep{wang2025diffusion} adds a dedicated diffusion head
and relies on DINO alignment.

\begin{figure}[t]
    \centering
    \includegraphics[width=\linewidth]{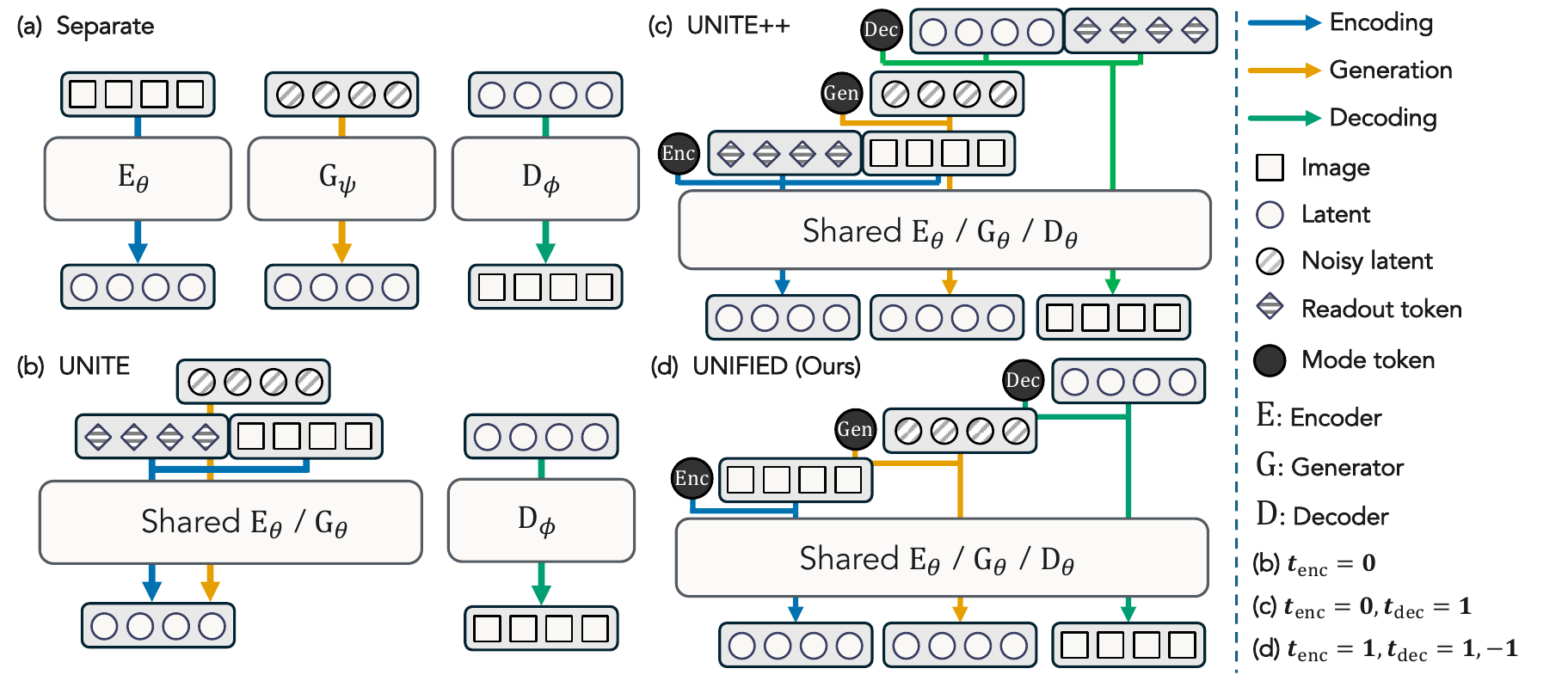}
    \caption{\textbf{From three networks to one shared transformer.}
    \textbf{(a)}~Separate: the standard latent-diffusion pipeline trains an encoder
    $E_\theta$, a generator $G_\psi$, and a decoder $D_\phi$.
    \textbf{(b)}~UNITE shares one backbone between encoding and generation; the
    encoding pass appends $\mathcal{N}(0,I)$ readout tokens to the image patches
    and takes the latent $z$ from their outputs, but a standalone decoder $D_\phi$
    remains.
    \textbf{(c)}~UNITE++ folds decoding into the same backbone as a third mode,
    selected by a mode token, but keeps readout tokens for encoding (latent
    readouts) and decoding (pixel readouts).
    \textbf{(d)}~UNIFIED (ours) removes all readout tokens: every pass maps a
    same-length, position-aligned sequence (image $x$, noisy latent $z_t$, or
    latent $z$) after a mode token at position~$0$, and encodes clean images at
    $t_{\mathrm{enc}}=1$ rather than $0$. Bottom right: each model's
    encoding and decoding timesteps.}
    \label{fig:architectures}
\end{figure}

UNITE~\citep{duggal2026end} is the closest step in this direction. It treats
tokenization and generation as the same latent-inference problem and solves both
with one \emph{generative encoder}, trained in a single stage with only
reconstruction and flow-matching losses, but it still decodes pixels with a separate network
(Figure~\ref{fig:architectures}b). A natural baseline is to let UNITE's backbone
decode as well. We call this extension \textbf{UNITE++}
(Figure~\ref{fig:architectures}c): it shares every weight across the three
modes and selects the mode with a learned mode token. Like UNITE, it reads its outputs
from \emph{readout tokens}, which are extra input tokens whose outputs are taken as the
latent or as pixels. UNITE++ does not improve upon UNITE: at ${\sim}215$M
parameters it reaches a gFID of $2.24$ vs.\ UNITE's $2.23$ and reconstructs markedly
worse (rFID $2.20$ vs.\ $1.41$). We attribute this to the readout design: there
is no direct mapping between latent tokens and image locations, so to decode,
every image location must gather its content from the latent tokens through
attention.

We therefore propose \textbf{UNIFIED} (Figure~\ref{fig:architectures}d), a single
DiT that maps images to latents, latents to latents, and latents to images as
\emph{position-aligned} sequence maps: a given sequence index refers to the same
image location whether it holds a pixel patch, a latent, or a noisy latent. The
backbone is fully shared, and each mode is selected by a learned mode token and
lightweight input and output heads. Relative to UNITE, UNIFIED changes only the
sequence layout, the timesteps for encoding and decoding, and the mode tokens,
and replaces the standalone decoder with a lightweight pixel output head.
The training recipe is unchanged.

The remaining design choice is the timestep that conditions encoding and
decoding. Both are noiseless maps, so the natural choice is the clean endpoint ($t=1$). This improves gFID over UNITE ($2.15$ vs.\ $2.23$ at
${\sim}215$M). At $t=1$, however, three modes disagree on what to output:
encoding and generation both produce the clean latent, whereas decoding produces pixels. 
Sharing $t=1$ thus forces the last
block to serve two output spaces, and the conflict is sharpest near the clean endpoint at the end of
the layers (Section~\ref{sec:representation-analysis}). 
We therefore give decoding its own learned timestep embedding, denoted
$t_{\mathrm{dec}}=-1$ (a label outside $[0,1]$), which leaves $t=1$ to the two
modes whose target is the clean latent. This further lowers gFID to $2.08$ and
restores the alignment between the two passes near the clean endpoint
(Figure~\ref{fig:cka}).

On ImageNet-1K $256\!\times\!256$ at matched parameter counts, UNIFIED improves
gFID over UNITE at all three sizes (${\sim}60$M, ${\sim}115$M, and ${\sim}215$M). Tiny UNIFIED nearly matches Small UNITE on gFID with about half the parameters, suggesting that a
standalone decoder uses capacity less efficiently than the shared backbone.
Against published models, UNIFIED-B reaches a gFID of $2.08$ with $215$M
parameters, the lowest among compared models with at most $250$M parameters,
without a pretrained autoencoder or vision foundation model
(Section~\ref{sec:sota}). On text-to-image generation at $256\times256$,
UNIFIED-L ($554$M) outperforms both the pixel-space JiT~\citep{li2026back}
($615$M) and DiT$^{\mathrm{DH}}$-XL~\citep{zheng2026diffusion} with a pretrained FLUX-VAE ($959$M) on
GenEval, DPG-Bench, and GenAI-Bench. At $512\times512$, UNIFIED-B ($218$M)
reaches $0.87$ on GenEval and $0.827$ on DPG-Bench, exceeding single-stage
pixel diffusion models with around one billion parameters
(Section~\ref{sec:t2i}).

\Needspace*{5\baselineskip}
Our contributions are:

\begin{itemize}
    \item We propose UNIFIED, a single DiT that encodes, generates, and decodes with
    every backbone weight shared, trained from scratch in a single stage with
    only reconstruction and flow-matching losses, without adversarial losses or
    pretrained encoders (e.g., DINO).
    \item We analyze how shared weights keep the three modes apart. We show that
    position alignment restores the reconstruction lost by naive full
    sharing, that the encoding timestep must match what the encoder reads, and
    that a learned decoding timestep removes the conflict between decoding and
    generation at the clean endpoint, leading to better performance.
    \item At matched parameter counts on ImageNet, UNIFIED improves over UNITE at
    three parameter scales and attains the lowest gFID among compared models
    with at most $250$M parameters. It also extends to text-to-image
    generation: at $512\times512$,
    UNIFIED-B ($218$M) exceeds single-stage pixel diffusion models with around one
    billion parameters.
\end{itemize}

\section{Related Work}
\label{sec:related}

\noindent\textbf{Autoencoders and latent diffusion.}
Latent generative models typically follow a two-stage recipe: an autoencoder is trained for reconstruction and frozen, and a separate generator is then fit to its latents. The tokenizer is usually a KL-regularized VAE~\citep{kingma2013auto,rezende2014stochastic} for latent diffusion~\citep{rombach2022high,podell2024sdxl} or a vector-quantized autoencoder~\citep{van2017neural,razavi2019generating} for discrete generation, in both cases trained with the perceptual~\citep{zhang2018unreasonable} and adversarial losses popularized by VQGAN~\citep{esser2021taming}. Because the two stages are decoupled, progress has largely come one network at a time: generators have moved to transformer backbones~\citep{peebles2023scalable,zheng2023fast,gao2023masked,ma2024sit,wang2025ddt} and flow-based objectives~\citep{lipman2022flow,liu2022flow}, while tokenizers have improved latent quantization~\citep{mentzer2024finite,yu2024language} and compactness~\citep{yu2024image,chen2025deep}. This decoupling has a cost, however. The two stages optimize different objectives, which creates an optimization dilemma: a latent that favors reconstruction can degrade generation. VA-VAE~\citep{yao2025reconstruction} addresses this dilemma by aligning the VAE latent with a pretrained vision encoder~\citep{caron2021emerging,oquab2023dinov2}. Such encoders also serve as an alignment target for the generator~\citep{yu2024representation}, or even replace the tokenizer's encoder in the latent diffusion pipeline~\citep{zheng2026diffusion,shi2026latent}. We instead revisit the recipe itself: a single backbone serves as encoder, decoder, and generator, and is trained in a single stage without relying on a pretrained vision encoder.

\noindent\textbf{End-to-end tokenization and generation.}
Closer to our setting, several methods train the tokenizer and generator jointly~\citep{leng2025repa,duggal2026end}. The central difficulty is latent collapse: once the generation loss can shape the latent, the generator can lower its loss by making the latent trivial rather than modeling it~\citep{leng2025repa,wang2025diffusion,zheng2026genfirst}. Existing methods differ mainly in how they prevent this. Likelihood-based approaches keep a variational bound whose KL or entropy terms hold the latent in place, and train a score-based~\citep{vahdat2021score}, normalizing-flow~\citep{zhao2026simflow}, or diffusion~\citep{zheng2026genfirst,heek2026unified} prior jointly with the autoencoder. REPA-E~\citep{leng2025repa} instead anchors the latent to an external representation, backpropagating a representation-alignment loss, rather than the diffusion loss, into the VAE. Other approaches stop the gradient through the denoising target and predict clean latents~\citep{duggal2026end,wang2025diffusion,zhang2026ldmisae}; this needs neither a variational bound nor an external encoder, and we adopt it. Closest to our work, three methods built on this recipe also share weights. UNITE~\citep{duggal2026end} merges the encoder and generator but keeps a standalone decoder; LDM-is-AE~\citep{zhang2026ldmisae} splits one DiT by depth into encoding and decoding blocks, which requires manually choosing the split depth and the feature-mixing schedule; and DSD~\citep{wang2025diffusion} uses a single ViT for all three modes with lightweight blocks for diffusion, but reintroduces an external alignment loss. UNIFIED instead shares every block of a single DiT backbone across all three roles, with no role-specific blocks or alignment loss.

\noindent\textbf{Pixel diffusion.}
A more direct way to avoid the two-stage recipe is to drop the latent and run diffusion on pixels. Early pixel-space models handle the high dimensionality with U-Net designs, adjusted noise schedules, and loss weightings~\citep{dhariwal2021diffusion,hoogeboom2023simple,kingma2023understanding,hoogeboom2025simpler}, adaptive computation over a small set of latent tokens~\citep{jabri2022scalable}, or cascades across resolutions~\citep{chen2025pixelflow}. JiT~\citep{li2026back,starodubcev2026registers} shows that a plain ViT on large patches suffices when it predicts the clean image. Follow-up works improve details with dedicated pixel-level modules~\citep{yu2025pixeldit,wang2025pixnerd,ma2026deco} or U-shaped skip connections~\citep{guo2026pixelu}, or draws on a pretrained vision encoder for representation alignment~\citep{shin2026representation}, perceptual losses~\citep{ma2026pixelgen}, or denoising its features first~\citep{baade2026latent}. Likewise, UNIFIED trains one network in a single stage, but keeps a learned latent that the same backbone encodes and decodes.

\noindent\textbf{One transformer, many tasks.}
A single transformer can serve many tasks when the task is specified through its input. The most common way is the input format: a text prefix or prompt~\citep{raffel2020exploring,chen2022unified,lu2022unified,lu2024unified}, the choice of input and target modalities~\citep{mizrahi20234m}, in-context example pairs~\citep{wang2023images}, or a multimodal instruction~\citep{xiao2025omnigen}. Unified multimodal models~\citep{team2024chameleon,zhou2025transfusion,xie2025show,wu2025janus,deng2025emerging} extend this to understanding and generation in one backbone. Another way is the corruption level of each input, set by per-modality or per-token timesteps~\citep{bao2023one,chen2024diffusion,le2025one} or by masking ratios~\citep{li2023mage,hansen2024unified}: clean inputs act as conditions and heavily corrupted ones are generated, so one network can switch among conditional, joint, and marginal generation, or between representation learning and generation. In this paper, we use a learned mode token and a mode-specific timestep to direct a single DiT backbone between tokenization and generation.

\section{Method}
\label{sec:method}

UNIFIED replaces the encoder, generator, and decoder of latent diffusion with a
single diffusion transformer $T_\theta$ that operates in three modes: encoding,
generation, and decoding. Each training step runs one pass per mode, and every
pass is a position-aligned sequence map, in which index $i$ refers to the same
image location whether it holds a pixel patch, a latent token, or a noised latent
token. The backbone weights are fully shared; the modes differ only in
lightweight input and output heads, in a learned mode token prepended to the
sequence, and in the AdaLN conditioning, namely the timestep and a class label
that is null outside generation (Table~\ref{tab:mode-spec}).

\subsection{Preliminaries: latent diffusion and UNITE}
\label{sec:preliminaries}

Let $x\in\mathbb{R}^{H\times W\times 3}$ denote an image, which a transformer
reads as $N=HW/p^2$ patches of size $p\times p$, and let
$z\in\mathbb{R}^{K\times d}$ denote its latent, a sequence of $K$ tokens of
width $d$. $t\in[0,1]$ is the noise
level, with $t=1$ corresponding to clean data and $t=0$ to Gaussian noise, and
$c$ is the class label, with $\emptyset$ denoting the null label. A noised
latent is formed by the rectified-flow
interpolation~\citep{lipman2022flow}
\begin{equation}
    z_t=t\,z+(1-t)\,\epsilon,
    \qquad \epsilon\sim\mathcal{N}(0,I),\quad t\sim\mathcal{U}[0,1],
    \label{eq:flow-path}
\end{equation}
so that $z_1=z$ is the clean latent and $z_0$ is pure noise.

\noindent\textbf{Latent diffusion.}
The standard pipeline (Figure~\ref{fig:architectures}a) trains three networks: an encoder $E_\theta$ that maps an
image to its latent, a generator $G_\psi$ that recovers the clean latent from a
noised one, and a decoder $D_\phi$ that maps latents back to
pixels~\citep{rombach2022high,peebles2023scalable}. The encoder and decoder are
trained first under a reconstruction objective and then frozen, so generative
gradients cannot shape the latent space that the generator must later model.

\noindent\textbf{UNITE.}
UNITE~\citep{duggal2026end} observes that tokenization and generation are the same latent-inference
problem under different conditioning: tokenization infers the latent from a
fully observed image, and generation infers it from a noised latent. It therefore implements the encoding and generation
modes with one network, the \emph{generative encoder}, trained in a single
stage (Figure~\ref{fig:architectures}b). Its encoding pass reads the image patches alongside $K$ tokens
drawn from $\mathcal{N}(0,I)$ and returns their outputs as the latent, and its
generation pass places $z_t$ at the same positions. UNITE calls these tokens
registers; we call them \emph{latent readouts}, since unlike the registers
of~\citet{darcet2024vision}, whose outputs are discarded, their outputs are the
latent. We apply a stop-gradient to the encoder output $z$ in the generation loss, so the loss treats the latent as a constant. The same detached latent serves as the clean target and, once noised, as the generator’s input. Because no gradient from the generation loss reaches $z$, this loss cannot push the latent toward a trivial solution. The reconstruction and generation losses still jointly update the shared weights. Writing each pipeline in terms of the three
functions it learns, gives
\begin{alignat}{4}
    &\text{LDM:}\qquad     && z=E_\theta(x),\quad && \hat z=G_\psi(z_t,t,c),\quad   && \hat x=D_\phi(z),
    \label{eq:ldm-interface}\\
    &\text{UNITE:}\qquad   && z=E_\theta(x),\quad && \hat z=G_\theta(z_t,t,c),\quad && \hat x=D_\phi(z),
    \label{eq:unite-interface}\\
    &\text{UNIFIED:}\qquad && z=E_\theta(x),\quad && \hat z=G_\theta(z_t,t,c),\quad && \hat x=D_\theta(z).
    \label{eq:unified-interface}
\end{alignat}
\begin{table}[t]
\centering
\caption{\textbf{What each pass receives} in the shared backbone $T_\theta$.
$r\sim\mathcal{N}(0,I)$: latent readouts; $x_{\mathrm{readout}}$: pixel
readouts. UNIFIED decodes either at the clean endpoint ($t=1$) or with the
learned decoding timestep ($t=-1$).}
\label{tab:mode-spec}
\small
\resizebox{\linewidth}{!}{%
\begin{tabular}{llllccl}
\toprule
Model & Mode & Input sequence & Output & $t$ & $c$ & Length \\
\midrule
\multirow{3}{*}{UNITE++}
 & Encode   & $[q_{\mathrm{enc}};r;P_x^{\mathrm{in}}(x)]$ & $P_z^{\mathrm{out}}$ at the $K$ latent readouts & $0$ & $\emptyset$ & $1+K+N$ \\
 & Generate & $[q_{\mathrm{gen}};P_z^{\mathrm{in}}(z_t)]$ & $P_z^{\mathrm{out}}$ at positions $1{:}K$ & $t$ & $c$         & $1+K$ \\
 & Decode   & $[q_{\mathrm{dec}};P_z^{\mathrm{in}}(z);x_{\mathrm{readout}}]$ & $P_x^{\mathrm{out}}$ at the $N$ pixel readouts & $1$ & $\emptyset$ & $1+K+N$ \\
\midrule
\multirow{3}{*}{UNIFIED}
 & Encode   & $[q_{\mathrm{enc}};P_x^{\mathrm{in}}(x)]$   & $P_z^{\mathrm{out}}$ at positions $1{:}N$ & $1$ & $\emptyset$ & $N+1$ \\
 & Generate & $[q_{\mathrm{gen}};P_z^{\mathrm{in}}(z_t)]$ & $P_z^{\mathrm{out}}$ at positions $1{:}N$ & $t$ & $c$         & $N+1$ \\
 & Decode   & $[q_{\mathrm{dec}};P_z^{\mathrm{in}}(z)]$   & $P_x^{\mathrm{out}}$ at positions $1{:}N$ & $1/{-1}$ & $\emptyset$ & $N+1$ \\
\bottomrule
\end{tabular}%
}
\end{table}%
Functions with a common subscript share every backbone weight; they differ only
in lightweight input and output heads and, in UNIFIED, a learned mode token
(Section~\ref{sec:architecture}).

\noindent\textbf{A naive full-sharing baseline: UNITE++.}
UNITE reduces three networks to two, but the decoder $D_\phi$ remains a
standalone network. The interface in Equation~\ref{eq:unified-interface} does
not by itself determine how the three passes are laid out. Its most direct
realization keeps UNITE's readout layout and lets the generative encoder decode
as well; we call this baseline UNITE++ (Figure~\ref{fig:architectures}c).
Its encoding pass is UNITE's, reading the image patches alongside $K$ latent
readouts at $t_{\mathrm{enc}}=0$; its decoding pass reads the latent alongside
$N$ pixel readouts at $t_{\mathrm{dec}}=1$ and returns pixels from them; and a
learned mode token selects the mode (Table~\ref{tab:mode-spec}). UNITE++ thus
shares every backbone weight, the mode tokens, and the training objective with
UNIFIED, and differs only in the sequence layout and the encoding and decoding
timesteps. Section~\ref{sec:task-specification} motivates each of these two
choices, and Section~\ref{sec:conditioning-ablations} isolates their effects.

\subsection{UNIFIED: one backbone, three modes}
\label{sec:architecture}

UNIFIED computes all three functions in Equation~\ref{eq:unified-interface}
with one backbone $T_\theta$, a stack of DiT blocks~\citep{peebles2023scalable}
with RoPE~\citep{su2024roformer}, in which the timestep and class label reach
every block through AdaLN. The backbone takes a token sequence together with
$(t,c)$ and returns a sequence of the same length. Four lightweight heads
translate between values and the backbone width: $P_x^{\mathrm{in}}$ and
$P_x^{\mathrm{out}}$ for pixel patches, and $P_z^{\mathrm{in}}$ and
$P_z^{\mathrm{out}}$ for latent tokens. Three of the four heads,
$P_x^{\mathrm{in}}$, $P_z^{\mathrm{in}}$, and $P_z^{\mathrm{out}}$, are those
of UNITE's generative encoder and are used unchanged. The fourth,
$P_x^{\mathrm{out}}$, is the only head that UNIFIED adds: it lets the backbone
itself return pixel patches and thereby replaces UNITE's standalone decoder. A learned mode token,
$q_{\mathrm{enc}}$, $q_{\mathrm{gen}}$, or $q_{\mathrm{dec}}$, is prepended at
position~$0$ to select the mode. The three passes are
\begin{align}
    E_\theta(x)
    &=P_z^{\mathrm{out}}\,T_\theta
      \!\left([q_{\mathrm{enc}};P_x^{\mathrm{in}}(x)],\,
      t_{\mathrm{enc}},\,\emptyset\right),
      \label{eq:unified-encode}\\
    G_\theta(z_t,t,c)
    &=P_z^{\mathrm{out}}\,T_\theta
      \!\left([q_{\mathrm{gen}};P_z^{\mathrm{in}}(z_t)],\,t,\,c\right),
      \label{eq:unified-generate}\\
    D_\theta(z)
    &=P_x^{\mathrm{out}}\,T_\theta
      \!\left([q_{\mathrm{dec}};P_z^{\mathrm{in}}(z)],\,
      t_{\mathrm{dec}},\,\emptyset\right),
      \label{eq:unified-decode}
\end{align}
where $[\,\cdot\,;\,\cdot\,]$ denotes concatenation along the sequence dimension and the
output at the mode-token position is discarded before the output head. The
latent has one token per patch, $K=N$, so every pass has length $N+1$; the
encoding and decoding timesteps $t_{\mathrm{enc}}$ and $t_{\mathrm{dec}}$ are
discussed in Section~\ref{sec:task-specification}. Every backbone parameter is shared
across Equations~\ref{eq:unified-encode}--\ref{eq:unified-decode}; the only
mode-specific parameters are the four heads, the three mode tokens, and the
learned decoding timestep (Section~\ref{sec:task-specification}), which
together make up $1.63\%$ of the parameters at Base (${\sim}215$M).
Because the generation and decoding passes read latents through the same head
$P_z^{\mathrm{in}}$, the two differ only in their mode token, their
conditioning, and their output head. Table~\ref{tab:mode-spec} lists what each
pass receives.

\subsection{Specifying modes in shared weights}
\label{sec:task-specification}

With every backbone weight shared, the three passes differ only in what enters
the backbone: the values supplied by the input heads, the layout of the
sequence, the mode token, and the conditioning $(t,c)$. The class label is null
in both encoding and decoding, so the layout, the mode token, and the timestep
must keep the modes apart. We set each by a single principle.

\noindent\textbf{Position alignment.}
In UNITE and UNITE++, the latent occupies its own group of readout tokens
(Table~\ref{tab:mode-spec}), so pixels and latents interact only through
attention: encoding must route the content of every image location into latent
readouts with no fixed spatial correspondence, and in UNITE++ decoding must
route it back out to the pixel readouts. The backbone thus learns this routing
on top of encoding and decoding themselves. UNIFIED instead aligns the sequences
of all three modes: the pixel patch, latent token, and noised latent token share one sequence index and one RoPE position. This leaves attention to model context rather than to move
content between pixels and latents. Alignment comes at a cost. Readouts decouple the
latent length $K$ from the number of patches, as in Perceiver-style
readouts~\citep{jaegle2021perceiver}, while alignment ties $K=N$. In exchange, each encoding or
decoding pass has length $N+1$ instead of $1+K+N$.

\noindent\textbf{Encoding timestep.}
The timestep is the backbone's explicit signal of how much of the content at
its output positions is observed: $t=0$ marks pure noise and $t=1$ a clean
latent. We set it in every mode to match that content. The generation pass reads
$z_t$ and uses the sampled $t$. For the encoding pass, what the output positions
hold depends on the layout. In UNITE and UNITE++, they hold $\mathcal{N}(0,I)$
latent readouts, the same input from which sampling starts, so the matched
choice is the noisy endpoint, $t_{\mathrm{enc}}=0$: encoding is generation from
noise, conditioned on the image through attention. Under position alignment,
the same positions hold the observed image patches, so the matched choice
becomes the clean endpoint, $t_{\mathrm{enc}}=1$.

\noindent\textbf{Decoding timestep.}
Decoding reads a clean latent, so the principle above would set
$t_{\mathrm{dec}}=1$. At the clean endpoint, however, the modes disagree on what
to output: encoding and generation return the clean latent, whereas decoding
returns pixels. The conflict is sharpest at the end of sampling. There the
generation input $z_t\approx z$ enters through the same head $P_z^{\mathrm{in}}$
as the decoding input, and in the unconditional branch of classifier-free
guidance both passes also carry the null label, so their AdaLN conditioning is
identical and only the mode token, acting through attention, tells them apart.
The conflict surfaces in the last blocks near $t=1$, where
the two passes must diverge into different output spaces. We resolve it by
conditioning decoding on its own learned embedding in place of the timestep
embedding, denoted as $t_{\mathrm{dec}}=-1$ for timestep outside the flow
interval.

\noindent\textbf{Mode token.}
A learned mode token at position~$0$ tells the backbone which pass it is
running. It enters as a
prefix token rather than through AdaLN or mode-specific blocks, so it adds no per-block parameters and every block stays fully shared.

\subsection{Single-stage training and inference}
\label{sec:objective}

Each training step runs the three passes of Section~\ref{sec:architecture} on a
batch of images. The
encoding pass produces $z=E_\theta(x)$.
The latent is perturbed before reconstruction,
$\tilde z=z+\sigma_{\mathrm{rec}}\epsilon_r$ with
$\epsilon_r\sim\mathcal{N}(0,I)$, which makes the decoding mode robust. The decoding pass reconstructs
$\hat x=D_\theta(\tilde z)$ under
\begin{equation}
    \mathcal{L}_{\mathrm{rec}}
    =\mathbb{E}_{x,\epsilon_r}\!\left[
      \lVert \hat{x}-x\rVert_1
      +\lambda_{\mathrm{perc}}\,\mathrm{LPIPS}(\hat{x},x)\right],
    \label{eq:reconstruction-loss}
\end{equation}
where LPIPS is the perceptual distance~\citep{zhang2018unreasonable}. The
generation pass predicts the clean latent directly from the noised latent $z_t$:
\begin{equation}
    \mathcal{L}_{\mathrm{gen}}
    =\mathbb{E}_{x,t,\epsilon}
      \left[
      \left\lVert
      G_\theta\!\left(
      \operatorname{sg}(z_t),\,t,\,c
      \right)
      -\operatorname{sg}(z)
      \right\rVert_2^2
      \right],
    \label{eq:generation-loss}
\end{equation}
where $\operatorname{sg}(\cdot)$ is the stop-gradient operator, which treats its argument as a constant during backpropagation. The complete objective is
\begin{equation}
    \mathcal{L}
    =\lambda_{\mathrm{rec}}\mathcal{L}_{\mathrm{rec}}
    +\lambda_{\mathrm{gen}}\mathcal{L}_{\mathrm{gen}}.
    \label{eq:total-loss}
\end{equation}

\noindent\textbf{Gradient flow.}
As in UNITE, the stop-gradient keeps the generation loss from reaching the
encoding pass through its target, but all three passes run through $T_\theta$,
so both losses update the same weights. 

\noindent\textbf{Inference.}
Generation starts from Gaussian noise at $t=0$ and integrates the learned flow
over the full interval $t\in[0,1]$ with repeated generation passes;
classifier-free guidance is applied only for $t\in[0.1,1]$. A single decoding pass then maps the final latent to pixels.
Every step runs the same network, switching only the mode token and
conditioning. Appendix~\ref{app:eval-protocols} gives the solver, time grid, and guidance
settings of every experiment.

\begin{table}[t]
\centering
\begin{minipage}[t]{0.52\linewidth}
\vspace{0pt}
\centering
\caption{\textbf{Separate, UNITE, and UNIFIED at matched sizes} on ImageNet-1K $256\times256$, with best CFG. UNITE uses its best allocation at Tiny. $t_{\mathrm{dec}}{=}1$: decoding at clean endpoint; $t_{\mathrm{dec}}{=}{-1}$: learned decoding timestep. gFID and IS: mean over five evaluation seeds (Table~\ref{tab:seeds}); $^\dagger$: single run. Full results in Table~\ref{tab:in1k-full}.}
\label{tab:in1k}
\small
\setlength{\tabcolsep}{3pt}
\begin{tabular}{@{}llccc@{}}
\toprule
Size & Method & gFID $\downarrow$ & IS $\uparrow$ & rFID $\downarrow$ \\
\midrule
\multirow{3}{*}{Tiny}
 & UNITE ($2{:}1$) & 3.00 & 272.4 & \textbf{1.27} \\
 & UNIFIED ($t_{\mathrm{dec}}{=}1$) & \textbf{2.75} & \textbf{282.0} & 1.30 \\
 & UNIFIED ($t_{\mathrm{dec}}{=}{-1}$) & 2.78 & 280.1 & 1.75 \\
\midrule
\multirow{4}{*}{Small}
 & Separate$^\dagger$ & 5.17 & 244.8 & 1.64 \\
 & UNITE & 2.76 & 283.9 & 1.13 \\
 & UNIFIED ($t_{\mathrm{dec}}{=}1$) & \textbf{2.41} & 287.4 & \textbf{0.95} \\
 & UNIFIED ($t_{\mathrm{dec}}{=}{-1}$) & 2.43 & \textbf{295.9} & 1.04 \\
\midrule
\multirow{4}{*}{Base}
 & Separate$^\dagger$ & 3.96 & 267.7 & 1.38 \\
 & UNITE & 2.23 & 282.1 & 1.41 \\
 & UNIFIED ($t_{\mathrm{dec}}{=}1$) & 2.15 & 312.8 & 0.97 \\
 & UNIFIED ($t_{\mathrm{dec}}{=}{-1}$) & \textbf{2.08} & \textbf{314.4} & \textbf{0.95} \\
\bottomrule
\end{tabular}
\end{minipage}\hfill
\begin{minipage}[t]{0.44\linewidth}
\vspace{0pt}
\centering
\caption{\textbf{Timestep conditioning of encoding and decoding} (Base, with CFG). $t_{\mathrm{enc}}$, $t_{\mathrm{dec}}$: timesteps of the encoding and decoding passes; $t_{\mathrm{dec}}{=}{-1}$: learned decoding timestep; --: not applicable. gFID and IS: mean over five evaluation seeds (Table~\ref{tab:seeds}); $^\dagger$: single run. Full results in Table~\ref{tab:timestep-full}.}
\label{tab:timestep}
\small
\setlength{\tabcolsep}{3pt}
\begin{tabular}{@{}lccccc@{}}
\toprule
Method & $t_{\mathrm{enc}}$ & $t_{\mathrm{dec}}$ & gFID $\downarrow$ & IS $\uparrow$ & rFID $\downarrow$ \\
\midrule
UNITE   & $0$  & --  & 2.23 & 282.1 & 1.41 \\
UNITE++ & $0$ & $1$ & 2.24 & \textbf{327.1} & 2.20 \\
\midrule
UNIFIED$^\dagger$ & $0$ & $1$  & 2.31 & 300.0 & 1.39 \\
UNIFIED & $1$ & $1$  & 2.15 & 312.8 & 0.97 \\
UNIFIED & $1$ & $-1$ & \textbf{2.08} & 314.4 & \textbf{0.95} \\
\bottomrule
\end{tabular}
\end{minipage}
\end{table}

\section{Experiments}
\label{sec:experiments}

All ImageNet-1K $256\!\times\!256$ models are trained for $200$ epochs, and each optimizer step combines one reconstruction step and $12$ flow-matching steps on the same batch into a single gradient update. Otherwise, we follow the training setup of UNITE~\citep{duggal2026end}, and refer to Appendices~\ref{app:train-details} and~\ref{app:eval-protocols} for more training and evaluation details.

\subsection{Full unification improves generation at matched size}
\label{sec:imagenet-results}

Table~\ref{tab:in1k} compares three ways of spending a fixed parameter budget:
separate networks, a shared generative encoder with a standalone decoder
(UNITE), and one fully shared backbone (UNIFIED).
Both UNIFIED variants improve gFID over UNITE at every size.
The gain is largest when
capacity is scarce ($0.22$ and $0.33$ at Tiny and Small vs.\ $0.15$ at Base),
and Tiny UNIFIED nearly matches Small UNITE with about half the parameters
($2.78$ at $62$M vs.\ $2.76$ at $117$M). This suggests that the shared backbone is more parameter-efficient than separate networks.

\subsection{Comparison with pixel-space and latent diffusion models}
\label{sec:sota}

\begin{table}[t]
\centering
\caption{\textbf{Class-conditional generation on ImageNet-1K $256\times256$ with at most $250$M parameters.} Models: generator (Gen.), pretrained autoencoder (AE), pretrained vision foundation model (VFM), or standalone decoder trained jointly with the generator (Dec.). No Aux.\ Info.: \cmark{} if training uses no auxiliary pretrained representation (e.g., DINOv2 features). Repr.: the space in which diffusion runs, namely a frozen pretrained latent (Fixed), pixels (Pixel), or a latent learned jointly with the generator (Dynamic). Params counts all networks used. $^*$: UNITE trained under our protocol. UNITE$^*$ and UNIFIED: mean over five evaluation seeds. Larger models and all baselines: Table~\ref{tab:sota}.}
\label{tab:sota-small}
\small
\setlength{\tabcolsep}{3.8pt}
\begin{tabular}{@{}clccccccc@{}}
\toprule
Scale & Method & Models & \begin{tabular}[c]{@{}c@{}}No Aux.\\Info.\end{tabular} & Repr. & Epoch & Params & gFID $\downarrow$ & IS $\uparrow$ \\
\midrule
\rowcolor{unitebg}
\nocol & UNITE-T$^*$                     & Gen.+Dec.   & \cmark & Dynamic & 200   & 61M   & 3.00  & 272.4 \\
\rowcolor{unifiedbg}
\nocol\multirow{-2}{*}{${\le}$100M}
 & \textbf{UNIFIED-T (Ours)} & \textbf{Gen.} & \textbf{\cmark} & \textbf{Dynamic} & \textbf{200} & \textbf{62M} & \textbf{2.78} & \textbf{280.1} \\
\midrule
 & JiT-B/16                        & Gen.        & \cmark & Pixel   & 600   & 131M  & 3.66  & 275.1 \\
 & PixelREPA-B/16                  & Gen.+VFM    &        & Pixel   & 600   & 131M  & 3.17  & 284.6 \\
 & JiT-B/16 + RG                   & Gen.        & \cmark & Pixel   & 600   & 131M  & 2.96  & 251.1 \\
\rowcolor{unitebg}
\nocol & UNITE-S$^*$                     & Gen.+Dec.   & \cmark & Dynamic & 200   & 117M  & 2.76  & 283.9 \\
\rowcolor{unifiedbg}
\nocol\multirow{-5}{*}{100--150M}
 & \textbf{UNIFIED-S (Ours)} & \textbf{Gen.} & \textbf{\cmark} & \textbf{Dynamic} & \textbf{200} & \textbf{116M} & \textbf{2.43} & \textbf{295.9} \\
\midrule
 & PixelU-B/16                     & Gen.+VFM    &        & Pixel   & 600   & 164M  & 3.40  & 276.0 \\
 & DSD                             & Gen.+VFM    &        & Dynamic & 50    & 205M  & 3.35  & 255.0 \\
 & REPA-B                          & Gen.+AE+VFM &        & Fixed   & --    & 214M  & 2.15  & 268.3 \\
\rowcolor{unitebg}
\nocol & UNITE-B$^*$                     & Gen.+Dec.   & \cmark & Dynamic & 200   & 217M  & 2.23  & 282.1 \\
\rowcolor{unifiedbg}
\nocol\multirow{-5}{*}{150--250M}
 & \textbf{UNIFIED-B (Ours)} & \textbf{Gen.} & \textbf{\cmark} & \textbf{Dynamic} & \textbf{200} & \textbf{215M} & \textbf{2.08} & \textbf{314.4} \\
\bottomrule
\end{tabular}
\end{table}

Table~\ref{tab:sota-small} compares UNIFIED with published class-conditional generators of at most $250$M parameters on ImageNet-1K $256\times256$. Appendix~\ref{app:sota} describes all baselines and extends the comparison to larger models (Table~\ref{tab:sota}). In every parameter range, UNIFIED attains the lowest gFID without relying on a pretrained autoencoder or vision foundation model, although it is trained for only $200$ epochs, against $600$ for the pixel-space baselines. Even UNIFIED-T ($62$M, $2.78$) improves on every pixel-space baseline of $131$--$164$M ($2.96$--$3.66$), including PixelREPA-B/16 and PixelU-B/16, which align to a pretrained vision encoder. At $100$--$150$M, UNIFIED-S ($116$M) reaches $2.43$, against $2.96$ for the best pixel-space model, JiT-B/16 with register guidance (RG). At $150$--$250$M, UNIFIED-B ($215$M) reaches $2.08$ and the highest IS in the table ($314.4$), improving on the end-to-end latent model DSD ($3.35$) and on two-stage REPA-B ($2.15$), which uses both a pretrained autoencoder and a vision foundation model. In the full comparison, every model with a lower gFID than UNIFIED-B uses at least $545$M parameters.

\subsection{How are three tasks kept apart in shared weights?}
\label{sec:conditioning-ablations}

Table~\ref{tab:timestep} traces the model design path from UNITE to UNIFIED at Base level. Each step changes one factor: where the decoder lives, the sequence layout, the encoding timestep, and the decoding timestep.

\noindent\textbf{Readout tokens.}
UNITE encodes from $\mathcal{N}(0,I)$ readout tokens at $t_{\mathrm{enc}}=0$.
The natural way to fold its decoder into the backbone is UNITE++, which adds
learnable readout tokens for the image and decodes at the clean endpoint,
$t_{\mathrm{dec}}=1$. UNITE++ generates on par with UNITE but reconstructs 
worst of all models (rFID $2.20$ vs.\ $1.41$). We
attribute this to the readout design: every image location must be recovered
by attending from readout tokens to latents that have no fixed spatial
correspondence to it, an indirect route for dense decoding. UNITE++
keeps $t_{\mathrm{enc}}=0$ because its encoder still reads $\mathcal{N}(0,I)$
readouts: $t_{\mathrm{enc}}=1$ would label pure noise as clean, so the encoding
timestep can be varied only once the encoder reads image patches.

\noindent\textbf{Position alignment.}
UNIFIED removes this detour by placing each pixel patch, latent, and noisy
latent at the same sequence index. With the same timesteps as UNITE++
($t_{\mathrm{enc}}=0$, $t_{\mathrm{dec}}=1$) and the same fully shared
backbone, the layout is the only change: it improves rFID from $2.20$ to
$1.39$, but its gFID is worse ($2.31$). We attribute this to a mismatch between
the encoder's input and its timestep. The timestep tells the backbone how much
of its input is observed: $t=0$ means pure noise. In UNITE++, the encoder reads
$\mathcal{N}(0,I)$ readout tokens, so $t_{\mathrm{enc}}=0$ describes its input
correctly. In UNIFIED, the encoder instead reads clean image patches, yet
$t_{\mathrm{enc}}=0$ still labels them as pure noise, so the backbone receives
the same conditioning for a fully observed image as for the pure-noise input at
the start of generation.

\noindent\textbf{Clean endpoint for encoding.}
Encoding and decoding are both deterministic maps from fully observed inputs,
so both naturally belong at the clean endpoint. Changing only the
encoding timestep to $t_{\mathrm{enc}}=1$, with the layout and
$t_{\mathrm{dec}}=1$ fixed, improves gFID from $2.31$ to $2.15$ and rFID from
$1.39$ to $0.97$, now better than UNITE on both.

\begin{figure}[t]
    \centering
    \includegraphics[width=\linewidth]{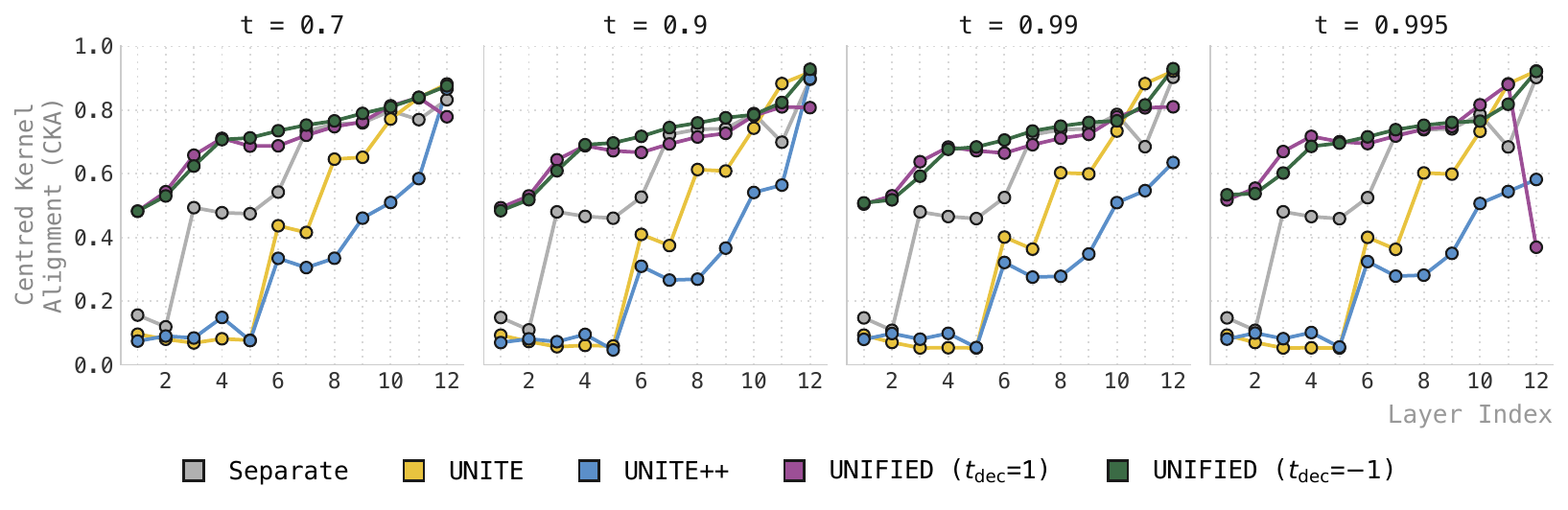}
    \vspace{-8mm}
    \caption{\textbf{Layer-wise CKA between the encoding and generation passes.} Following UNITE~\citep{duggal2026end}, we record each block's activations while encoding an image and while running the generation pass on its encoded latent corrupted to level $t$ ($t=1$ clean); each panel reports CKA~\citep{kornblith2019similarity} at one $t$.}
    \label{fig:cka}
\end{figure}

\noindent\textbf{CKA analysis.}\phantomsection\label{sec:representation-analysis}
Following UNITE, we compare what the shared backbone computes when encoding and
when generating with layer-wise CKA (Figure~\ref{fig:cka}).
Section~\ref{sec:task-specification} predicts that position alignment lets the
encoding pass track the generation pass from the early blocks, and the CKA
is consistent with this: both UNIFIED variants reach $0.48$--$0.72$ in blocks $1$--$5$ at
every $t$, whereas UNITE and UNITE++ stay at or below $0.15$. Their encoders
start from $\mathcal{N}(0,I)$ readout tokens, which carry no image information
until attention writes it in, so they align only in the last blocks
($0.84$--$0.92$ for UNITE). The one exception is the last block near the clean
endpoint. At $t=0.995$, its CKA drops in both models that decode at $t=1$,
whatever their layout ($0.37$ for UNIFIED with $t_{\mathrm{enc}}=t_{\mathrm{dec}}=1$
and $0.58$ for UNITE++), but stays at $0.92$ for UNITE, whose decoder is a
separate network. We attribute this to the output conflict at the shared
endpoint: encoding and generation output the clean latent, whereas decoding
outputs pixel features, and near $t=1$ the generation input is nearly identical
to the decoding input, so the last block must serve both output spaces.
We use CKA only to locate where the two passes diverge, while the effect of each design choice on
generation is measured by the ablations in Table~\ref{tab:timestep}.

\noindent\textbf{Learned decoding timestep.}
We therefore condition decoding on its own learned timestep,
$t_{\mathrm{dec}}=-1$. This removes the drop, keeping the last-block CKA at
$0.92$--$0.93$ from $t=0.9$ to $0.995$, on par with UNITE, and gives the best
gFID and rFID at Base ($2.08$ and $0.95$).

\subsection[Generation quality across training and sampling budgets]{Generation quality across training and sampling budgets}
\label{sec:nfe}

\begin{figure}[t]
    \centering
    \includegraphics[width=\linewidth]{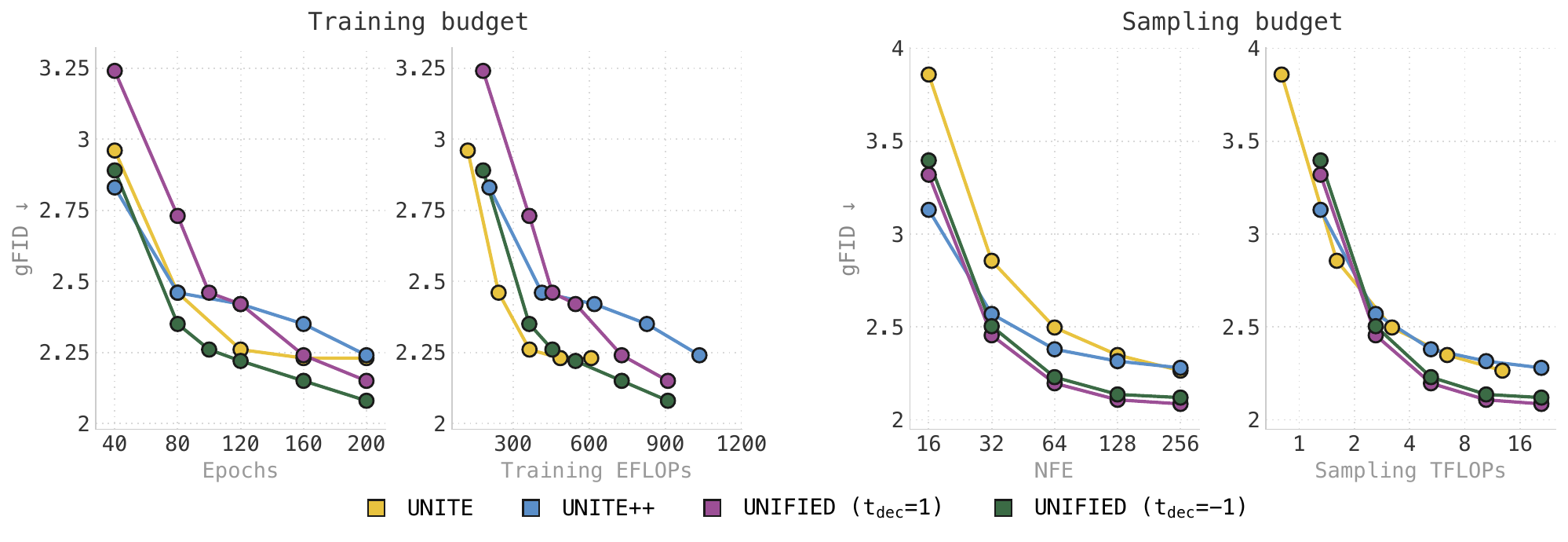}
    \caption{\textbf{Generation quality across training and sampling budgets.} Base models (${\sim}215$M) on ImageNet-1K $256\times256$, with CFG. \textbf{Training budget}: adaptive dopri5 solver over $[0,1]$, epoch-$40$ to $200$ checkpoints; epoch $200$ (five evaluation seeds) and the UNIFIED ($t_{\mathrm{dec}}{=}{-1}$) checkpoints of Table~\ref{tab:matched-flops} are means over multiple evaluation seeds, other checkpoints are single runs. \textbf{Sampling budget}: epoch-$200$ checkpoints, fixed-step Euler solver over $[0,1]$ on the shifted time grid of Appendix~\ref{app:eval-protocols}; sampling FLOPs per image $=$ NFE $\times$ one flow-matching forward pass. Small and Tiny: Figure~\ref{fig:train-budget-st}.}
    \label{fig:budget}
    \vspace{-4mm}
\end{figure}
Figure~\ref{fig:budget} varies the training and sampling budgets at Base. UNIFIED ($t_{\mathrm{dec}}{=}{-1}$) has a lower gFID than UNITE at every training checkpoint ($2.08$ vs.\ $2.23$ at epoch $200$), as it does at Small and Tiny (Figure~\ref{fig:train-budget-st}). Per training FLOP, UNITE is ahead early, but UNIFIED at $120$ epochs already matches UNITE's final gFID ($2.22$ vs.\ $2.23$) with about $10\%$ less compute (Table~\ref{tab:matched-flops}). At sampling time, UNIFIED improves on both UNITE and UNITE++ from $32$ NFE to $256$ NFE, whereas UNITE++ is best at $16$ NFE, suggesting that its separate readout pathway may help under coarser steps. UNIFIED at $64$ NFE ($2.23$) already surpasses both at $256$ NFE ($2.28$ for both), with fewer sampling FLOPs ($5.2$ vs.\ $12.8$ TFLOPs per image for UNITE). More generally, at matched quality, UNIFIED needs fewer sampling FLOPs than UNITE for any target gFID of $2.50$ or better; UNITE is cheaper only for coarser targets.

\subsection{Text-to-image generation}
\label{sec:t2i}

\begin{table}[t]
\centering
\caption{\textbf{Text-to-image generation on GenEval, DPG-Bench, and GenAI-Bench at $256\times256$.} All models are pretrained on same data (JourneyDB and BLIP3o) for $100$k iterations with the same text encoder. ``SFT'' denotes further fine-tuning on BLIP3o-60k for $2$k steps. \#Params: generator $+$ autoencoder. Higher is better for all metrics.}
\label{tab:t2i-align}
\small
\setlength{\tabcolsep}{4pt}
\resizebox{\linewidth}{!}{%
\begin{tabular}{@{}lccccccc@{}}
\toprule
 & & \multicolumn{3}{c}{Pretraining ($100$k steps)} & \multicolumn{3}{c}{SFT ($2$k steps)} \\
\cmidrule(lr){3-5} \cmidrule(l){6-8}
Method & \#Params & GenEval & DPG-Bench & GenAI-Bench & GenEval & DPG-Bench & GenAI-Bench \\
\midrule
UNITE-B                           & 219M     & 0.430 & 0.780 & 0.657 & 0.826 & 0.820 & 0.709 \\
DiT$^{\mathrm{DH}}$-XL (FLUX-VAE) & 875M+84M & 0.497 & 0.789 & 0.637 & 0.841 & 0.835 & 0.688 \\
JiT                               & 615M     & 0.516 & 0.782 & 0.674 & --    & --    & --    \\
NanoGen (SD-VAE)                  & 611M+84M & 0.578 & 0.804 & 0.691 & --    & --    & --    \\
\midrule
\rowcolor{unifiedbg}
\textbf{UNIFIED-B (Ours)}               & 218M     & 0.471 & 0.792 & 0.672 & 0.849 & 0.833 & 0.729 \\
\rowcolor{unifiedbg}
\textbf{UNIFIED-L (Ours)}               & 554M     & 0.527 & 0.794 & 0.685 & 0.878 & 0.848 & 0.748 \\
\bottomrule
\end{tabular}}
\end{table}

\begin{table}[t]
\centering
\caption{\textbf{Text-to-image generation at $512\times512$.} All models are single-stage: they are trained end to end without a pretrained autoencoder. No Aux.\ Info.: \cmark{} if training uses no auxiliary pretrained representation (e.g., DINOv2 features as a representation-alignment target), as in Table~\ref{tab:sota-small}. Results of other methods are taken from their papers. ``--'' means not reported.}
\label{tab:t2i-512}
\small
\setlength{\tabcolsep}{6pt}
\begin{tabular}{@{}lcccc@{}}
\toprule
Method & \#Params & No Aux.\ Info. & GenEval $\uparrow$ & DPG-Bench $\uparrow$ \\
\midrule
PixelFlow-XL/4   & 882M & \cmark & 0.60 & 0.779 \\
PixNerd-XXL/16   & 1.2B &        & 0.73 & 0.809 \\
PixelGen-XXL/16  & 1.1B &        & 0.79 & --    \\
DeCo-XXL/16      & 1.1B &        & 0.86 & 0.814 \\
\midrule
\rowcolor{unifiedbg}
\textbf{UNIFIED-B (Ours)}                & 218M & \textbf{\cmark} & \textbf{0.87} & \textbf{0.827} \\
\bottomrule
\end{tabular}
\end{table}

\begin{figure}[t]
\centering
\begin{minipage}[b]{0.243\linewidth}
    \centering
    {\scriptsize\parbox[b][2\baselineskip][b]{\linewidth}{\centering\textit{A bare oak tree silhouetted against a winter sunset.}}}\\[2pt]
    \includegraphics[width=\linewidth]{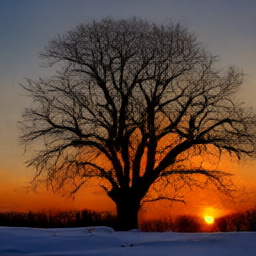}
\end{minipage}\hfill
\begin{minipage}[b]{0.243\linewidth}
    \centering
    {\scriptsize\parbox[b][2\baselineskip][b]{\linewidth}{\centering\textit{A phoenix of fire in anime}}}\\[2pt]
    \includegraphics[width=\linewidth]{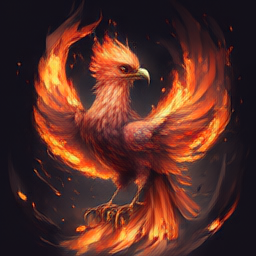}
\end{minipage}\hfill
\begin{minipage}[b]{0.243\linewidth}
    \centering
    {\scriptsize\parbox[b][2\baselineskip][b]{\linewidth}{\centering\textit{A Birman cat with blue eyes.}}}\\[2pt]
    \includegraphics[width=\linewidth]{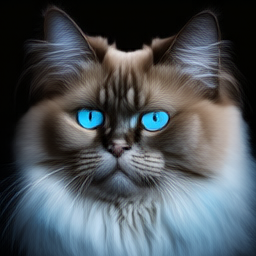}
\end{minipage}\hfill
\begin{minipage}[b]{0.243\linewidth}
    \centering
    {\scriptsize\parbox[b][2\baselineskip][b]{\linewidth}{\centering\textit{the path to Heaven}}}\\[2pt]
    \includegraphics[width=\linewidth]{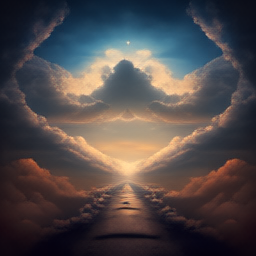}
\end{minipage}\\[4pt]
\begin{minipage}[b]{0.243\linewidth}
    \centering
    {\scriptsize\parbox[b][4\baselineskip][b]{\linewidth}{\centering\textit{a panda bear with aviator glasses on its head}}}\\[2pt]
    \includegraphics[width=\linewidth]{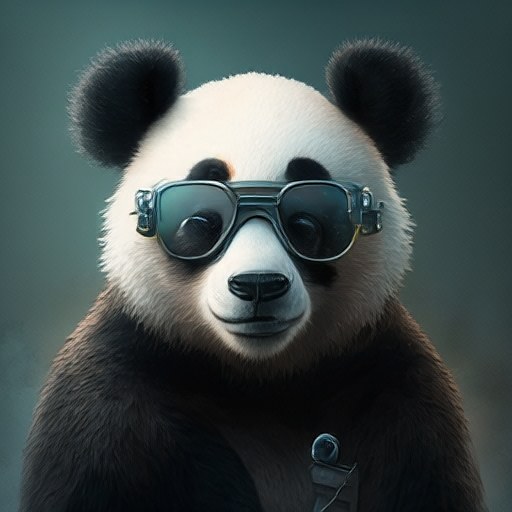}
\end{minipage}\hfill
\begin{minipage}[b]{0.243\linewidth}
    \centering
    {\scriptsize\parbox[b][4\baselineskip][b]{\linewidth}{\centering\textit{A red-haired queen wearing a green and black dress and veil depicted in an intricate and elegant digital painting.}}}\\[2pt]
    \includegraphics[width=\linewidth]{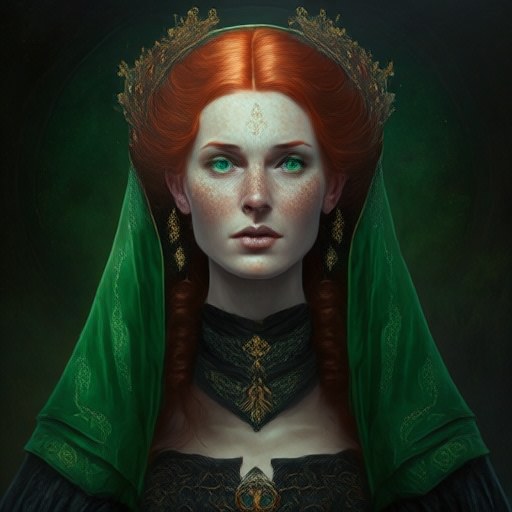}
\end{minipage}\hfill
\begin{minipage}[b]{0.243\linewidth}
    \centering
    {\scriptsize\parbox[b][4\baselineskip][b]{\linewidth}{\centering\textit{Portrait of a gecko wearing a train conductor's hat and holding a flag. Chinese ink.}}}\\[2pt]
    \includegraphics[width=\linewidth]{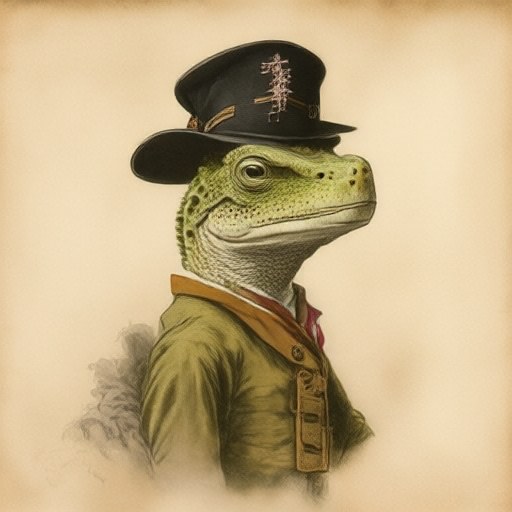}
\end{minipage}\hfill
\begin{minipage}[b]{0.243\linewidth}
    \centering
    {\scriptsize\parbox[b][4\baselineskip][b]{\linewidth}{\centering\textit{Blueberry ice cream in a stylish modern glass, ice cubes, nuts, mint leaves, splashing milk cream.}}}\\[2pt]
    \includegraphics[width=\linewidth]{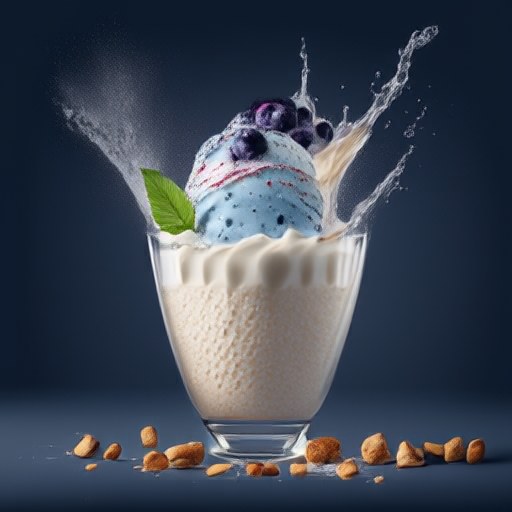}
\end{minipage}
\caption{\textbf{Text-to-image samples from UNIFIED.} \textbf{Top}: UNIFIED-L ($554$M) after $100$K pretraining steps at $256\times256$ (more in Figure~\ref{fig:t2i-prompts}). \textbf{Bottom}: UNIFIED-B ($218$M) after $10$K fine-tuning steps at $512\times512$ (more in Figure~\ref{fig:t2i-512-samples}). All samples use a $50$-step Euler solver and CFG scale $6.0$. Prompts are shown above each image.}
\label{fig:t2i-samples}
\end{figure}

\paragraph{Implementation details.} We extend UNIFIED to text-to-image generation by replacing the class-embedding conditioner with a text-encoder conditioner. We use Qwen3-0.6B~\citep{yang2025qwen3} as the text encoder and take its final-layer hidden states as the text conditioning tokens. Following DiffusionBench~\citep{leng2026diffusionbench}, we pretrain on JourneyDB~\citep{sun2023journeydb} and the long- and short-caption subsets of BLIP3-o~\citep{chen2025blip3} for $100$K iterations, with a batch size of $1024$ and $10\%$ text-condition dropout. All other settings, including the learning rate schedule, optimizer, EMA decay, gradient clipping, and sampler, follow the ImageNet setup. Optionally, we fine-tune the pretrained model on BLIP3o-60k for $2$K steps.

\paragraph{Evaluation.} All models use a $50$-step Euler solver with CFG scale $6.0$ applied over the entire integration. Table~\ref{tab:t2i-align} reports GenEval~\citep{ghosh2023geneval}, DPG-Bench~\citep{hu2024ella}, and GenAI-Bench~\citep{li2024genai}. Figure~\ref{fig:t2i-samples} shows samples. 
After pretraining, UNIFIED-L ($554$M) improves on pixel-space JiT
($615$M) on all three benchmarks, and UNIFIED-B ($218$M) is on par with JiT on
DPG-Bench and GenAI-Bench. Unlike DiT$^{\mathrm{DH}}$-XL and NanoGen, UNIFIED does
not rely on a pretrained VAE, which may have been trained on large-scale external
data; it learns its latent space from scratch during training.
After fine-tuning, UNIFIED-L improves on
DiT$^{\mathrm{DH}}$-XL with a pretrained FLUX-VAE ($959$M in total) on all three
benchmarks, and UNIFIED-B matches it on GenEval and DPG-Bench and improves on it
on GenAI-Bench. Compared with UNITE-B ($219$M), UNIFIED-B improves on all
three benchmarks both after pretraining and after fine-tuning, so
the gain of full sharing on ImageNet carries over to text-to-image generation.

\paragraph{Scaling to $512\times512$.} Following DeCo~\citep{ma2026deco} and PixelGen~\citep{ma2026pixelgen}, we pretrain UNIFIED-B at $256\times256$ for $200$K iterations, fine-tune it at $512\times512$ for $10$K iterations, and apply SFT for $2$K iterations (Appendix~\ref{app:t2i-512}). With the patch size unchanged, each image now yields $1{,}024$ latent tokens, a $4\times$ longer sequence. Table~\ref{tab:t2i-512} compares UNIFIED-B with single-stage pixel diffusion models, which, like UNIFIED, train without a pretrained autoencoder: PixelFlow~\citep{chen2025pixelflow}, PixNerd~\citep{wang2025pixnerd}, PixelGen~\citep{ma2026pixelgen}, and DeCo~\citep{ma2026deco}. UNIFIED-B ($218$M) reaches the highest GenEval ($0.87$) and DPG-Bench ($0.827$) scores. It improves on the strongest baseline, DeCo-XXL/16 ($1.1$B), with about $5\times$ fewer parameters, one eighth of its $512\times512$ fine-tuning iterations ($10$K vs.\ $80$K), and no DINO supervision, and on PixelFlow-XL/4 ($882$M), the only baseline that likewise uses no auxiliary pretrained representation, by $0.27$ on GenEval. A single fully shared network therefore still encodes, generates, and decodes well with a $4\times$ longer latent sequence. Figure~\ref{fig:t2i-samples} (bottom) and Figure~\ref{fig:t2i-512-samples} show $512\times512$ samples.

\vspace{-1mm}
\section{Conclusion}
\label{sec:conclusion}
\vspace{-1mm}
We presented UNIFIED, a single diffusion transformer that encodes, generates,
and decodes with fully shared weights. We systematically study how to unify these modes, including partial versus full weight sharing, sequence layout, and the timestep conditioning of each mode, and find that it works best when the three modes are cast as
position-aligned sequence maps and decoding receives its own learned timestep.
At matched parameters on ImageNet-1K, UNIFIED improves generation over UNITE
at every size.
This also extends to text-to-image generation, where it is competitive with
substantially larger models on GenEval, DPG-Bench, and GenAI-Bench, and at
$512\times512$ exceeds single-stage pixel diffusion models with around one
billion parameters.
Overall, full unification of latent diffusion is not merely feasible but beneficial.

\bibliography{main}

\begin{thebibliography}{92}
\providecommand{\natexlab}[1]{#1}
\providecommand{\url}[1]{\texttt{#1}}
\expandafter\ifx\csname urlstyle\endcsname\relax
  \providecommand{\doi}[1]{doi: #1}\else
  \providecommand{\doi}{doi: \begingroup \urlstyle{rm}\Url}\fi

\bibitem[Baade et~al.(2026)Baade, Chan, Sargent, Chen, Johnson, Adeli, and
  Fei-Fei]{baade2026latent}
Alan Baade, Eric~Ryan Chan, Kyle Sargent, Changan Chen, Justin Johnson, Ehsan
  Adeli, and Li~Fei-Fei.
\newblock Latent forcing: Reordering the diffusion trajectory for pixel-space
  image generation.
\newblock In \emph{ICML}, pp.\  4694--4707, 2026.

\bibitem[Bao et~al.(2023)Bao, Nie, Xue, Li, Pu, Wang, Yue, Cao, Su, and
  Zhu]{bao2023one}
Fan Bao, Shen Nie, Kaiwen Xue, Chongxuan Li, Shi Pu, Yaole Wang, Gang Yue, Yue
  Cao, Hang Su, and Jun Zhu.
\newblock One transformer fits all distributions in multi-modal diffusion at
  scale.
\newblock In \emph{ICML}, pp.\  1692--1717, 2023.

\bibitem[Bento et~al.(2020)Bento, Hersey, F{\'e}lix, Landrum, Gaulton,
  Atkinson, Bellis, De~Veij, and Leach]{bento2020open}
A~Patr{\'\i}cia Bento, Anne Hersey, Eloy F{\'e}lix, Greg Landrum, Anna Gaulton,
  Francis Atkinson, Louisa~J Bellis, Marleen De~Veij, and Andrew~R Leach.
\newblock An open source chemical structure curation pipeline using {RDKit}.
\newblock \emph{Journal of Cheminformatics}, 12\penalty0 (1):\penalty0 51,
  2020.

\bibitem[Caron et~al.(2021)Caron, Touvron, Misra, J{\'e}gou, Mairal,
  Bojanowski, and Joulin]{caron2021emerging}
Mathilde Caron, Hugo Touvron, Ishan Misra, Herv{\'e} J{\'e}gou, Julien Mairal,
  Piotr Bojanowski, and Armand Joulin.
\newblock Emerging properties in self-supervised vision transformers.
\newblock In \emph{ICCV}, pp.\  9630--9640, 2021.

\bibitem[{Chameleon Team}(2024)]{team2024chameleon}
{Chameleon Team}.
\newblock Chameleon: Mixed-modal early-fusion foundation models.
\newblock \emph{arXiv preprint arXiv:2405.09818}, 2024.

\bibitem[Chen et~al.(2024)Chen, Mart{\'\i}~Mons{\'o}, Du, Simchowitz, Tedrake,
  and Sitzmann]{chen2024diffusion}
Boyuan Chen, Diego Mart{\'\i}~Mons{\'o}, Yilun Du, Max Simchowitz, Russ
  Tedrake, and Vincent Sitzmann.
\newblock Diffusion forcing: Next-token prediction meets full-sequence
  diffusion.
\newblock In \emph{NeurIPS}, volume~37, pp.\  24081--24125, 2024.

\bibitem[Chen et~al.(2025{\natexlab{a}})Chen, Xu, Pan, Hu, Qin, Goldstein,
  Huang, Zhou, Xie, Savarese, Xue, Xiong, and Xu]{chen2025blip3}
Jiuhai Chen, Zhiyang Xu, Xichen Pan, Yushi Hu, Can Qin, Tom Goldstein, Lifu
  Huang, Tianyi Zhou, Saining Xie, Silvio Savarese, Le~Xue, Caiming Xiong, and
  Ran Xu.
\newblock {BLIP3-o}: A family of fully open unified multimodal
  models---architecture, training and dataset.
\newblock \emph{arXiv preprint arXiv:2505.09568}, 2025{\natexlab{a}}.

\bibitem[Chen et~al.(2025{\natexlab{b}})Chen, Cai, Chen, Xie, Yang, Tang, Li,
  and Han]{chen2025deep}
Junyu Chen, Han Cai, Junsong Chen, Enze Xie, Shang Yang, Haotian Tang, Muyang
  Li, and Song Han.
\newblock Deep compression autoencoder for efficient high-resolution diffusion
  models.
\newblock In \emph{ICLR}, 2025{\natexlab{b}}.

\bibitem[Chen et~al.(2025{\natexlab{c}})Chen, Ge, Zhang, Sun, and
  Luo]{chen2025pixelflow}
Shoufa Chen, Chongjian Ge, Shilong Zhang, Peize Sun, and Ping Luo.
\newblock {PixelFlow}: Pixel-space generative models with flow.
\newblock \emph{arXiv preprint arXiv:2504.07963}, 2025{\natexlab{c}}.

\bibitem[Chen et~al.(2022)Chen, Saxena, Li, Lin, Fleet, and
  Hinton]{chen2022unified}
Ting Chen, Saurabh Saxena, Lala Li, Tsung-Yi Lin, David Fleet, and Geoffrey~E
  Hinton.
\newblock A unified sequence interface for vision tasks.
\newblock In \emph{NeurIPS}, volume~35, pp.\  31333--31346, 2022.

\bibitem[Darcet et~al.(2024)Darcet, Oquab, Mairal, and
  Bojanowski]{darcet2024vision}
Timoth{\'e}e Darcet, Maxime Oquab, Julien Mairal, and Piotr Bojanowski.
\newblock Vision transformers need registers.
\newblock In \emph{ICLR}, 2024.

\bibitem[Deng et~al.(2025)Deng, Zhu, Li, Gou, Li, Wang, Zhong, Yu, Nie, Song,
  et~al.]{deng2025emerging}
Chaorui Deng, Deyao Zhu, Kunchang Li, Chenhui Gou, Feng Li, Zeyu Wang, Shu
  Zhong, Weihao Yu, Xiaonan Nie, Ziang Song, et~al.
\newblock Emerging properties in unified multimodal pretraining.
\newblock \emph{arXiv preprint arXiv:2505.14683}, 2025.

\bibitem[Dhariwal \& Nichol(2021)Dhariwal and Nichol]{dhariwal2021diffusion}
Prafulla Dhariwal and Alexander Nichol.
\newblock Diffusion models beat {GAN}s on image synthesis.
\newblock In \emph{NeurIPS}, pp.\  8780--8794, 2021.

\bibitem[Duggal et~al.(2026)Duggal, Bai, Wu, Zhang, Shechtman, Torralba, Isola,
  and Freeman]{duggal2026end}
Shivam Duggal, Xingjian Bai, Zongze Wu, Richard Zhang, Eli Shechtman, Antonio
  Torralba, Phillip Isola, and William~T Freeman.
\newblock End-to-end training for unified tokenization and latent denoising.
\newblock \emph{arXiv preprint arXiv:2603.22283}, 2026.

\bibitem[Esser et~al.(2021)Esser, Rombach, and Ommer]{esser2021taming}
Patrick Esser, Robin Rombach, and Bj{\"o}rn Ommer.
\newblock Taming transformers for high-resolution image synthesis.
\newblock In \emph{CVPR}, pp.\  12873--12883, 2021.

\bibitem[Esser et~al.(2024)Esser, Kulal, Blattmann, Entezari, M{\"u}ller,
  Saini, Levi, Lorenz, Sauer, Boesel, et~al.]{esser2024scaling}
Patrick Esser, Sumith Kulal, Andreas Blattmann, Rahim Entezari, Jonas
  M{\"u}ller, Harry Saini, Yam Levi, Dominik Lorenz, Axel Sauer, Frederic
  Boesel, et~al.
\newblock Scaling rectified flow transformers for high-resolution image
  synthesis.
\newblock In \emph{ICML}, 2024.

\bibitem[Gao et~al.(2023)Gao, Zhou, Cheng, and Yan]{gao2023masked}
Shanghua Gao, Pan Zhou, Ming-Ming Cheng, and Shuicheng Yan.
\newblock Masked diffusion transformer is a strong image synthesizer.
\newblock In \emph{ICCV}, pp.\  23164--23173, 2023.

\bibitem[Ghosh et~al.(2023)Ghosh, Hajishirzi, and Schmidt]{ghosh2023geneval}
Dhruba Ghosh, Hannaneh Hajishirzi, and Ludwig Schmidt.
\newblock {GenEval}: An object-focused framework for evaluating text-to-image
  alignment.
\newblock In \emph{NeurIPS}, pp.\  52132--52152, 2023.

\bibitem[Guo et~al.(2026)Guo, Ma, He, Fu, Fu, Huang, and Li]{guo2026pixelu}
Zipeng Guo, Lichen Ma, Yu~He, Xiaolong Fu, Jingling Fu, Junshi Huang, and Yan
  Li.
\newblock {PixelU}: A {U}-shaped transformer for efficient end-to-end pixel
  diffusion.
\newblock In \emph{ECCV}, pp.\  425--442, 2026.

\bibitem[Hansen-Estruch et~al.(2024)Hansen-Estruch, Vishwanath, Zhang, and
  Tomar]{hansen2024unified}
Philippe Hansen-Estruch, Sriram Vishwanath, Amy Zhang, and Manan Tomar.
\newblock Unified auto-encoding with masked diffusion.
\newblock \emph{arXiv preprint arXiv:2406.17688}, 2024.

\bibitem[Hansen-Estruch et~al.(2025)Hansen-Estruch, Yan, Chuang, Zohar, Wang,
  Hou, Xu, Vishwanath, Vajda, and Chen]{hansen2025learnings}
Philippe Hansen-Estruch, David Yan, Ching-Yao Chuang, Orr Zohar, Jialiang Wang,
  Tingbo Hou, Tao Xu, Sriram Vishwanath, Peter Vajda, and Xinlei Chen.
\newblock Learnings from scaling visual tokenizers for reconstruction and
  generation.
\newblock In \emph{ICML}, pp.\  22023--22043, 2025.

\bibitem[Heek et~al.(2026)Heek, Hoogeboom, Mensink, and
  Salimans]{heek2026unified}
Jonathan Heek, Emiel Hoogeboom, Thomas Mensink, and Tim Salimans.
\newblock Unified latents ({UL}): How to train your latents.
\newblock \emph{arXiv preprint arXiv:2602.17270}, 2026.

\bibitem[Heusel et~al.(2017)Heusel, Ramsauer, Unterthiner, Nessler, and
  Hochreiter]{heusel2017gans}
Martin Heusel, Hubert Ramsauer, Thomas Unterthiner, Bernhard Nessler, and Sepp
  Hochreiter.
\newblock {GANs} trained by a two time-scale update rule converge to a local
  {N}ash equilibrium.
\newblock In \emph{NeurIPS}, 2017.

\bibitem[Hoogeboom et~al.(2023)Hoogeboom, Heek, and
  Salimans]{hoogeboom2023simple}
Emiel Hoogeboom, Jonathan Heek, and Tim Salimans.
\newblock simple diffusion: End-to-end diffusion for high resolution images.
\newblock In \emph{ICML}, pp.\  13213--13232, 2023.

\bibitem[Hoogeboom et~al.(2025)Hoogeboom, Mensink, Heek, Lamerigts, Gao, and
  Salimans]{hoogeboom2025simpler}
Emiel Hoogeboom, Thomas Mensink, Jonathan Heek, Kay Lamerigts, Ruiqi Gao, and
  Tim Salimans.
\newblock Simpler diffusion: 1.5 {FID} on {ImageNet512} with pixel-space
  diffusion.
\newblock In \emph{CVPR}, 2025.

\bibitem[Hu et~al.(2024)Hu, Wang, Fang, Fu, Cheng, and Yu]{hu2024ella}
Xiwei Hu, Rui Wang, Yixiao Fang, Bin Fu, Pei Cheng, and Gang Yu.
\newblock {ELLA}: Equip diffusion models with {LLM} for enhanced semantic
  alignment.
\newblock \emph{arXiv preprint arXiv:2403.05135}, 2024.

\bibitem[Jabri et~al.(2023)Jabri, Fleet, and Chen]{jabri2022scalable}
Allan Jabri, David~J Fleet, and Ting Chen.
\newblock Scalable adaptive computation for iterative generation.
\newblock In \emph{ICML}, pp.\  14569--14589, 2023.

\bibitem[Jaegle et~al.(2022)Jaegle, Borgeaud, Alayrac, Doersch, Ionescu, Ding,
  Koppula, Zoran, Brock, Shelhamer, et~al.]{jaegle2021perceiver}
Andrew Jaegle, Sebastian Borgeaud, Jean-Baptiste Alayrac, Carl Doersch, Catalin
  Ionescu, David Ding, Skanda Koppula, Daniel Zoran, Andrew Brock, Evan
  Shelhamer, et~al.
\newblock Perceiver {IO}: A general architecture for structured inputs \&
  outputs.
\newblock In \emph{ICLR}, 2022.

\bibitem[Joshi et~al.(2025)Joshi, Fu, Liao, Gharakhanyan, Miller, Sriram, and
  Ulissi]{joshi2025all}
Chaitanya~K Joshi, Xiang Fu, Yi-Lun Liao, Vahe Gharakhanyan, Benjamin~Kurt
  Miller, Anuroop Sriram, and Zachary~W Ulissi.
\newblock All-atom diffusion transformers: Unified generative modelling of
  molecules and materials.
\newblock In \emph{ICML}, pp.\  28393--28417, 2025.

\bibitem[Kingma \& Gao(2023)Kingma and Gao]{kingma2023understanding}
Diederik~P Kingma and Ruiqi Gao.
\newblock Understanding diffusion objectives as the {ELBO} with simple data
  augmentation.
\newblock In \emph{NeurIPS}, volume~36, 2023.

\bibitem[Kingma \& Welling(2014)Kingma and Welling]{kingma2013auto}
Diederik~P Kingma and Max Welling.
\newblock Auto-encoding variational {B}ayes.
\newblock In \emph{ICLR}, 2014.

\bibitem[Kornblith et~al.(2019)Kornblith, Norouzi, Lee, and
  Hinton]{kornblith2019similarity}
Simon Kornblith, Mohammad Norouzi, Honglak Lee, and Geoffrey Hinton.
\newblock Similarity of neural network representations revisited.
\newblock In \emph{ICML}, pp.\  3519--3529, 2019.

\bibitem[Le et~al.(2025)Le, Pham, Lee, Clark, Kembhavi, Mandt, Krishna, and
  Lu]{le2025one}
Duong~H Le, Tuan Pham, Sangho Lee, Christopher Clark, Aniruddha Kembhavi,
  Stephan Mandt, Ranjay Krishna, and Jiasen Lu.
\newblock One diffusion to generate them all.
\newblock In \emph{CVPR}, pp.\  2671--2682, 2025.

\bibitem[Leng et~al.(2025)Leng, Singh, Hou, Xing, Xie, and Zheng]{leng2025repa}
Xingjian Leng, Jaskirat Singh, Yunzhong Hou, Zhenchang Xing, Saining Xie, and
  Liang Zheng.
\newblock {REPA-E}: Unlocking {VAE} for end-to-end tuning with latent diffusion
  transformers.
\newblock In \emph{ICCV}, pp.\  18262--18272, 2025.

\bibitem[Leng et~al.(2026)Leng, Singh, Liang, Smith, Bell, Saha, Yuan, and
  Zheng]{leng2026diffusionbench}
Xingjian Leng, Jaskirat Singh, Zhanhao Liang, Ethan Smith, Martin Bell, Aninda
  Saha, Yuhui Yuan, and Liang Zheng.
\newblock {DiffusionBench}: On holistic evaluation of diffusion transformers.
\newblock \emph{arXiv preprint arXiv:2606.24888}, 2026.

\bibitem[Li et~al.(2024)Li, Lin, Pathak, Li, Fei, Wu, Ling, Xia, Zhang, Neubig,
  and Ramanan]{li2024genai}
Baiqi Li, Zhiqiu Lin, Deepak Pathak, Jiayao Li, Yixin Fei, Kewen Wu, Tiffany
  Ling, Xide Xia, Pengchuan Zhang, Graham Neubig, and Deva Ramanan.
\newblock {GenAI-Bench}: Evaluating and improving compositional text-to-visual
  generation.
\newblock \emph{arXiv preprint arXiv:2406.13743}, 2024.

\bibitem[Li \& He(2026)Li and He]{li2026back}
Tianhong Li and Kaiming He.
\newblock Back to basics: Let denoising generative models denoise.
\newblock In \emph{CVPR}, pp.\  36115--36125, 2026.

\bibitem[Li et~al.(2023)Li, Chang, Mishra, Zhang, Katabi, and
  Krishnan]{li2023mage}
Tianhong Li, Huiwen Chang, Shlok Mishra, Han Zhang, Dina Katabi, and Dilip
  Krishnan.
\newblock {MAGE}: Masked generative encoder to unify representation learning
  and image synthesis.
\newblock In \emph{CVPR}, pp.\  2142--2152, 2023.

\bibitem[Lin et~al.(2014)Lin, Maire, Belongie, Hays, Perona, Ramanan,
  Doll{\'a}r, and Zitnick]{lin2014microsoft}
Tsung-Yi Lin, Michael Maire, Serge Belongie, James Hays, Pietro Perona, Deva
  Ramanan, Piotr Doll{\'a}r, and C.~Lawrence Zitnick.
\newblock Microsoft {COCO}: Common objects in context.
\newblock In \emph{ECCV}, pp.\  740--755, 2014.

\bibitem[Lipman et~al.(2023)Lipman, Chen, Ben-Hamu, Nickel, and
  Le]{lipman2022flow}
Yaron Lipman, Ricky~TQ Chen, Heli Ben-Hamu, Maximilian Nickel, and Matt Le.
\newblock Flow matching for generative modeling.
\newblock In \emph{ICLR}, 2023.

\bibitem[Liu et~al.(2023)Liu, Gong, and Liu]{liu2022flow}
Xingchao Liu, Chengyue Gong, and Qiang Liu.
\newblock Flow straight and fast: Learning to generate and transfer data with
  rectified flow.
\newblock In \emph{ICLR}, 2023.

\bibitem[Lu et~al.(2023)Lu, Clark, Zellers, Mottaghi, and
  Kembhavi]{lu2022unified}
Jiasen Lu, Christopher Clark, Rowan Zellers, Roozbeh Mottaghi, and Aniruddha
  Kembhavi.
\newblock {Unified-IO}: A unified model for vision, language, and multi-modal
  tasks.
\newblock In \emph{ICLR}, 2023.

\bibitem[Lu et~al.(2024{\natexlab{a}})Lu, Clark, Lee, Zhang, Khosla, Marten,
  Hoiem, and Kembhavi]{lu2024unified}
Jiasen Lu, Christopher Clark, Sangho Lee, Zichen Zhang, Savya Khosla, Ryan
  Marten, Derek Hoiem, and Aniruddha Kembhavi.
\newblock {Unified-IO 2}: Scaling autoregressive multimodal models with vision,
  language, audio, and action.
\newblock In \emph{CVPR}, pp.\  26429--26445, 2024{\natexlab{a}}.

\bibitem[Lu et~al.(2024{\natexlab{b}})Lu, Wang, Huang, Wu, Liu, Ouyang, and
  Bai]{lu2024fit}
Zeyu Lu, Zidong Wang, Di~Huang, Chengyue Wu, Xihui Liu, Wanli Ouyang, and Lei
  Bai.
\newblock {FiT}: Flexible vision transformer for diffusion model.
\newblock In \emph{ICML}, pp.\  33160--33176, 2024{\natexlab{b}}.

\bibitem[Ma et~al.(2024)Ma, Goldstein, Albergo, Boffi, Vanden-Eijnden, and
  Xie]{ma2024sit}
Nanye Ma, Mark Goldstein, Michael~S Albergo, Nicholas~M Boffi, Eric
  Vanden-Eijnden, and Saining Xie.
\newblock {SiT}: Exploring flow and diffusion-based generative models with
  scalable interpolant transformers.
\newblock In \emph{ECCV}, pp.\  23--40, 2024.

\bibitem[Ma et~al.(2026{\natexlab{a}})Ma, Wei, Wang, Zhang, and
  Tian]{ma2026deco}
Zehong Ma, Longhui Wei, Shuai Wang, Shiliang Zhang, and Qi~Tian.
\newblock {DeCo}: Frequency-decoupled pixel diffusion for end-to-end image
  generation.
\newblock In \emph{CVPR}, pp.\  43600--43610, 2026{\natexlab{a}}.

\bibitem[Ma et~al.(2026{\natexlab{b}})Ma, Xu, and Zhang]{ma2026pixelgen}
Zehong Ma, Ruihan Xu, and Shiliang Zhang.
\newblock {PixelGen}: Improving pixel diffusion with perceptual supervision.
\newblock \emph{arXiv preprint arXiv:2602.02493}, 2026{\natexlab{b}}.

\bibitem[Mentzer et~al.(2024)Mentzer, Minnen, Agustsson, and
  Tschannen]{mentzer2024finite}
Fabian Mentzer, David Minnen, Eirikur Agustsson, and Michael Tschannen.
\newblock Finite scalar quantization: {VQ-VAE} made simple.
\newblock In \emph{ICLR}, 2024.

\bibitem[Mizrahi et~al.(2023)Mizrahi, Bachmann, Kar, Yeo, Gao, Dehghan, and
  Zamir]{mizrahi20234m}
David Mizrahi, Roman Bachmann, Oguzhan Kar, Teresa Yeo, Mingfei Gao, Afshin
  Dehghan, and Amir Zamir.
\newblock {4M}: Massively multimodal masked modeling.
\newblock In \emph{NeurIPS}, volume~36, pp.\  58363--58408, 2023.

\bibitem[Nichol et~al.(2022)Nichol, Dhariwal, Ramesh, Shyam, Mishkin, McGrew,
  Sutskever, and Chen]{nichol2021glide}
Alexander~Quinn Nichol, Prafulla Dhariwal, Aditya Ramesh, Pranav Shyam, Pamela
  Mishkin, Bob McGrew, Ilya Sutskever, and Mark Chen.
\newblock {GLIDE}: Towards photorealistic image generation and editing with
  text-guided diffusion models.
\newblock In \emph{ICML}, pp.\  16784--16804, 2022.

\bibitem[Obukhov et~al.(2020)Obukhov, Seitzer, Wu, Zhydenko, Kyl, and
  Lin]{obukhov2020torchfidelity}
Anton Obukhov, Maximilian Seitzer, Po-Wei Wu, Semen Zhydenko, Jonathan Kyl, and
  Elvis Yu-Jing Lin.
\newblock High-fidelity performance metrics for generative models in {PyTorch}.
\newblock \url{https://github.com/toshas/torch-fidelity}, 2020.

\bibitem[Ong et~al.(2013)Ong, Richards, Jain, Hautier, Kocher, Cholia, Gunter,
  Chevrier, Persson, and Ceder]{ong2013python}
Shyue~Ping Ong, William~Davidson Richards, Anubhav Jain, Geoffroy Hautier,
  Michael Kocher, Shreyas Cholia, Dan Gunter, Vincent~L Chevrier, Kristin~A
  Persson, and Gerbrand Ceder.
\newblock {Python} {Materials Genomics} (pymatgen): A robust, open-source
  python library for materials analysis.
\newblock \emph{Computational Materials Science}, 68:\penalty0 314--319, 2013.

\bibitem[Oquab et~al.(2024)Oquab, Darcet, Moutakanni, Vo, Szafraniec, Khalidov,
  Fernandez, Haziza, Massa, El-Nouby, et~al.]{oquab2023dinov2}
Maxime Oquab, Timoth{\'e}e Darcet, Th{\'e}o Moutakanni, Huy Vo, Marc
  Szafraniec, Vasil Khalidov, Pierre Fernandez, Daniel Haziza, Francisco Massa,
  Alaaeldin El-Nouby, et~al.
\newblock {DINOv2}: Learning robust visual features without supervision.
\newblock \emph{TMLR}, 2024.

\bibitem[Peebles \& Xie(2023)Peebles and Xie]{peebles2023scalable}
William Peebles and Saining Xie.
\newblock Scalable diffusion models with transformers.
\newblock In \emph{ICCV}, pp.\  4172--4182, 2023.

\bibitem[Podell et~al.(2024)Podell, English, Lacey, Blattmann, Dockhorn,
  M{\"u}ller, Penna, and Rombach]{podell2024sdxl}
Dustin Podell, Zion English, Kyle Lacey, Andreas Blattmann, Tim Dockhorn, Jonas
  M{\"u}ller, Joe Penna, and Robin Rombach.
\newblock {SDXL}: Improving latent diffusion models for high-resolution image
  synthesis.
\newblock In \emph{ICLR}, 2024.

\bibitem[Raffel et~al.(2020)Raffel, Shazeer, Roberts, Lee, Narang, Matena,
  Zhou, Li, and Liu]{raffel2020exploring}
Colin Raffel, Noam Shazeer, Adam Roberts, Katherine Lee, Sharan Narang, Michael
  Matena, Yanqi Zhou, Wei Li, and Peter~J Liu.
\newblock Exploring the limits of transfer learning with a unified text-to-text
  transformer.
\newblock \emph{Journal of Machine Learning Research}, 21\penalty0
  (140):\penalty0 1--67, 2020.

\bibitem[Ramakrishnan et~al.(2014)Ramakrishnan, Dral, Rupp, and von
  Lilienfeld]{ramakrishnan2014quantum}
Raghunathan Ramakrishnan, Pavlo~O Dral, Matthias Rupp, and O~Anatole von
  Lilienfeld.
\newblock Quantum chemistry structures and properties of 134 kilo molecules.
\newblock \emph{Scientific Data}, 1:\penalty0 140022, 2014.

\bibitem[Ramesh et~al.(2021)Ramesh, Pavlov, Goh, Gray, Voss, Radford, Chen, and
  Sutskever]{ramesh2021zero}
Aditya Ramesh, Mikhail Pavlov, Gabriel Goh, Scott Gray, Chelsea Voss, Alec
  Radford, Mark Chen, and Ilya Sutskever.
\newblock Zero-shot text-to-image generation.
\newblock In \emph{ICML}, pp.\  8821--8831, 2021.

\bibitem[Razavi et~al.(2019)Razavi, {van den Oord}, and
  Vinyals]{razavi2019generating}
Ali Razavi, Aaron {van den Oord}, and Oriol Vinyals.
\newblock Generating diverse high-fidelity images with {VQ-VAE-2}.
\newblock In \emph{NeurIPS}, volume~32, 2019.

\bibitem[Rezende et~al.(2014)Rezende, Mohamed, and
  Wierstra]{rezende2014stochastic}
Danilo~Jimenez Rezende, Shakir Mohamed, and Daan Wierstra.
\newblock Stochastic backpropagation and approximate inference in deep
  generative models.
\newblock In \emph{ICML}, pp.\  1278--1286, 2014.

\bibitem[Rombach et~al.(2022)Rombach, Blattmann, Lorenz, Esser, and
  Ommer]{rombach2022high}
Robin Rombach, Andreas Blattmann, Dominik Lorenz, Patrick Esser, and Bj{\"o}rn
  Ommer.
\newblock High-resolution image synthesis with latent diffusion models.
\newblock In \emph{CVPR}, pp.\  10674--10685, 2022.

\bibitem[Saharia et~al.(2022)Saharia, Chan, Saxena, Li, Whang, Denton,
  Ghasemipour, Gontijo~Lopes, Karagol~Ayan, Salimans, Ho, Fleet, and
  Norouzi]{saharia2022photorealistic}
Chitwan Saharia, William Chan, Saurabh Saxena, Lala Li, Jay Whang, Emily~L
  Denton, Seyed Kamyar~Seyed Ghasemipour, Raphael Gontijo~Lopes, Burcu
  Karagol~Ayan, Tim Salimans, Jonathan Ho, David~J Fleet, and Mohammad Norouzi.
\newblock Photorealistic text-to-image diffusion models with deep language
  understanding.
\newblock In \emph{NeurIPS}, pp.\  36479--36494, 2022.

\bibitem[Shi et~al.(2026)Shi, Wang, Zheng, Yuan, Wu, Wang, Wan, Zhou, and
  Lu]{shi2026latent}
Minglei Shi, Haolin Wang, Wenzhao Zheng, Ziyang Yuan, Xiaoshi Wu, Xintao Wang,
  Pengfei Wan, Jie Zhou, and Jiwen Lu.
\newblock Latent diffusion model without variational autoencoder.
\newblock In \emph{ICLR}, pp.\  154506--154537, 2026.

\bibitem[Shin et~al.(2026)Shin, Kim, and Shim]{shin2026representation}
Jaeyo Shin, Jiwook Kim, and Hyunjung Shim.
\newblock Representation alignment for just image transformers is not easier
  than you think.
\newblock In \emph{ECCV}, pp.\  603--620, 2026.

\bibitem[Song et~al.(2012)Song, Smola, Gretton, Bedo, and
  Borgwardt]{song2012feature}
Le~Song, Alex Smola, Arthur Gretton, Justin Bedo, and Karsten Borgwardt.
\newblock Feature selection via dependence maximization.
\newblock \emph{Journal of Machine Learning Research}, 13:\penalty0 1393--1434,
  2012.

\bibitem[Starodubcev et~al.(2026)Starodubcev, Sudakov, Drobyshevskiy, Babenko,
  and Baranchuk]{starodubcev2026registers}
Nikita Starodubcev, Ilia Sudakov, Ilya Drobyshevskiy, Artem Babenko, and Dmitry
  Baranchuk.
\newblock Registers matter for pixel-space diffusion transformers.
\newblock \emph{arXiv preprint arXiv:2605.16147}, 2026.

\bibitem[Su et~al.(2024)Su, Ahmed, Lu, Pan, Bo, and Liu]{su2024roformer}
Jianlin Su, Murtadha Ahmed, Yu~Lu, Shengfeng Pan, Wen Bo, and Yunfeng Liu.
\newblock {RoFormer}: Enhanced transformer with rotary position embedding.
\newblock \emph{Neurocomputing}, 568:\penalty0 127063, 2024.

\bibitem[Sun et~al.(2023)Sun, Pan, Ge, Li, Duan, Wu, Zhang, Zhou, Qin, Wang,
  Dai, Qiao, Wang, and Li]{sun2023journeydb}
Keqiang Sun, Junting Pan, Yuying Ge, Hao Li, Haodong Duan, Xiaoshi Wu, Renrui
  Zhang, Aojun Zhou, Zipeng Qin, Yi~Wang, Jifeng Dai, Yu~Qiao, Limin Wang, and
  Hongsheng Li.
\newblock {JourneyDB}: A benchmark for generative image understanding.
\newblock In \emph{NeurIPS}, pp.\  49659--49678, 2023.

\bibitem[Szegedy et~al.(2016)Szegedy, Vanhoucke, Ioffe, Shlens, and
  Wojna]{szegedy2016rethinking}
Christian Szegedy, Vincent Vanhoucke, Sergey Ioffe, Jon Shlens, and Zbigniew
  Wojna.
\newblock Rethinking the inception architecture for computer vision.
\newblock In \emph{CVPR}, pp.\  2818--2826, 2016.

\bibitem[Vahdat et~al.(2021)Vahdat, Kreis, and Kautz]{vahdat2021score}
Arash Vahdat, Karsten Kreis, and Jan Kautz.
\newblock Score-based generative modeling in latent space.
\newblock In \emph{NeurIPS}, volume~34, pp.\  11287--11302, 2021.

\bibitem[{van den Oord} et~al.(2017){van den Oord}, Vinyals, and
  Kavukcuoglu]{van2017neural}
Aaron {van den Oord}, Oriol Vinyals, and Koray Kavukcuoglu.
\newblock Neural discrete representation learning.
\newblock In \emph{NeurIPS}, volume~30, 2017.

\bibitem[Wang et~al.(2026{\natexlab{a}})Wang, Gao, Zhu, Huang, and
  Wang]{wang2025pixnerd}
Shuai Wang, Ziteng Gao, Chenhui Zhu, Weilin Huang, and Limin Wang.
\newblock {PixNerd}: Pixel neural field diffusion.
\newblock In \emph{ICLR}, 2026{\natexlab{a}}.

\bibitem[Wang et~al.(2026{\natexlab{b}})Wang, Tian, Huang, and
  Wang]{wang2025ddt}
Shuai Wang, Zhi Tian, Weilin Huang, and Limin Wang.
\newblock {DDT}: Decoupled diffusion transformer.
\newblock In \emph{CVPR}, pp.\  40633--40642, 2026{\natexlab{b}}.

\bibitem[Wang et~al.(2023)Wang, Wang, Cao, Shen, and Huang]{wang2023images}
Xinlong Wang, Wen Wang, Yue Cao, Chunhua Shen, and Tiejun Huang.
\newblock Images speak in images: A generalist {P}ainter for in-context visual
  learning.
\newblock In \emph{CVPR}, pp.\  6830--6839, 2023.

\bibitem[Wang \& Zhang(2025)Wang and Zhang]{wang2025diffusion}
Xiyuan Wang and Muhan Zhang.
\newblock Diffusion as self-distillation: End-to-end latent diffusion in one
  model.
\newblock \emph{arXiv preprint arXiv:2511.14716}, 2025.

\bibitem[Wu et~al.(2025)Wu, Chen, Wu, Ma, Liu, Pan, Liu, Xie, Yu, Ruan,
  et~al.]{wu2025janus}
Chengyue Wu, Xiaokang Chen, Zhiyu Wu, Yiyang Ma, Xingchao Liu, Zizheng Pan, Wen
  Liu, Zhenda Xie, Xingkai Yu, Chong Ruan, et~al.
\newblock Janus: Decoupling visual encoding for unified multimodal
  understanding and generation.
\newblock In \emph{CVPR}, pp.\  12966--12977, 2025.

\bibitem[Xiao et~al.(2025)Xiao, Wang, Zhou, Yuan, Xing, Yan, Li, Wang, Huang,
  and Liu]{xiao2025omnigen}
Shitao Xiao, Yueze Wang, Junjie Zhou, Huaying Yuan, Xingrun Xing, Ruiran Yan,
  Chaofan Li, Shuting Wang, Tiejun Huang, and Zheng Liu.
\newblock {OmniGen}: Unified image generation.
\newblock In \emph{CVPR}, pp.\  13294--13304, 2025.

\bibitem[Xie et~al.(2025)Xie, Mao, Bai, Zhang, Wang, Lin, Gu, Chen, Yang, and
  Shou]{xie2025show}
Jinheng Xie, Weijia Mao, Zechen Bai, David~Junhao Zhang, Weihao Wang,
  Kevin~Qinghong Lin, Yuchao Gu, Zhijie Chen, Zhenheng Yang, and Mike~Zheng
  Shou.
\newblock Show-o: One single transformer to unify multimodal understanding and
  generation.
\newblock In \emph{ICLR}, 2025.

\bibitem[Yang et~al.(2025)Yang, Li, Yang, Zhang, Hui, Zheng, Yu, Gao, Huang,
  Lv, et~al.]{yang2025qwen3}
An~Yang, Anfeng Li, Baosong Yang, Beichen Zhang, Binyuan Hui, Bo~Zheng, Bowen
  Yu, Chang Gao, Chengen Huang, Chenxu Lv, et~al.
\newblock {Qwen3} technical report.
\newblock \emph{arXiv preprint arXiv:2505.09388}, 2025.

\bibitem[Yao et~al.(2025)Yao, Yang, and Wang]{yao2025reconstruction}
Jingfeng Yao, Bin Yang, and Xinggang Wang.
\newblock Reconstruction vs. generation: Taming optimization dilemma in latent
  diffusion models.
\newblock In \emph{CVPR}, pp.\  15703--15712, 2025.

\bibitem[Yu et~al.(2022)Yu, Xu, Koh, Luong, Baid, Wang, Vasudevan, Ku, Yang,
  Karagol~Ayan, Hutchinson, Han, Parekh, Li, Zhang, Baldridge, and
  Wu]{yu2022scaling}
Jiahui Yu, Yuanzhong Xu, Jing~Yu Koh, Thang Luong, Gunjan Baid, Zirui Wang,
  Vijay Vasudevan, Alexander Ku, Yinfei Yang, Burcu Karagol~Ayan, Ben
  Hutchinson, Wei Han, Zarana Parekh, Xin Li, Han Zhang, Jason Baldridge, and
  Yonghui Wu.
\newblock Scaling autoregressive models for content-rich text-to-image
  generation.
\newblock \emph{TMLR}, 2022.

\bibitem[Yu et~al.(2024{\natexlab{a}})Yu, Lezama, Gundavarapu, Versari, Sohn,
  Minnen, Cheng, Gupta, Gu, Hauptmann, et~al.]{yu2024language}
Lijun Yu, Jos{\'e} Lezama, Nitesh~Bharadwaj Gundavarapu, Luca Versari, Kihyuk
  Sohn, David Minnen, Yong Cheng, Agrim Gupta, Xiuye Gu, Alexander~G Hauptmann,
  et~al.
\newblock Language model beats diffusion -- tokenizer is key to visual
  generation.
\newblock In \emph{ICLR}, 2024{\natexlab{a}}.

\bibitem[Yu et~al.(2024{\natexlab{b}})Yu, Weber, Deng, Shen, Cremers, and
  Chen]{yu2024image}
Qihang Yu, Mark Weber, Xueqing Deng, Xiaohui Shen, Daniel Cremers, and
  Liang-Chieh Chen.
\newblock An image is worth 32 tokens for reconstruction and generation.
\newblock In \emph{NeurIPS}, volume~37, pp.\  128940--128966,
  2024{\natexlab{b}}.

\bibitem[Yu et~al.(2025)Yu, Kwak, Jang, Jeong, Huang, Shin, and
  Xie]{yu2024representation}
Sihyun Yu, Sangkyung Kwak, Huiwon Jang, Jongheon Jeong, Jonathan Huang, Jinwoo
  Shin, and Saining Xie.
\newblock Representation alignment for generation: Training diffusion
  transformers is easier than you think.
\newblock In \emph{ICLR}, pp.\  29100--29142, 2025.

\bibitem[Yu et~al.(2026)Yu, Xiong, Nie, Sheng, Liu, and Luo]{yu2025pixeldit}
Yongsheng Yu, Wei Xiong, Weili Nie, Yichen Sheng, Shiqiu Liu, and Jiebo Luo.
\newblock {PixelDiT}: Pixel diffusion transformers for image generation.
\newblock In \emph{CVPR}, pp.\  14273--14282, 2026.

\bibitem[Zhang et~al.(2018)Zhang, Isola, Efros, Shechtman, and
  Wang]{zhang2018unreasonable}
Richard Zhang, Phillip Isola, Alexei~A Efros, Eli Shechtman, and Oliver Wang.
\newblock The unreasonable effectiveness of deep features as a perceptual
  metric.
\newblock In \emph{CVPR}, pp.\  586--595, 2018.

\bibitem[Zhang et~al.(2026)Zhang, Sun, Wu, Yi, Kong, Xiao, and
  Zhang]{zhang2026ldmisae}
Zhengqiang Zhang, Lingchen Sun, Rongyuan Wu, Qiaosi Yi, Xiangtao Kong, Chaodong
  Xiao, and Lei Zhang.
\newblock {LDM-is-AE}: Latent diffusion model is an auto-encoder for end-to-end
  image generation.
\newblock In \emph{NeurIPS}, 2026.

\bibitem[Zhao et~al.(2026)Zhao, Zheng, Yang, Zhu, Leng, Gould, and
  Zheng]{zhao2026simflow}
Qinyu Zhao, Guangting Zheng, Tao Yang, Rui Zhu, Xingjian Leng, Stephen Gould,
  and Liang Zheng.
\newblock {SimFlow}: Simplified and end-to-end training of latent normalizing
  flows.
\newblock In \emph{ECCV}, pp.\  679--697, 2026.

\bibitem[Zheng et~al.(2026{\natexlab{a}})Zheng, Ma, Tong, and
  Xie]{zheng2026diffusion}
Boyang Zheng, Nanye Ma, Shengbang Tong, and Saining Xie.
\newblock Diffusion transformers with representation autoencoders.
\newblock In \emph{ICLR}, pp.\  35791--35820, 2026{\natexlab{a}}.

\bibitem[Zheng et~al.(2026{\natexlab{b}})Zheng, Zhang, Yang, Chen, Zhu, Deng,
  and Zhang]{zheng2026genfirst}
Guangting Zheng, Yiyuan Zhang, Tao Yang, Yunpeng Chen, Rui Zhu, Jiajun Deng,
  and Yanyong Zhang.
\newblock {GenFirst}: Generation before reconstruction for stable end-to-end
  latent generative modeling.
\newblock \emph{arXiv preprint arXiv:2608.29335}, 2026{\natexlab{b}}.

\bibitem[Zheng et~al.(2024)Zheng, Nie, Vahdat, and Anandkumar]{zheng2023fast}
Hongkai Zheng, Weili Nie, Arash Vahdat, and Anima Anandkumar.
\newblock Fast training of diffusion models with masked transformers.
\newblock \emph{TMLR}, 2024.

\bibitem[Zhou et~al.(2025)Zhou, Yu, Babu, Tirumala, Yasunaga, Shamis, Kahn, Ma,
  Zettlemoyer, and Levy]{zhou2025transfusion}
Chunting Zhou, Lili Yu, Arun Babu, Kushal Tirumala, Michihiro Yasunaga, Leonid
  Shamis, Jacob Kahn, Xuezhe Ma, Luke Zettlemoyer, and Omer Levy.
\newblock Transfusion: Predict the next token and diffuse images with one
  multi-modal model.
\newblock In \emph{ICLR}, 2025.

\end{thebibliography}
\bibliographystyle{iclr2027_conference}

\clearpage

\appendix
\section*{Appendix}

In this appendix, we first provide the ImageNet-1K training details and the evaluation protocols used across the paper. Then, we provide a compute analysis on model training, the full comparison with pixel-space and latent diffusion models, and additional ablation results on ImageNet-1K. Next, we detail the text-to-image training and evaluation protocol and report zero-shot results on MS-COCO, and present results and implementation details for the QM9 molecule experiment. Finally, we discuss limitations. The last appendix shows uncurated ImageNet-1K samples.

\section{ImageNet training details}
\label{app:train-details}

Table~\ref{tab:in1k-hparams} lists the training configuration shared by all
ImageNet-1K $256\times256$ models, including the baselines. We follow
UNITE~\citep{duggal2026end} except for two settings: we train for $200$ epochs
instead of $240$, and each optimizer step combines one reconstruction step with
$12$ flow-matching steps instead of $14$.

\begin{table}[h]
\centering
\caption{Training configuration for ImageNet-1K $256\times256$, shared by all
models.
}
\label{tab:in1k-hparams}
\small
\begin{tabular}{@{}ll|ll@{}}
\toprule
Hyperparameter & Value & Hyperparameter & Value \\
\midrule
Base learning rate & $1\times10^{-4}$ & Warmup epochs & $20$ \\
Global batch size & $1024$ & Total epochs & $200$ \\
Optimizer & Muon & LR schedule & Cosine \\
AdamW betas & $(0.9, 0.999)$ & Min LR & $1\times10^{-6}$ \\
Weight decay & $0$ & Gradient clip & $3.0$ \\
Reconstruction noise ($\sigma_{\mathrm{rec}}$) & $0.7$ & EMA decay & $0.9978$ \\
Flow timesteps (training) & $1000$ & Noise schedule shift ($\alpha$) & $0.5$ \\
Flow steps per recon.\ step & $12$ & Precision & BF16 (mixed) \\
Loss weights ($\lambda_{\mathrm{rec}},\lambda_{\mathrm{perc}},\lambda_{\mathrm{gen}}$) & $(1, 1, 1)$ & Class-label dropout & $0.1$ \\
\bottomrule
\end{tabular}
\end{table}

\noindent\textbf{Architecture details.}
Table~\ref{tab:arch} lists the UNIFIED configurations. All sizes use a patch
size of $16$, so a $256\times256$ image gives $N=256$ patches and, with $K=N$,
a $16\times16$ latent with $d=32$ channels per token, as in UNITE. UNIFIED-T,
-S, and -B are the ImageNet-1K models of Table~\ref{tab:in1k}, and UNIFIED-L is
the larger model used for text-to-image generation
(Table~\ref{tab:t2i-align}).

\begin{table}[h]
\centering
\caption{Architecture configurations of UNIFIED. Parameters are for the
class-conditional models; the text-to-image models replace the class embedding
with a text-conditioning head ($218$M for UNIFIED-B and $554$M for UNIFIED-L).}
\label{tab:arch}
\small
\begin{tabular}{@{}lcccc@{}}
\toprule
 & UNIFIED-T & UNIFIED-S & UNIFIED-B & UNIFIED-L \\
\midrule
Hidden dimension   & 528 & 720 & 984 & 1120 \\
Layers             & 12 & 12 & 12 & 24 \\
Attention heads    & 12 & 12 & 12 & 16 \\
MLP ratio          & 4 & 4 & 4 & 4 \\
Patch size         & 16 & 16 & 16 & 16 \\
Latent dimension   & 32 & 32 & 32 & 32 \\
Latent resolution  & $16\times16$ & $16\times16$ & $16\times16$ & $16\times16$ \\
Parameters (M)     & 62 & 116 & 215 & 550 \\
\bottomrule
\end{tabular}
\end{table}

\section{ImageNet evaluation protocols}
\label{app:eval-protocols}

\noindent\textbf{FID computation.}
We report Fréchet Inception Distance (FID) computed with the \texttt{torch-fidelity} library using InceptionV3 features. Reference statistics are computed over the full ImageNet-1K training set (1,281,167 images), and all FID scores are evaluated on 50K generated samples. Reconstruction metrics (rFID, rIS, PSNR, LPIPS) use the $50$K ImageNet-1K validation images, center-cropped to $256\times256$ and passed through one encoding and one decoding pass with EMA weights.

\noindent\textbf{Sampling protocol.}
Following UNITE~\citep{duggal2026end}, we use class-balanced sampling, generating exactly 50 images for each of the 1,000 ImageNet classes (50K images in total).

\noindent\textbf{ODE solver.}
Every sample starts from Gaussian noise $z_0\sim\mathcal{N}(0,I)$ at $t=0$, and we integrate the probability flow ODE over the full interval $t\in[0,1]$. The main solver is the adaptive fifth-order Dormand–Prince solver (\texttt{dopri5}; absolute tolerance $10^{-6}$, relative tolerance $10^{-3}$), which chooses its own steps, so no time grid is involved. The velocity is obtained from the predicted clean latent as $(\hat z-z_t)/(1-t)$, with the denominator clamped at $0.05$.

\noindent\textbf{Classifier-free guidance.}
Following UNITE~\citep{duggal2026end}, classifier-free guidance is applied only for $t\in[0.1,1.0]$; for $t<0.1$ the model is run without guidance. This interval restricts the guidance, not the integration, which always covers $[0,1]$. We sweep the CFG scale $\omega$ from $1.0$ to $4.0$ in increments of $0.2$ and report the best FID for each model.

\noindent\textbf{Protocol of each result.}
The protocol above produces every ImageNet-1K result in the paper (Tables~\ref{tab:in1k}, \ref{tab:timestep}, \ref{tab:matched-flops}, \ref{tab:sota}, \ref{tab:allocation}, \ref{tab:timestep-full}, \ref{tab:in1k-full}, and the dopri5 columns of Table~\ref{tab:seeds}, and the training-budget panels of Figures~\ref{fig:budget} and~\ref{fig:train-budget-st}), with two exceptions. (i)~The early-termination experiment (Appendix~\ref{app:ode-end}, Table~\ref{tab:ode-end}) keeps \texttt{dopri5} and the initial condition but ends the integration at $t_{\mathrm{end}}<1$, at a fixed CFG scale of $2.0$. (ii)~The sampling-budget panels of Figure~\ref{fig:budget} and the Euler columns of Table~\ref{tab:seeds} replace \texttt{dopri5} with a fixed-step Euler solver over $[0,1]$. With a budget of $K$ NFE, it takes $K$ steps on the grid $t_i=\alpha s_i/\bigl(1+(\alpha-1)s_i\bigr)$, where $s_i=i/K$ for $i=0,\dots,K$ and $\alpha=0.5$ is the noise-schedule shift used in training (Table~\ref{tab:in1k-hparams}); the grid is therefore denser near $t=0$. Each model uses the best CFG scale found with \texttt{dopri5}, and the guidance interval and the $50$K class-balanced samples are unchanged.

\phantomsection
\label{app:cka}
\noindent\textbf{CKA analysis.} We compare the backbone's activations in the
encoding and generation passes with linear CKA~\citep{kornblith2019similarity},
using the debiased estimator~\citep{song2012feature}.
For each validation
image, we run the encoding pass on the image and the generation pass on its
layer-normalized latent noised to level $t$ ($t=1$ is clean), both with the null
class label; the noise is drawn once and reused at every $t$, so the curves
differ only in $t$. We record each block's output and keep only the positions
that carry the latent: all $N$ positions for UNIFIED, and the readout positions
for UNITE and UNITE++. For Separate, the two passes run in different networks,
the encoder and the generator. CKA is computed within each image, with its tokens
as samples, so it measures whether the two passes induce the same token-to-token
similarity structure.

\section{Compute analysis}
\label{app:compute}

UNIFIED reserves no parameters for any single mode, so training can assign the
whole backbone to whichever task needs it, rather than fixing the split in
advance as UNITE's encoder-to-decoder ratio does (Table~\ref{tab:allocation}).
We attribute its gFID gains at matched size to this (Table~\ref{tab:in1k};
Appendix~\ref{app:parameter-allocation}). The cost is compute: every mode runs
the full backbone, whereas UNITE denoises with its generative encoder alone.

\noindent\textbf{Per step.} At matched parameters ($215$M),
the shared backbone is wider ($984$ vs.\ $768$), so one flow-matching forward
pass for UNIFIED costs $1.63\times$ more than UNITE ($81.7$ vs.\ $50.1$ GFLOPs).

\noindent\textbf{Per training step.} The reconstruction branch costs about the
same in both models ($0.62$ vs.\ $0.57$ TFLOPs): UNIFIED encodes and decodes
position-aligned $257$-token sequences, while UNITE encodes $512$ tokens of
readouts and patches and decodes with a separate decoder. Flow matching
dominates the step ($83\%$ and $76\%$ of compute), so a full step for UNIFIED costs
$1.50\times$ more than UNITE ($3.58$ vs.\ $2.39$ TFLOPs per image).

\noindent\textbf{At matched training compute.} Training for the same
number of epochs therefore gives UNIFIED $1.35\times$, $1.72\times$, and
$1.50\times$ the training compute of UNITE at Tiny, Small, and Base.
Table~\ref{tab:matched-flops} instead compares the two at matched total training
FLOPs, using the UNIFIED checkpoints whose compute brackets UNITE's $200$-epoch
budget (the full curves are in Figures~\ref{fig:budget}
and~\ref{fig:train-budget-st}). With $10$--$19\%$ less compute, UNIFIED is within $0.03$ gFID of UNITE's final
model at every size ($3.03$ vs.\ $3.00$ at Tiny, $2.78$ vs.\ $2.76$ at Small,
and $2.22$ vs.\ $2.23$ at Base); with $3$--$20\%$ more, it improves on UNITE at
every size ($2.91$, $2.65$, and $2.15$). At UNITE's full training budget, UNIFIED is
thus on par with or better than UNITE. The gain is, however, smaller than at
matched epochs ($0.15$ at Base in Table~\ref{tab:in1k}), and UNITE is ahead
early in training (Figure~\ref{fig:budget}). 
However, UNITE's curve flattens over its last 240 EFLOPs ($2.26 \to 2.23 \to 2.23$), while UNIFIED keeps improving up to and beyond UNITE's total budget ($2.22 \to 2.15 \to 2.08$).

\begin{table}[t]
\centering
\caption{\textbf{UNITE and UNIFIED at matched training compute} on ImageNet-1K $256\times256$, with CFG. Training EFLOPs: total training FLOPs up to the checkpoint, in units of $10^{18}$. UNITE: final ($200$-epoch) models of Table~\ref{tab:in1k}. UNIFIED ($t_{\mathrm{dec}}{=}{-1}$): the two checkpoints whose training compute brackets UNITE's. gFID: mean over multiple evaluation seeds.}
\label{tab:matched-flops}
\small
\setlength{\tabcolsep}{6pt}
\begin{tabular}{@{}llccc@{}}
\toprule
Size & Method & Epochs & Training EFLOPs & gFID $\downarrow$ \\
\midrule
\multirow{3}{*}{Tiny}
 & UNITE ($2{:}1$) & 200 & 218 & 3.00 \\
 & UNIFIED & 120 & 176 & 3.03 \\
 & UNIFIED & 160 & 235 & 2.91 \\
\midrule
\multirow{3}{*}{Small}
 & UNITE & 200 & 296 & 2.76 \\
 & UNIFIED & 100 & 254 & 2.78 \\
 & UNIFIED & 120 & 305 & 2.65 \\
\midrule
\multirow{3}{*}{Base}
 & UNITE & 200 & 608 & 2.23 \\
 & UNIFIED & 120 & 546 & 2.22 \\
 & UNIFIED & 160 & 728 & 2.15 \\
\bottomrule
\end{tabular}
\end{table}

\noindent\textbf{Per sample.} A higher cost per step need not mean a higher
cost at equal quality: UNIFIED at $64$ NFE reaches a lower gFID than UNITE at
$128$ NFE ($2.23$ vs.\ $2.35$; Section~\ref{sec:nfe}) with $18\%$ fewer
denoising FLOPs ($5.2$ vs.\ $6.4$ TFLOPs per sample).

\section{Comparison with pixel-space and latent diffusion models}
\label{app:sota}

Table~\ref{tab:sota} extends Section~\ref{sec:sota} and places UNIFIED among published class-conditional generators on ImageNet-1K $256\times256$, grouped by parameter count.

\noindent\textbf{Baselines.} Two-stage baselines run diffusion in the frozen latent of a pretrained autoencoder: DiT~\citep{peebles2023scalable}, MaskDiT~\citep{zheng2023fast}, SiT~\citep{ma2024sit}, and MDTv2~\citep{gao2023masked} use only the autoencoder, whereas REPA~\citep{yu2024representation}, REPA-E~\citep{leng2025repa}, LightningDiT~\citep{yao2025reconstruction}, DDT~\citep{wang2025ddt}, RAE~\citep{zheng2026diffusion}, SVG~\citep{shi2026latent}, and EiT~\citep{zheng2026genfirst} additionally rely on a pretrained vision foundation model. Single-stage baselines train without a pretrained autoencoder: pixel-space models ADM~\citep{dhariwal2021diffusion}, RIN~\citep{jabri2022scalable}, SiD~\citep{hoogeboom2023simple}, VDM++~\citep{kingma2023understanding}, SiD2~\citep{hoogeboom2025simpler}, PixelFlow~\citep{chen2025pixelflow}, JiT~\citep{li2026back} and its register-guided variant~\citep{starodubcev2026registers}, DeCo~\citep{ma2026deco}, PixelGen~\citep{ma2026pixelgen}, PixNerd~\citep{wang2025pixnerd}, PixelDiT~\citep{yu2025pixeldit}, PixelREPA~\citep{shin2026representation}, PixelU~\citep{guo2026pixelu}, and Latent Forcing (LF-DiT)~\citep{baade2026latent}, and end-to-end latent models DSD~\citep{wang2025diffusion}, UNITE~\citep{duggal2026end}, and LDM-is-AE~\citep{zhang2026ldmisae}.

\noindent\textbf{Comparability.} Results of other methods are taken from their papers and therefore differ in training length, sampler, and guidance settings. Our models are trained for $200$ epochs, fewer than most baselines ($600$ for JiT and PixelREPA, and $400$--$1600$ for the two-stage models). UNITE$^*$ is retrained under the same protocol as UNIFIED. Given our computational budget, we train UNIFIED up to ${\sim}215$M parameters at $256\times256$; the $250$M--$1$B and ${\ge}1$B groups are included for reference, and we leave larger models and higher-resolution ImageNet to future work.

\noindent\textbf{Results.} Among single-stage models of at most ${\sim}220$M parameters, UNIFIED attains the lowest gFID at every size: UNIFIED-T ($62$M, $2.78$) already improves on all pixel-space baselines of $131$--$164$M ($2.96$--$3.66$). UNIFIED-B ($215$M, $2.08$) also improves on the pixel-space JiT-L/16 ($459$M, $2.36$), which has about twice its parameters and is trained for three times as many epochs. UNIFIED-B further improves on two-stage REPA-B ($2.15$, $214$M) and matches RAE-B ($2.08$, $631$M) without a pretrained autoencoder or vision foundation model. Every model with a lower gFID than UNIFIED-B uses at least $545$M parameters.

\begin{table}[tp]
\centering
\caption{\textbf{Class-conditional generation on ImageNet-1K $256\times256$}, comparing pixel-space and latent diffusion models. Models: networks the method uses, namely a generator (Gen.), a pretrained autoencoder (AE), a pretrained vision foundation model (VFM), or a standalone decoder trained jointly with the generator (Dec.). No Aux.\ Info.: \cmark{} if training uses no auxiliary pretrained representation (e.g., DINOv2 features as the latent or as a representation-alignment target). Repr.: the space in which diffusion runs, namely a frozen pretrained latent (Fixed), pixels (Pixel), or a latent learned jointly with the generator (Dynamic). Epoch: training epochs. Params counts all networks used. Results of other methods are taken from their papers. $^*$: UNITE trained under our protocol. UNITE$^*$ and UNIFIED: mean over five evaluation seeds (Table~\ref{tab:seeds}). --: not reported.}
\label{tab:sota}
\small
\setlength{\tabcolsep}{3.8pt}
\begin{tabular}{@{}clccccccc@{}}
\toprule
Scale & Method & Models & \begin{tabular}[c]{@{}c@{}}No Aux.\\Info.\end{tabular} & Repr. & Epoch & Params & gFID $\downarrow$ & IS $\uparrow$ \\
\midrule
\rowcolor{gray!15}
\multicolumn{9}{@{}l}{\textit{Two-stage}} \\
150--250M & REPA-B                          & Gen.+AE+VFM &        & Fixed   & --    & 214M  & 2.15  & 268.3 \\
\dashedrule
 & MaskDiT                         & Gen.+AE     & \cmark & Fixed   & 1600  & 759M  & 2.28  & 276.6 \\
 & DiT-XL/2                        & Gen.+AE     & \cmark & Fixed   & 1400  & 759M  & 2.27  & 278.2 \\
 & RAE-B                           & Gen.+AE+VFM &        & Fixed   & --    & 631M  & 2.08  & 275.1 \\
 & SiT-XL/2                        & Gen.+AE     & \cmark & Fixed   & 1400  & 759M  & 2.06  & 270.3 \\
 & SVG-XL                          & Gen.+AE+VFM &        & Fixed   & 1400  & 758M  & 1.92  & 265.0 \\
 & MDTv2                           & Gen.+AE     & \cmark & Fixed   & 1080  & 759M  & 1.58  & 314.7 \\
 & LightningDiT-XL                 & Gen.+AE+VFM &        & Fixed   & 800   & 745M  & 1.35  & 295.3 \\
 & REPA-SiT-XL/2                   & Gen.+AE+VFM &        & Fixed   & 800   & 759M  & 1.29  & 306.3 \\
 & DDT-XL/2                        & Gen.+AE+VFM &        & Fixed   & 400   & 759M  & 1.26  & 310.6 \\
\multirow{-10}{*}{250M--1B}
 & REPA-E (tuning)                 & Gen.+AE+VFM &        & Fixed   & 800   & 745M  & 1.12  & 302.9 \\
\dashedrule
 & RAE-DiT$^{\mathrm{DH}}$-XL/2    & Gen.+AE+VFM &        & Fixed   & 800   & 1.3B  & 1.13  & 262.6 \\
\multirow{-2}{*}{${\ge}$1B}
 & EiT                             & Gen.+AE+VFM &        & Fixed   & 800   & 1.4B  & 0.97 & 283.4 \\
\midrule
\rowcolor{gray!15}
\multicolumn{9}{@{}l}{\textit{Single-stage}} \\
\rowcolor{unitebg}
\nocol & UNITE-T$^*$                     & Gen.+Dec.   & \cmark & Dynamic & 200   & 61M   & 3.00  & 272.4 \\
\rowcolor{unifiedbg}
\nocol\multirow{-2}{*}{${\le}$100M}
 & \textbf{UNIFIED-T (Ours)} & \textbf{Gen.} & \textbf{\cmark} & \textbf{Dynamic} & \textbf{200} & \textbf{62M} & \textbf{2.78} & \textbf{280.1} \\
\dashedrule
 & JiT-B/16                        & Gen.        & \cmark & Pixel   & 600   & 131M  & 3.66  & 275.1 \\
 & PixelREPA-B/16                  & Gen.+VFM    &        & Pixel   & 600   & 131M  & 3.17  & 284.6 \\
 & JiT-B/16 + RG                   & Gen.        & \cmark & Pixel   & 600   & 131M  & 2.96  & 251.1 \\
\rowcolor{unitebg}
\nocol & UNITE-S$^*$                     & Gen.+Dec.   & \cmark & Dynamic & 200   & 117M  & 2.76  & 283.9 \\
\rowcolor{unifiedbg}
\nocol\multirow{-5}{*}{100--150M}
 & \textbf{UNIFIED-S (Ours)} & \textbf{Gen.} & \textbf{\cmark} & \textbf{Dynamic} & \textbf{200} & \textbf{116M} & \textbf{2.43} & \textbf{295.9} \\
\dashedrule
 & PixelU-B/16                     & Gen.+VFM    &        & Pixel   & 600   & 164M  & 3.40  & 276.0 \\
 & DSD                             & Gen.+VFM    &        & Dynamic & 50    & 205M  & 3.35  & 255.0 \\
\rowcolor{unitebg}
\nocol & UNITE-B$^*$                     & Gen.+Dec.   & \cmark & Dynamic & 200   & 217M  & 2.23  & 282.1 \\
\rowcolor{unifiedbg}
\nocol\multirow{-4}{*}{150--250M}
 & \textbf{UNIFIED-B (Ours)} & \textbf{Gen.} & \textbf{\cmark} & \textbf{Dynamic} & \textbf{200} & \textbf{215M} & \textbf{2.08} & \textbf{314.4} \\
\dashedrule
 & ADM-G                           & Gen.        & \cmark & Pixel   & 400   & 554M  & 4.59  & 186.7 \\
 & RIN                             & Gen.        & \cmark & Pixel   & 480   & 410M  & 3.42  & 182.0 \\
 & LF-DiT-L                        & Gen.+VFM    &        & Pixel   & 200   & 465M  & 2.48  & --    \\
 & JiT-L/16                        & Gen.        & \cmark & Pixel   & 600   & 459M  & 2.36  & 298.5 \\
 & PixNerd-XL/16                   & Gen.+VFM    &        & Pixel   & 160   & 700M  & 2.15  & 297.0 \\
 & PixelFlow-XL/4                  & Gen.        & \cmark & Pixel   & 320   & 677M  & 1.98  & 282.1 \\
 & JiT-H/16                        & Gen.        & \cmark & Pixel   & 600   & 953M  & 1.86  & 303.4 \\
 & PixelGen-XL/16                  & Gen.+VFM    &        & Pixel   & 160   & 676M  & 1.83  & 293.6 \\
 & LDM-is-AE                       & Gen.        & \cmark & Dynamic & 300   & 961M  & 1.80  & 314.0 \\
 & UNITE-XL                        & Gen.+Dec.   & \cmark & Dynamic & 240   & 762M  & 1.75  & 309.9 \\
 & UNITE-L                         & Gen.+Dec.   & \cmark & Dynamic & 240   & 545M  & 1.73  & 296.0 \\
 & DeCo                            & Gen.+VFM    &        & Pixel   & 600   & 682M  & 1.69  & 304.0 \\
\multirow{-13}{*}{250M--1B}
 & PixelDiT                        & Gen.+VFM    &        & Pixel   & 320   & 797M  & 1.61  & 292.7 \\
\dashedrule
 & SiD                             & Gen.        & \cmark & Pixel   & 800   & 2B    & 2.44  & 256.3 \\
 & VDM++                           & Gen.        & \cmark & Pixel   & --    & 2B    & 2.12  & 267.7 \\
\multirow{-3}{*}{${\ge}$1B}
 & JiT-G/16                        & Gen.        & \cmark & Pixel   & 600   & 2B    & 1.82  & 292.6 \\
\bottomrule
\end{tabular}
\end{table}

\section{Additional ablation experiments}

\phantomsection
\label{app:parameter-allocation}
\noindent\textbf{Decoder parameters are better spent in the shared backbone.}
At a fixed Tiny budget (Table~\ref{tab:allocation}), moving UNITE's capacity from
the decoder to the encoder/generator improves gFID ($1{:}2\to1{:}1\to2{:}1$:
$5.01\to3.79\to3.00$), even though the decoder-heavy $1{:}2$ reconstructs best.
Removing the decoder altogether (UNIFIED) improves gFID further, to $2.78$
($2.75$ with $t_{\mathrm{dec}}{=}1$). Parameters thus help generation most when
they sit in the shared backbone rather than in a standalone decoder.

\begin{table}[t]
\centering
\caption{\textbf{Encoder/decoder allocation at the Tiny level} (${\sim}60$M) on ImageNet-1K $256\times256$, epoch $200$, ODE over $[0,1]$; with CFG, each model uses its best scale. Generation with CFG: mean over five evaluation seeds (Table~\ref{tab:seeds}); $^\dagger$: single run. UNITE ($a{:}b$): ratio of encoder/generator to decoder parameters ($60$--$61$M); UNIFIED ($62$M) has no standalone decoder.}
\label{tab:allocation}
\small
\setlength{\tabcolsep}{3.5pt}
\begin{tabular}{@{}lcccccccc@{}}
\toprule
 & \multicolumn{2}{c}{Gen.\ w/o CFG} & \multicolumn{2}{c}{Gen.\ w/ CFG} & \multicolumn{4}{c}{Reconstruction} \\
\cmidrule(lr){2-3}\cmidrule(lr){4-5}\cmidrule(lr){6-9}
Method & gFID $\downarrow$ & IS $\uparrow$ & gFID $\downarrow$ & IS $\uparrow$ & rFID $\downarrow$ & rIS $\uparrow$ & PSNR $\uparrow$ & LPIPS $\downarrow$ \\
\midrule
UNITE ($1{:}2$)$^\dagger$ & 41.19 & 37.9 & 5.01 & 239.6 & \textbf{1.24} & 196.4 & \textbf{26.60} & \textbf{0.115} \\
UNITE ($1{:}1$)$^\dagger$ & 31.18 & 49.8 & 3.79 & 262.5 & 1.59 & 190.4 & 26.14 & 0.127 \\
UNITE ($2{:}1$) & 26.02 & 60.8 & 3.00 & 272.4 & 1.27 & \textbf{197.2} & 26.31 & \textbf{0.115} \\
\midrule
UNIFIED ($t_{\mathrm{dec}}{=}1$) & \textbf{20.71} & \textbf{75.2} & \textbf{2.75} & \textbf{282.0} & 1.30 & 196.7 & 25.65 & 0.127 \\
UNIFIED ($t_{\mathrm{dec}}{=}{-1}$) & 21.25 & 72.3 & 2.78 & 280.1 & 1.75 & 188.6 & 25.23 & 0.142 \\
\bottomrule
\end{tabular}
\end{table}

\noindent\textbf{Full results.}
Tables~\ref{tab:allocation}, \ref{tab:timestep-full}, and~\ref{tab:in1k-full}
report every ImageNet-1K model at epoch $200$, without and with CFG. 
Both UNIFIED variants improve on every UNITE allocation, as shown in Table~\ref{tab:in1k-full}.
At Base, both UNIFIED variants reconstruct better than
UNITE (lower rFID, higher rIS) at a similar PSNR and LPIPS, while UNITE++
reconstructs worst. 
Moreover, encoding at $t_{\mathrm{enc}}{=}0$ instead
of $1$ hurts UNIFIED on both tasks (gFID $2.31$ vs.\ $2.15$, rFID $1.39$ vs.\
$0.97$), as shown in Table~\ref{tab:timestep-full}. The conclusions hold
with and without guidance, and full unification costs
reconstruction only at the smallest size.

\begin{table}[t]
\centering
\caption{\textbf{Timestep conditioning at the Base level} (${\sim}215$M): full results of Table~\ref{tab:timestep}, epoch $200$, ODE over $[0,1]$; with CFG, each model uses its best scale. Generation with CFG: mean over five evaluation seeds (Table~\ref{tab:seeds}). $t_{\mathrm{enc}}$, $t_{\mathrm{dec}}$: timesteps of the encoding and decoding passes; $t_{\mathrm{dec}}{=}{-1}$: learned decoding timestep; --: not applicable (UNITE). $^\dagger$: single run.}
\label{tab:timestep-full}
\small
\setlength{\tabcolsep}{3.2pt}
\begin{tabular}{@{}lcccccccccc@{}}
\toprule
 &  &  & \multicolumn{2}{c}{Gen.\ w/o CFG} & \multicolumn{2}{c}{Gen.\ w/ CFG} & \multicolumn{4}{c}{Reconstruction} \\
\cmidrule(lr){4-5}\cmidrule(lr){6-7}\cmidrule(lr){8-11}
Method & $t_{\mathrm{enc}}$ & $t_{\mathrm{dec}}$ & gFID $\downarrow$ & IS $\uparrow$ & gFID $\downarrow$ & IS $\uparrow$ & rFID $\downarrow$ & rIS $\uparrow$ & PSNR $\uparrow$ & LPIPS $\downarrow$ \\
\midrule
UNITE & $0$ & -- & 15.99 & 91.4 & 2.23 & 282.1 & 1.41 & 198.0 & 26.63 & 0.106 \\
UNITE++ & $0$ & $1$ & 16.16 & 95.7 & 2.24 & \textbf{327.1} & 2.20 & 182.0 & 24.99 & 0.145 \\
\midrule
UNIFIED$^\dagger$ & $0$ & $1$ & 14.95 & 100.3 & 2.31 & 300.0 & 1.39 & 194.5 & 25.74 & 0.126 \\
UNIFIED & $1$ & $1$ & 14.20 & 100.4 & 2.15 & 312.8 & 0.97 & 203.9 & 26.63 & 0.103 \\
UNIFIED & $1$ & $-1$ & \textbf{14.12} & \textbf{103.6} & \textbf{2.08} & 314.4 & \textbf{0.95} & \textbf{205.2} & \textbf{26.69} & \textbf{0.101} \\
\bottomrule
\end{tabular}
\end{table}

\begin{table}[t]
\centering
\caption{\textbf{Separate, UNITE, and UNIFIED at matched sizes}: full results of Table~\ref{tab:in1k}, epoch $200$, ODE over $[0,1]$; with CFG, each model uses its best scale. Generation with CFG: mean over five evaluation seeds (Table~\ref{tab:seeds}); $^\dagger$: single run. UNITE uses its best allocation at each size ($2{:}1$ at Tiny, Table~\ref{tab:allocation}). Parameters: Tiny $61$--$62$M, Small $113$--$117$M, Base $215$--$222$M.}
\label{tab:in1k-full}
\small
\setlength{\tabcolsep}{3.2pt}
\begin{tabular}{@{}llcccccccc@{}}
\toprule
 &  & \multicolumn{2}{c}{Gen.\ w/o CFG} & \multicolumn{2}{c}{Gen.\ w/ CFG} & \multicolumn{4}{c}{Reconstruction} \\
\cmidrule(lr){3-4}\cmidrule(lr){5-6}\cmidrule(lr){7-10}
Size & Method & gFID $\downarrow$ & IS $\uparrow$ & gFID $\downarrow$ & IS $\uparrow$ & rFID $\downarrow$ & rIS $\uparrow$ & PSNR $\uparrow$ & LPIPS $\downarrow$ \\
\midrule
\multirow{3}{*}{Tiny}
 & UNITE ($2{:}1$) & 26.02 & 60.8 & 3.00 & 272.4 & \textbf{1.27} & \textbf{197.2} & \textbf{26.31} & \textbf{0.115} \\
 & UNIFIED ($t_{\mathrm{dec}}{=}1$) & \textbf{20.71} & \textbf{75.2} & \textbf{2.75} & \textbf{282.0} & 1.30 & 196.7 & 25.65 & 0.127 \\
 & UNIFIED ($t_{\mathrm{dec}}{=}{-1}$) & 21.25 & 72.3 & 2.78 & 280.1 & 1.75 & 188.6 & 25.23 & 0.142 \\
\midrule
\multirow{4}{*}{Small}
 & Separate$^\dagger$ & 42.54 & 36.9 & 5.17 & 244.8 & 1.64 & 200.1 & \textbf{27.74} & \textbf{0.085} \\
 & UNITE & 22.80 & 69.2 & 2.76 & 283.9 & 1.13 & 200.1 & 26.76 & 0.104 \\
 & UNIFIED ($t_{\mathrm{dec}}{=}1$) & 17.61 & 86.7 & \textbf{2.41} & 287.4 & \textbf{0.95} & 204.0 & 26.43 & 0.108 \\
 & UNIFIED ($t_{\mathrm{dec}}{=}{-1}$) & \textbf{16.65} & \textbf{90.7} & 2.43 & \textbf{295.9} & 1.04 & \textbf{206.9} & 26.65 & 0.105 \\
\midrule
\multirow{4}{*}{Base}
 & Separate$^\dagger$ & 29.74 & 52.4 & 3.96 & 267.7 & 1.38 & 204.9 & \textbf{28.13} & \textbf{0.077} \\
 & UNITE & 15.99 & 91.4 & 2.23 & 282.1 & 1.41 & 198.0 & 26.63 & 0.106 \\
 & UNIFIED ($t_{\mathrm{dec}}{=}1$) & 14.20 & 100.4 & 2.15 & 312.8 & 0.97 & 203.9 & 26.63 & 0.103 \\
 & UNIFIED ($t_{\mathrm{dec}}{=}{-1}$) & \textbf{14.12} & \textbf{103.6} & \textbf{2.08} & \textbf{314.4} & \textbf{0.95} & \textbf{205.2} & 26.69 & 0.101 \\
\bottomrule
\end{tabular}
\end{table}

\noindent\textbf{Variation across evaluation seeds.}
Every epoch-$200$ gFID and IS with CFG that we report for UNITE, UNITE++, and
UNIFIED is the mean over repeated evaluations, and Table~\ref{tab:seeds} gives
these means with their standard deviations. Each seed draws a different set of
$50$K initial noises; the checkpoint, class-balanced labels, and CFG scale are
fixed. Every model is evaluated with five seeds.
At Base, we repeat the evaluation with the adaptive dopri5
solver and with a $256$-step Euler solver (Appendix~\ref{app:eval-protocols}).
With dopri5, the standard deviation of gFID is at most $0.025$, and the three ranges do not
overlap: all five runs of UNIFIED ($t_{\mathrm{dec}}{=}{-1}$) ($2.047$--$2.108$)
are below all five of UNIFIED ($t_{\mathrm{dec}}{=}1$) ($2.134$--$2.162$), which
are below all five of UNITE ($2.206$--$2.255$). On average, UNIFIED improves on
UNITE by $0.15$ gFID, about six standard deviations. With the Euler solver, both
UNIFIED variants again improve on UNITE in every run ($2.070$--$2.133$ vs.\
$2.253$--$2.318$), but their own order reverses ($2.089$ for
$t_{\mathrm{dec}}{=}1$ vs.\ $2.121$ for $t_{\mathrm{dec}}{=}{-1}$), so which
variant generates better depends on the solver. The gains at Tiny and Small
($0.22$ and $0.33$ gFID) are at least seven times the largest standard deviation
at that size.
Because one Base run costs $6$--$9\times10^{20}$ training FLOPs
(Appendix~\ref{app:compute}), we do not repeat training with different seeds, so
the table measures the evaluation variance only.

\begin{table}[h]
\centering
\caption{\textbf{Variation across evaluation seeds} (epoch $200$, with CFG, ODE over $[0,1]$, $50$K samples). Mean $\pm$ standard deviation over five evaluation seeds. Two solvers are tested: the adaptive dopri5 solver and a $256$-step Euler solver. --: not evaluated.}
\label{tab:seeds}
\small
\setlength{\tabcolsep}{4pt}
\begin{tabular}{@{}llcccccc@{}}
\toprule
 &  &  &  & \multicolumn{2}{c}{Adaptive dopri5} & \multicolumn{2}{c}{Euler, $256$ NFE} \\
\cmidrule(lr){5-6}\cmidrule(lr){7-8}
Size & Method & $t_{\mathrm{enc}}$ & $t_{\mathrm{dec}}$ & gFID $\downarrow$ & IS $\uparrow$ & gFID $\downarrow$ & IS $\uparrow$ \\
\midrule
\multirow{3}{*}{Tiny}
 & UNITE ($2{:}1$) & $0$ & --   & $2.998\pm0.030$ & $272.4\pm1.6$ & -- & -- \\
 & UNIFIED & $1$ & $1$  & $\mathbf{2.754}\pm0.026$ & $\mathbf{282.0}\pm1.0$ & -- & -- \\
 & UNIFIED & $1$ & $-1$ & $2.783\pm0.025$ & $280.1\pm0.6$ & -- & -- \\
\midrule
\multirow{3}{*}{Small}
 & UNITE   & $0$ & --   & $2.761\pm0.020$ & $283.9\pm1.1$ & -- & -- \\
 & UNIFIED & $1$ & $1$  & $\mathbf{2.414}\pm0.018$ & $287.4\pm1.2$ & -- & -- \\
 & UNIFIED & $1$ & $-1$ & $2.426\pm0.012$ & $\mathbf{295.9}\pm2.0$ & -- & -- \\
\midrule
\multirow{4}{*}{Base}
 & UNITE   & $0$ & --   & $2.226\pm0.019$ & $282.1\pm1.4$ & $2.277\pm0.029$ & $280.4\pm0.6$ \\
 & UNITE++ & $0$ & $1$  & $2.239\pm0.004$ & $\mathbf{327.1}\pm0.7$ & $2.279\pm0.013$ & $\mathbf{329.2}\pm1.4$ \\
 & UNIFIED & $1$ & $1$  & $2.150\pm0.011$ & $312.8\pm1.9$ & $\mathbf{2.089}\pm0.013$ & $315.0\pm1.1$ \\
 & UNIFIED & $1$ & $-1$ & $\mathbf{2.075}\pm0.025$ & $314.4\pm2.3$ & $2.121\pm0.010$ & $314.9\pm1.9$ \\
\bottomrule
\end{tabular}
\end{table}

\phantomsection
\label{app:ode-end}
\noindent\textbf{Where to end the sampling ODE.}
With $t_{\mathrm{dec}}{=}1$, decoding uses $t=1$ at every step, while generation
samples $t\in[0.99,1]$ only $1\%$ of the time, so the end of the generation
trajectory is learned least. Stopping the ODE at $t=0.99$ improves the gFID of
UNIFIED ($t_{\mathrm{enc}}{=}1$, $t_{\mathrm{dec}}{=}1$) from $2.145$ to $2.040$
(Table~\ref{tab:ode-end}; Base, dopri5, the same evaluation seed with and without early stopping). The decoding pass tolerates the residual noise because it
is trained on latents perturbed by $\sigma_{\mathrm{rec}}$. UNITE ($2.206$ vs.\ $2.208$) and
UNIFIED ($t_{\mathrm{enc}}{=}1$, $t_{\mathrm{dec}}{=}{-1}$) ($2.108$ vs.\ $2.111$), whose endpoint belongs to generation alone, are
unchanged, so the gain comes from the shared endpoint rather than the solver.
Early stopping thus works around this competition at sampling time, whereas the
learned decoding timestep removes it in training and is better under standard
$[0,1]$ integration with dopri5 ($2.08$ vs.\ $2.15$, Table~\ref{tab:seeds}).

Table~\ref{tab:ode-end} sweeps $t_{\mathrm{end}}$ in a single evaluation run,
so its $[0,1]$ column differs slightly from the seed means of Table~\ref{tab:in1k}. Stopping at any
$t_{\mathrm{end}}\in[0.95,0.995]$ and decoding (b) lowers
UNIFIED ($t_{\mathrm{enc}}{=}1$, $t_{\mathrm{dec}}{=}1$) from $2.145$ to $2.032$--$2.040$, while UNITE and
UNIFIED ($t_{\mathrm{enc}}{=}1$, $t_{\mathrm{dec}}{=}{-1}$) move by at most $0.005$. Jumping to the predicted clean
latent $\hat z$ before decoding (c) brings no further gain and degrades both
UNIFIED variants as $t_{\mathrm{end}}\to1$ ($2.168$ and $2.149$ at $0.995$).
The gain therefore comes from skipping the final stretch of the trajectory, and
the exact end point within $[0.95,0.995]$ barely matters.

\begin{table}[h]
\centering
\caption{\textbf{End point of the sampling ODE.} Base models, gFID-50K
($\downarrow$) with the adaptive dopri5 solver and CFG scale $2.0$. (a)~Integrate over $[0,1]$. (b)~Stop at
$t_{\mathrm{end}}$ and decode. (c)~Stop at $t_{\mathrm{end}}$ and jump to the
predicted clean latent before decoding.}
\label{tab:ode-end}
\small
\setlength{\tabcolsep}{4pt}
\begin{tabular}{lc|cccc|cccc}
\toprule
 & (a) & \multicolumn{4}{c|}{(b) stop at $t_{\mathrm{end}}$} & \multicolumn{4}{c}{(c) stop at $t_{\mathrm{end}}$ + jump} \\
Model & $1.0$ & $0.95$ & $0.98$ & $0.99$ & $0.995$ & $0.95$ & $0.98$ & $0.99$ & $0.995$ \\
\midrule
UNITE & 2.206 & 2.206 & 2.208 & 2.208 & 2.206 & 2.210 & 2.207 & 2.199 & 2.194 \\
UNIFIED ($t_{\mathrm{dec}}{=}1$) & 2.145 & 2.032 & 2.038 & 2.040 & 2.040 & 2.037 & 2.048 & 2.052 & 2.168 \\
UNIFIED ($t_{\mathrm{dec}}{=}{-1}$) & 2.108 & 2.103 & 2.108 & 2.111 & 2.112 & 2.107 & 2.120 & 2.132 & 2.149 \\
\bottomrule
\end{tabular}
\end{table}

\begin{figure}[h]
\centering
\includegraphics[width=\linewidth]{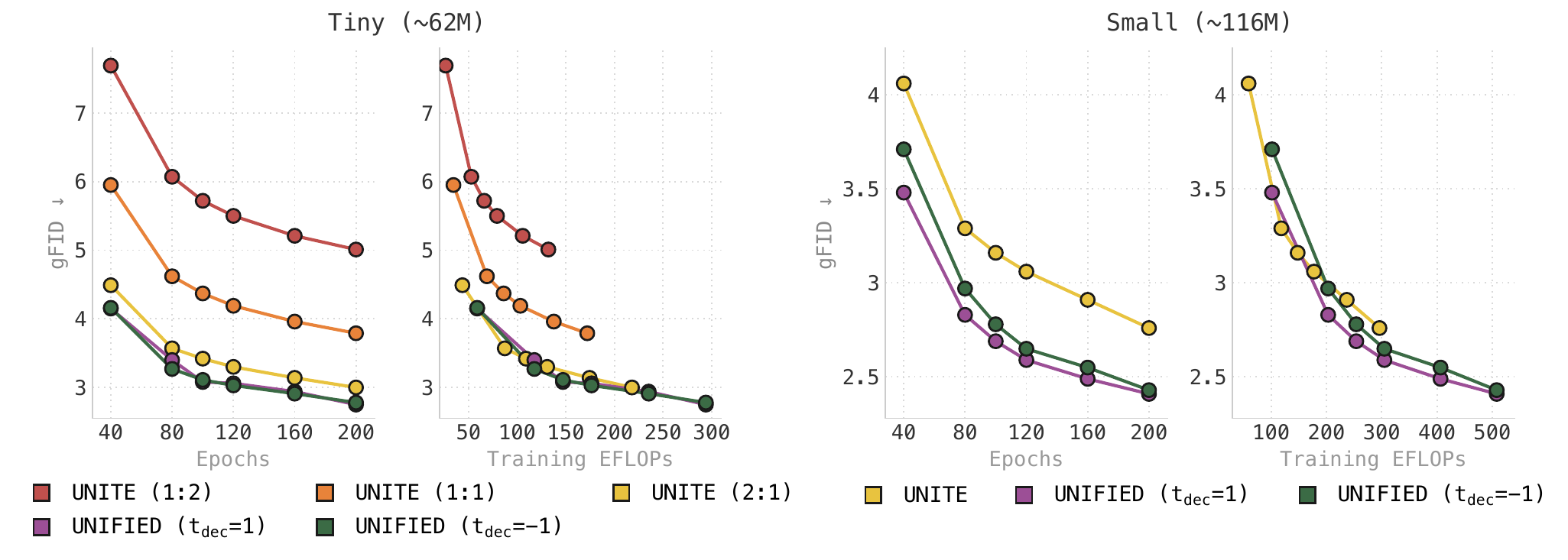}
\caption{\textbf{Generation quality across training budgets at Tiny and Small.} ImageNet-1K $256\times256$, with CFG and the adaptive dopri5 solver. For each size, gFID vs.\ training epochs and vs.\ total training FLOPs. Epoch $200$ of UNITE ($2{:}1$ at Tiny) and UNIFIED is the mean over five evaluation seeds (Table~\ref{tab:seeds}), and the UNIFIED ($t_{\mathrm{dec}}{=}{-1}$) checkpoints of Table~\ref{tab:matched-flops} are also means over multiple evaluation seeds; other points are single runs. Left: Tiny (${\sim}62$M), including UNITE at all three encoder/generator-to-decoder ratios (Table~\ref{tab:allocation}). Right: Small (${\sim}116$M).}
\label{fig:train-budget-st}
\end{figure}

\noindent\textbf{gFID across training budgets at Tiny and Small.}
Figure~\ref{fig:train-budget-st} repeats the training-budget panels of
Figure~\ref{fig:budget} at Tiny and Small parameter level. Similar to Base level, UNIFIED has a lower gFID
than UNITE at every checkpoint.

\begin{figure}[t]
\centering
\includegraphics[width=0.6\linewidth]{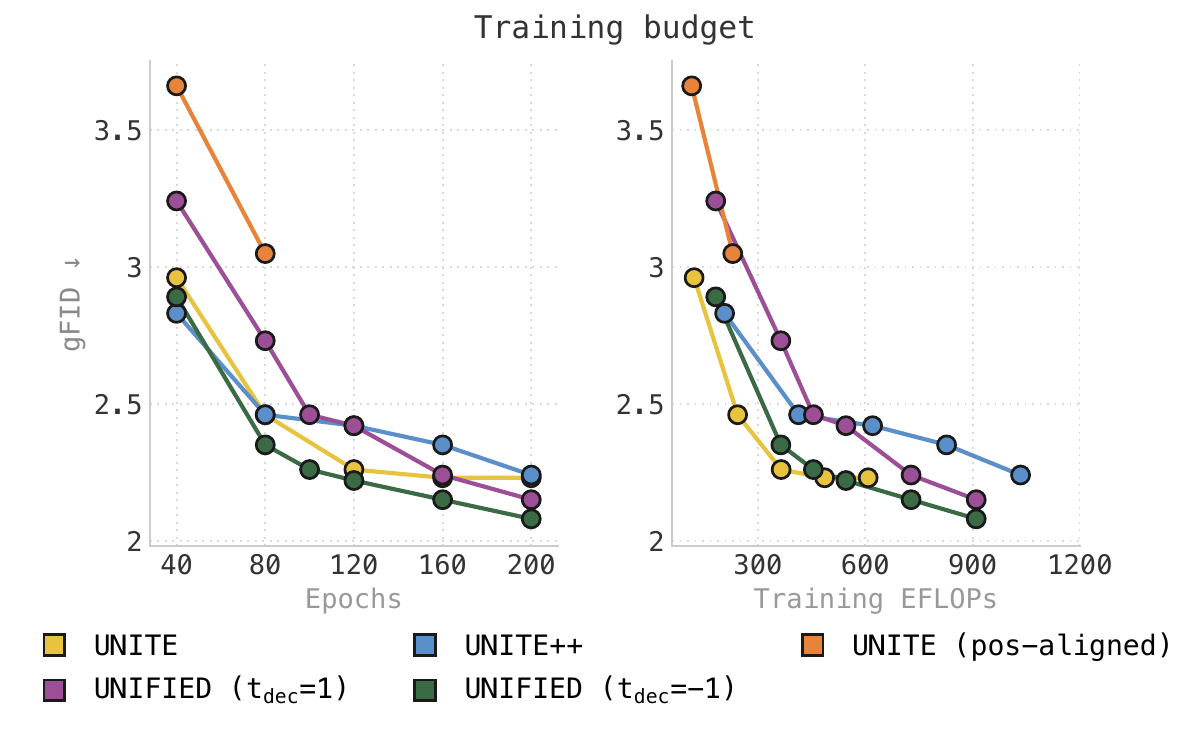}
\caption{\textbf{UNITE with position-aligned layout} (evaluated up to epoch $80$). Base models on ImageNet-1K $256\times256$, with CFG and the adaptive dopri5 solver: gFID vs.\ training epochs (left) and vs.\ total training FLOPs (right). Other models as in the training-budget panels of Figure~\ref{fig:budget}.}
\label{fig:unite-aligned}
\end{figure}

\phantomsection
\label{app:unite-aligned}
\noindent\textbf{UNITE with position-aligned layout.}
To separate the effect of the layout from that of full sharing, we train UNITE with UNIFIED's position-aligned layout: its generative encoder maps image patches directly to latents without readout tokens, at $t_{\mathrm{enc}}=1$ since it no longer reads $\mathcal{N}(0,I)$ readouts, and it keeps UNITE's standalone decoder, so it differs from UNIFIED only in whether the decoder is shared. We train it at Base with the recipe of Table~\ref{tab:timestep} and report the checkpoints evaluated so far, up to epoch $80$ (Figure~\ref{fig:unite-aligned}). At epoch $80$, it reaches gFID $3.05$, behind UNITE with readout tokens ($2.46$) and both UNIFIED variants ($2.35$ with $t_{\mathrm{dec}}{=}{-1}$, $2.73$ with $t_{\mathrm{dec}}{=}1$), and it remains behind UNITE and UNIFIED ($t_{\mathrm{dec}}{=}{-1}$) at matched training FLOPs. Together with UNITE++ (Table~\ref{tab:timestep}), which shares the decoder but keeps readout tokens, these early results suggest that the gain of UNIFIED comes from combining position alignment with full sharing rather than from either alone.

\section{Text-to-image generation}
\label{app:t2i}

This section gives the training and evaluation protocol behind Table~\ref{tab:t2i-align} (Section~\ref{sec:t2i}) and of the $512\times512$ results in Table~\ref{tab:t2i-512}, and reports zero-shot FID on MS-COCO. Figures~\ref{fig:t2i-512-samples} and~\ref{fig:t2i-prompts} show additional samples at $512\times512$ and $256\times256$.

\subsection{Training}

\noindent\textbf{Data.} We train from scratch on JourneyDB~\citep{sun2023journeydb} (${\sim}4.3$M images) and the long-caption (${\sim}27$M) and short-caption (${\sim}4.8$M) subsets of BLIP3-o~\citep{chen2025blip3}, ${\sim}36$M image--text pairs in total, all at $256\times256$.

\noindent\textbf{Text conditioning.} Captions go through the chat template and the frozen Qwen3-0.6B~\citep{yang2025qwen3}. We use its final-layer hidden states, truncated to $256$ tokens and padded only to the longest caption in the batch, with padded positions masked out. The state of the last real token also serves as a pooled embedding for AdaLN. Text encoder is frozen during both pretraining and fine-tuning stage.

\noindent\textbf{Pretraining.} Training runs for $100$K iterations at batch size $1024$. The whole model is trained end to end with the L1, LPIPS loss and the flow-matching loss, using Muon and AdamW, an EMA decay of $0.9978$, and bf16. The peak learning rate is $2.67\times10^{-4}$: $10$K warmup steps, constant until $15$K, cosine decay to half the peak by $100$K, and constant afterwards. Captions are dropped with probability $0.1$ for classifier-free guidance.

\noindent\textbf{Supervised fine-tuning (SFT).} We start from the pretrained EMA weights and train on BLIP3o-60K ($58{,}859$ samples) for $2$K iterations at batch size $1024$, with a fresh optimizer and a $100$-step warmup to $2.67\times10^{-4}$, followed by cosine decay to $2.67\times10^{-5}$. Everything else is unchanged.

\subsection{Evaluation on GenEval, DPG-Bench, and GenAI-Bench}
\label{app:t2i-eval}

\noindent\textbf{Sampling.} All models are evaluated at $256\times256$ using EMA weights in fp32. We sample with a fixed-step Euler ODE solver for $50$ steps at classifier-free guidance scale $6.0$ over the whole trajectory, $t\in[0,1]$. Guidance is applied to the model's $x$-prediction after normalization. The unconditional branch has no text tokens and uses a learned null pooled embedding, and no other guidance is used. Prompts are used verbatim, without rewriting, LLM expansion, or negative prompts.

\noindent\textbf{Benchmarks.} We evaluate prompt alignment on GenEval~\citep{ghosh2023geneval} ($553$ prompts, $4$ images each), DPG-Bench~\citep{hu2024ella} ($1{,}065$ prompts, $4$ images each), and GenAI-Bench~\citep{li2024genai} ($1{,}600$ prompts, $1$ image each), and score them with the evaluation code and protocol of DiffusionBench~\citep{leng2026diffusionbench}. We report the overall GenEval accuracy (mean over its six categories), the mean DPG-Bench score, and the mean GenAI-Bench VQAScore.

\subsection{Zero-shot FID on MS-COCO}
\label{app:t2i-coco}

\noindent\textbf{Evaluation.} Following standard practice~\citep{ramesh2021zero,nichol2021glide,saharia2022photorealistic}, we evaluate zero-shot text-to-image generation on MS-COCO~\citep{lin2014microsoft}. We randomly sample $30{,}000$ images from the MS-COCO 2014 validation set with a fixed seed and use one randomly chosen caption per image. We generate one $256\times256$ image per caption and compute FID~\citep{heusel2017gans} against the $30{,}000$ sampled images, resized and center-cropped to $256\times256$. FID is computed with the Inception-V3~\citep{szegedy2016rethinking} (pool3 features) from torch-fidelity~\citep{obukhov2020torchfidelity}. We sample with an Euler ODE solver for $50$ steps and classifier-free guidance. MS-COCO images and captions are not used during training.

\noindent\textbf{Performance.} UNIFIED-B achieves FID-30K $19.34$ with $218$M parameters, and UNIFIED-L achieves FID-30K $18.74$ with $554$M parameters. Both models are trained on ${\sim}36$M image--text pairs for 100K iterations (Appendix~\ref{app:t2i}). For comparison with similar-sized models, FiTv2-XL/2~\citep{lu2024fit} ($759$M) achieves FID-30K $27.88$ on MS-COCO.

\noindent\textbf{Caption and domain shift.} We argue that zero-shot FID on MS-COCO favors training data whose captions and images resemble MS-COCO, and that our training data differs from it in two ways. First, caption length: $27$M of our $36$M training pairs come from the BLIP3-o long-caption subset, whose captions average $100$ tokens, while MS-COCO captions are shorter (average $17$ tokens). Second, the image domain: JourneyDB consists mainly of artistic, stylized images, whereas MS-COCO consists mainly of real photographs. Both shifts move our training distribution away from MS-COCO and likely raise our FID-30K.

\subsection{\texorpdfstring{Text-to-image generation at $512\times512$}{Text-to-image generation at 512x512}}
\label{app:t2i-512}

\noindent\textbf{Training.} Following DeCo~\citep{ma2026deco} and PixelGen~\citep{ma2026pixelgen}, we train UNIFIED-B in three phases. (i)~\emph{Pretraining at $256\times256$} for $200$K iterations, with the data and recipe of Appendix~\ref{app:t2i}. (ii)~\emph{Fine-tuning at $512\times512$} on the same data, starting from the EMA weights of phase~(i) with a fresh optimizer, at batch size $512$ with a $2$K-step linear warmup to a constant learning rate of $6.67\times10^{-5}$. DeCo and PixelGen run this phase for $80$K iterations; we report the checkpoint after $10$K iterations. (iii)~\emph{SFT on BLIP3o-60k at $512\times512$} for $2$K iterations at batch size $512$, with a $100$-step warmup to $1.33\times10^{-4}$ and cosine decay to $1.33\times10^{-5}$. The losses, optimizer, EMA decay, text encoder, and caption dropout are unchanged.

\noindent\textbf{From $256\times256$ to $512\times512$.} The patch size is unchanged, so an image yields $1{,}024$ latent tokens instead of $256$, a $4\times$ longer sequence. We interpolate the position embeddings of the $256\times256$ checkpoint to the larger grid and, as is common at higher resolution~\citep{esser2024scaling}, shift the noise schedule further toward high noise, from $\alpha=0.5$ to $\alpha=0.25$ (Appendix~\ref{app:eval-protocols}). The evaluation protocol follows Appendix~\ref{app:t2i-eval}, with images generated at $512\times512$.

\noindent\textbf{Results.} Table~\ref{tab:t2i-512} and its discussion are in Section~\ref{sec:t2i}. Figure~\ref{fig:t2i-512-samples} shows further $512\times512$ samples from UNIFIED-B, complementing those in Figure~\ref{fig:t2i-samples}.

\newcommand{\tfcell}[2]{%
  \begin{minipage}[b]{0.243\linewidth}\centering
    {\scriptsize\parbox[b][3\baselineskip][b]{\linewidth}{\centering\textit{#2}}}\\[2pt]
    #1%
  \end{minipage}}

\begin{figure}[t]
\centering
\tfcell{\includegraphics[width=\linewidth]{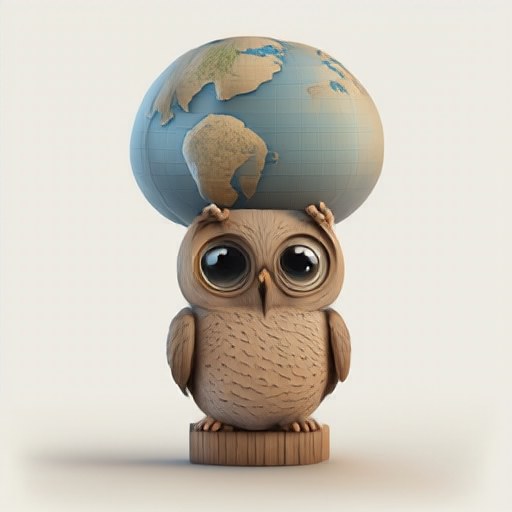}}{a cute wooden owl statue holding a large globe of the Earth above its head}\hfill
\tfcell{\includegraphics[width=\linewidth]{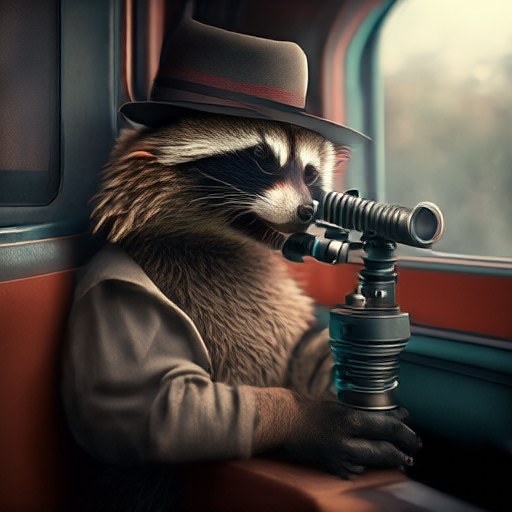}}{a racoon detective using a microscope while riding in a train}\hfill
\tfcell{\includegraphics[width=\linewidth]{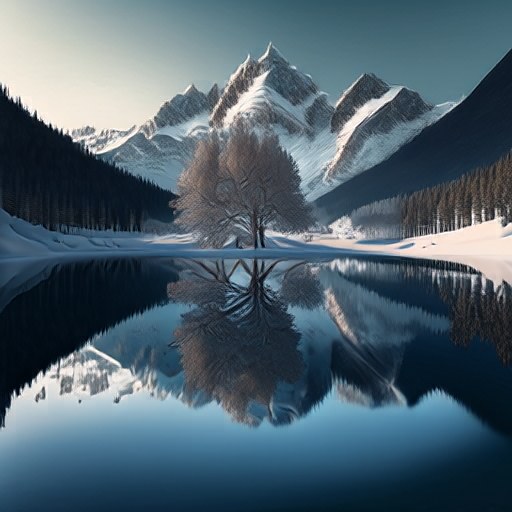}}{Snow mountain and tree reflection in the lake}\hfill
\tfcell{\includegraphics[width=\linewidth]{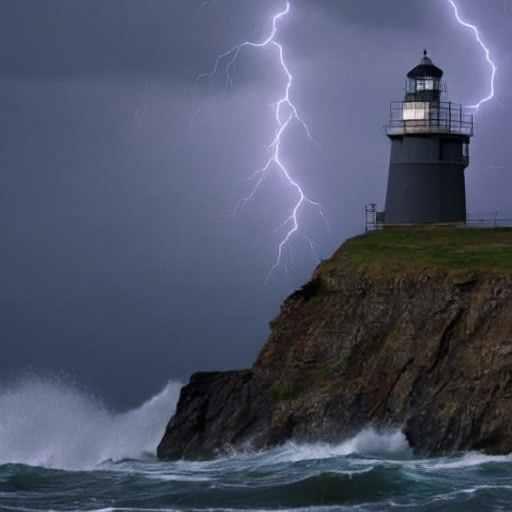}}{A lighthouse on a cliff during a thunderstorm, waves crashing, lightning in the sky.}\\[4pt]
\tfcell{\includegraphics[width=\linewidth]{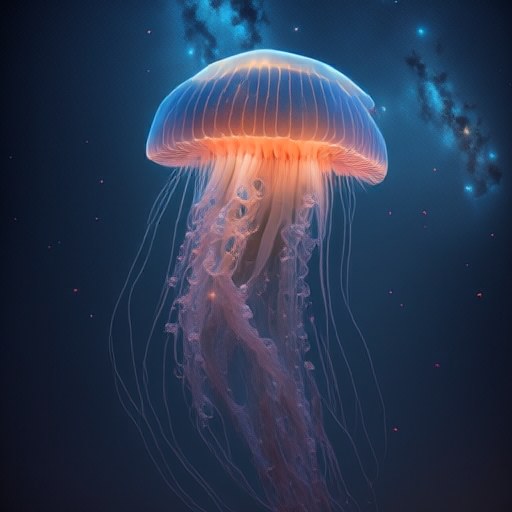}}{A jellyfish floating through a starry night sky.}\hfill
\tfcell{\includegraphics[width=\linewidth]{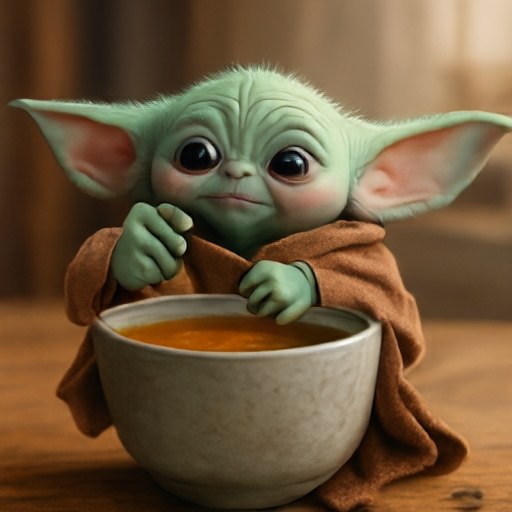}}{Baby Yoda (Grogu) drinking soup.}\hfill
\tfcell{\includegraphics[width=\linewidth]{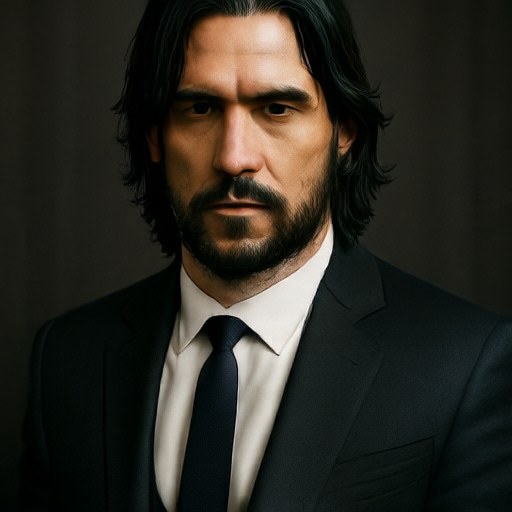}}{Keanu Reeves as John Wick.}\hfill
\tfcell{\includegraphics[width=\linewidth]{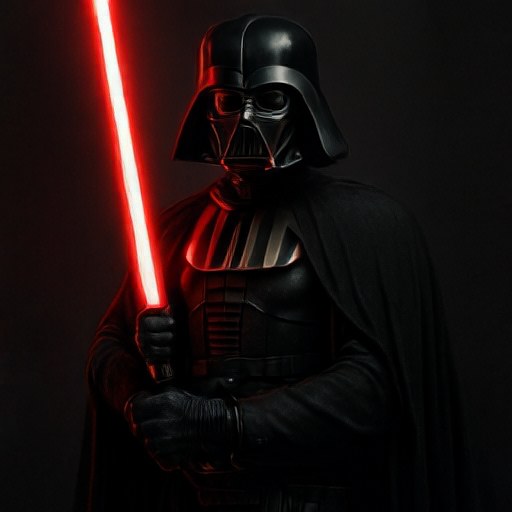}}{Darth Vader holding a red lightsaber.}
\caption{\textbf{Text-to-image samples from UNIFIED-B} ($218$M) after $10$K fine-tuning steps at $512\times512$, with a $50$-step Euler solver and CFG scale $6.0$. Prompts are shown above each image.}
\label{fig:t2i-512-samples}
\end{figure}

\section{Molecular experiments on QM9}
\label{sec:qm9}

\begin{table}[t]
\centering
\caption{\textbf{Molecule reconstruction and generation on QM9.} All models have
${\sim}33$M parameters: ADiT uses a separate encoder, decoder, and generator
($5.5$M/$10.7$M/$16.5$M), UNITE a generative encoder ($22$M) and a decoder
($10.7$M), and UNIFIED one shared backbone ($32.7$M). Reconstruction is measured on
the $12{,}831$ test molecules and generation on $10{,}000$ samples.}
\label{tab:qm9}
\small
\begin{tabular}{lcccc}
\toprule
& \multicolumn{2}{c}{Reconstruction} & \multicolumn{2}{c}{Generation} \\
\cmidrule(lr){2-3} \cmidrule(lr){4-5}
Method & Match (\%) $\uparrow$ & RMSD (\AA) $\downarrow$ & Valid (\%) $\uparrow$ & Unique (\%) $\uparrow$ \\
\midrule
ADiT    & \textbf{99.38} & \textbf{0.025} & 94.04 & 94.86 \\
UNITE   & 98.42 & 0.035 & 95.79 & \textbf{98.68} \\
UNIFIED ($t_{\mathrm{dec}}{=}{-1}$) & 97.30 & 0.058 & \textbf{96.11} & 98.49 \\
\bottomrule
\end{tabular}
\end{table}

\noindent\textbf{Overview.} To test whether UNIFIED works beyond images, we apply it to 3D molecule
generation on QM9~\citep{ramakrishnan2014quantum}, which contains ${\sim}130$K
small molecules with up to $29$ atoms. Unlike images, molecules have no fixed
token grid: each molecule is a variable-size set of atoms, given by atom types
and 3D coordinates. Each atom is one token with a $16$-dimensional latent.
UNITE uses an $8$-block generative encoder and a $6$-block decoder, and UNIFIED a
single $12$-block backbone with the same position-aligned layout as on ImageNet
(atom $i$ and latent $i$ share a position). Table~\ref{tab:qm9} compares them at
the same parameter count, with the two-stage ADiT~\citep{joshi2025all} as a
reference.

\noindent\textbf{Implementation details.}\label{app:qm9} We follow the QM9 setup of ADiT~\citep{joshi2025all}. UNITE and UNIFIED share the data, losses,
hyperparameters (Table~\ref{tab:qm9-hparams}), sampler, and evaluation, and
differ only in the backbone. We use the random $100{,}000$/$18{,}000$/$12{,}831$
train/validation/test split of ADiT. 

\noindent\textbf{Evaluation.}\label{app:qm9-details} For evaluation, we generate $10{,}000$ molecules with $500$ Euler steps and no guidance.
Following ADiT, a reconstruction counts as a match if it aligns with the input
molecule under pymatgen~\citep{ong2013python}, with RMSD averaged over matches;
a generated molecule is valid if RDKit~\citep{bento2020open} can parse it; and
uniqueness is the fraction of distinct valid molecules.

\noindent\textbf{Results.} UNIFIED generates at least as well as UNITE
(validity $96.11\%$ vs.\ $95.79\%$, uniqueness $98.49\%$ vs.\ $98.68\%$) but
reconstructs less accurately (match rate $97.30\%$ vs.\ $98.42\%$, RMSD $0.058$ vs.\ $0.035$\,\AA{}).
Most failed reconstructions differ from the input only in where a
single hydrogen atom sits, and the reconstructed molecules are as chemically
valid as the originals. 

\begin{table}[!ht]
\centering
\caption{QM9 configuration, shared by UNITE and UNIFIED.}
\label{tab:qm9-hparams}
\small
\begin{tabular}{@{}lp{0.62\linewidth}@{}}
\toprule
Setting & Value \\
\midrule
Backbone & width $384$, $6$ heads, AdaLN blocks \\
Blocks & UNITE: $8$ (generative encoder) $+\,6$ (decoder); UNIFIED: $12$ \\
Parameters & $32.7$M each \\
Latent & $16$ channels per atom \\
Batch size / epochs & $512$ / $8{,}000$ \\
Reconstruction noise $\sigma_{\mathrm{rec}}$ & $0.7$ \\
Generation : reconstruction updates & $1{:}1$ \\
As in ADiT & AdamW (constant LR $10^{-4}$), EMA $0.9999$, coordinate-loss weight $10$, label dropout $0.1$, encoder-input corruption, Euler sampler ($500$ steps) \\
\bottomrule
\end{tabular}
\end{table}

\section{Limitations}
\label{sec:limitations}
Our evidence that full unification helps comes from images, where every sample
shares one token grid that position alignment can exploit. On QM9
(Appendix~\ref{sec:qm9}), where samples are unordered sets without such a grid,
UNIFIED generates as well as UNITE but reconstructs slightly worse. Whether this
gap stems from limited decoding capacity at small scale or from position
alignment itself, and how to adapt position alignment to such data, are left to
future work. Our comparisons with UNITE are at $256\times256$. UNIFIED
also works at $512\times512$ (Section~\ref{sec:t2i}), but whether its gains
over a separate decoder persist at higher resolutions and longer token sequences
is left to future work.

Full weight sharing also rules out freezing the latent space. With a separate
tokenizer, the latent space can be fixed, from the start or after joint
training, and the generator trained alone on precomputed latents. In UNIFIED,
every update to the shared backbone also changes the encoding and decoding
passes, so the reconstruction objective must stay active throughout training,
and each step pays for encoding and decoding passes alongside generation.

\section{Uncurated ImageNet samples}
\label{app:uncurated}

Figures~\ref{fig:uncurated-1} and~\ref{fig:uncurated-2} show uncurated class-conditional samples from UNIFIED ($t_{\mathrm{dec}}{=}{-1}$), trained on ImageNet-1K $256\times256$ and sampled with CFG scale $4.0$ and the adaptive dopri5 solver. For each of $24$ classes, we show the first $10$ samples without selection.

\newcommand{\uncls}[2]{%
  \begin{minipage}[b]{0.485\linewidth}\raggedright
    {\small #2}\\[2pt]
    #1%
  \end{minipage}}

\begin{figure}[p]
\centering
\parbox[c][\dimexpr\textheight-4\baselineskip\relax][s]{\linewidth}{%
\uncls{\includegraphics[width=\linewidth]{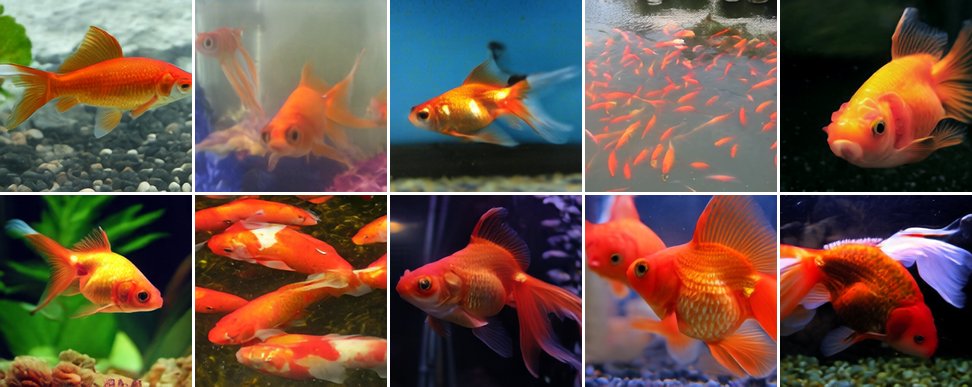}}{goldfish (1)}\hfill
\uncls{\includegraphics[width=\linewidth]{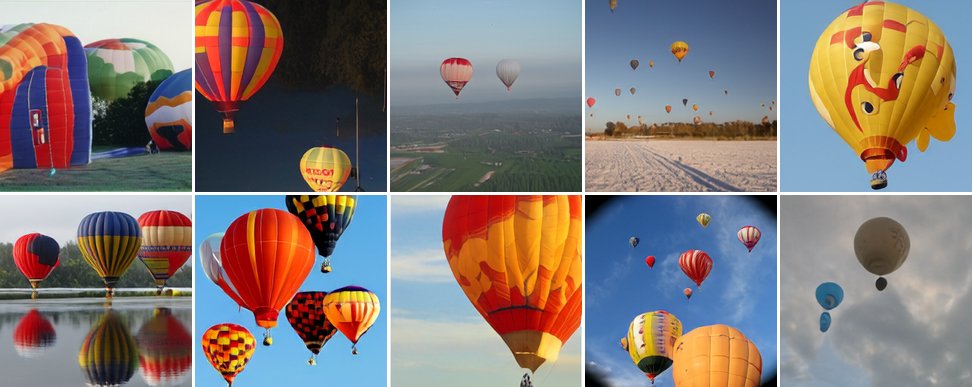}}{balloon (417)}
\vfill
\uncls{\includegraphics[width=\linewidth]{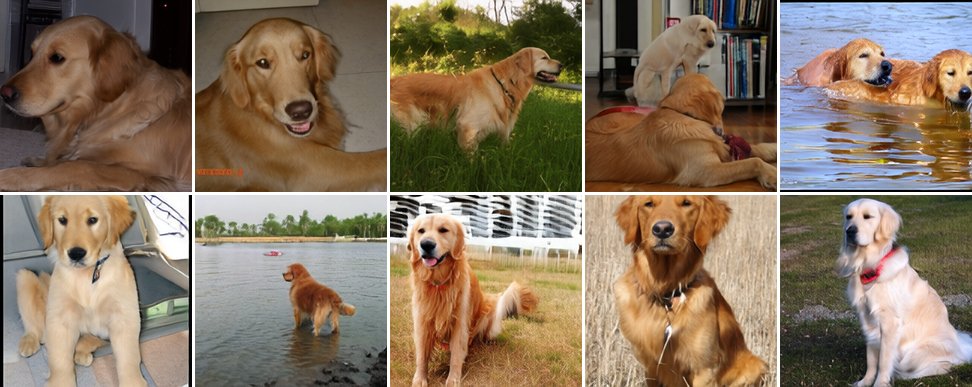}}{golden retriever (207)}\hfill
\uncls{\includegraphics[width=\linewidth]{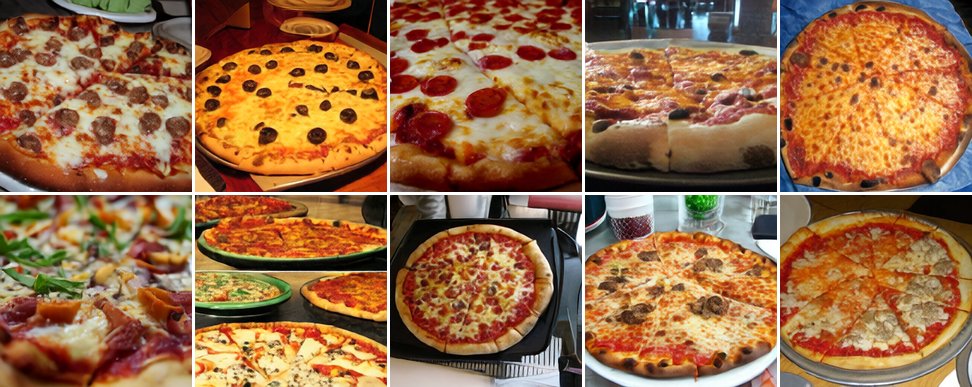}}{pizza (963)}
\vfill
\uncls{\includegraphics[width=\linewidth]{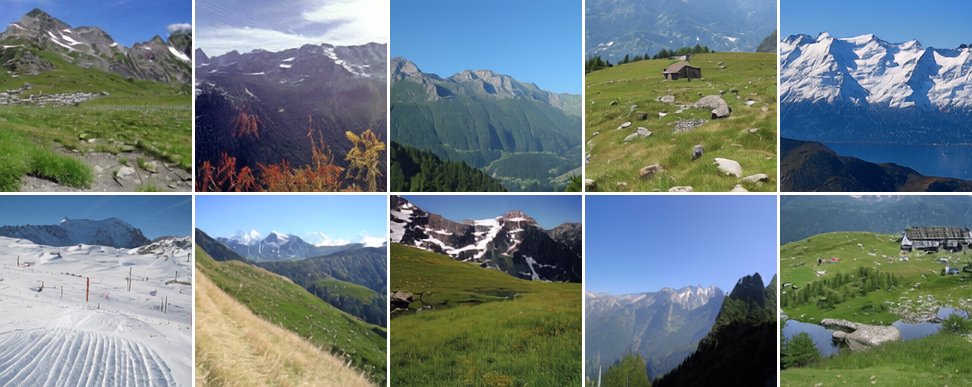}}{alp (970)}\hfill
\uncls{\includegraphics[width=\linewidth]{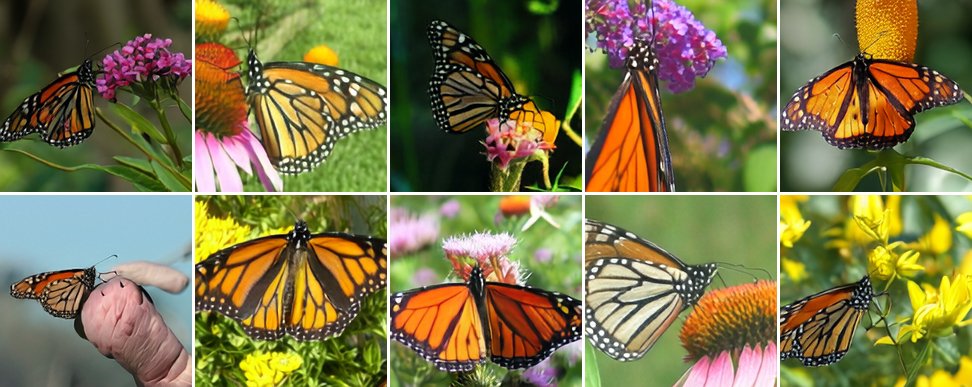}}{monarch (323)}
\vfill
\uncls{\includegraphics[width=\linewidth]{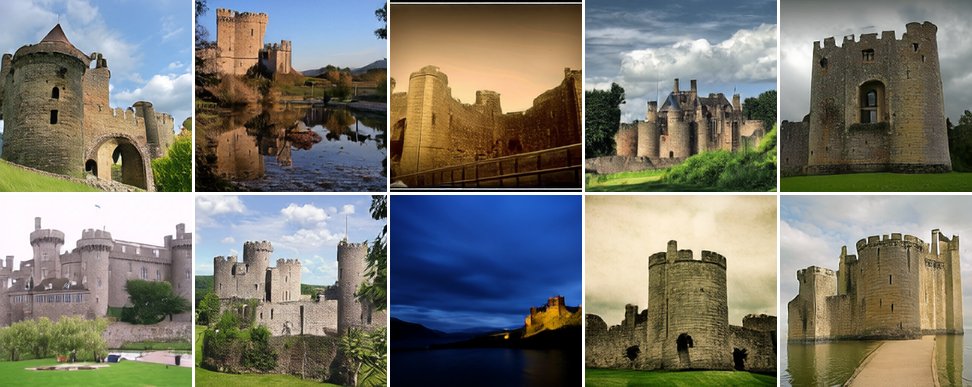}}{castle (483)}\hfill
\uncls{\includegraphics[width=\linewidth]{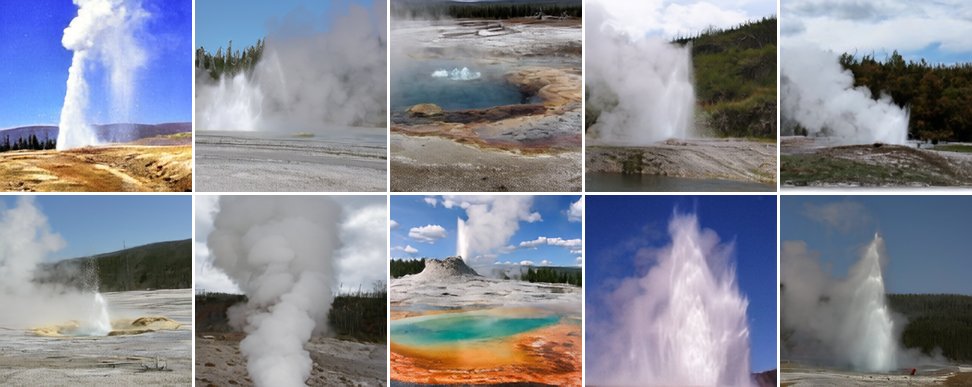}}{geyser (974)}
\vfill
\uncls{\includegraphics[width=\linewidth]{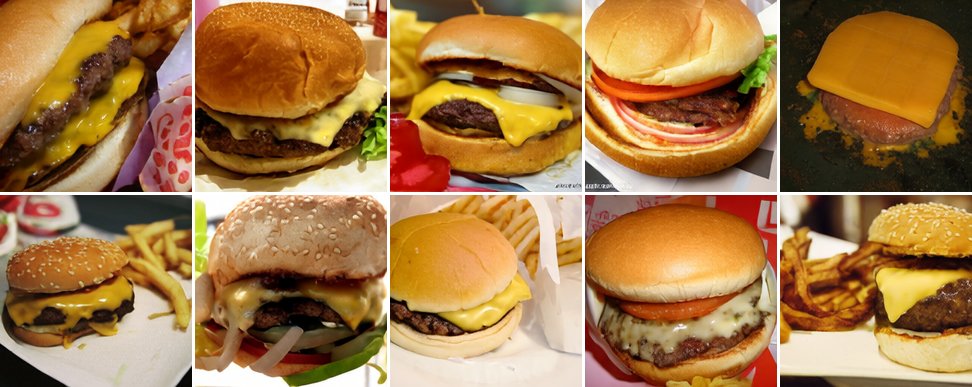}}{cheeseburger (933)}\hfill
\uncls{\includegraphics[width=\linewidth]{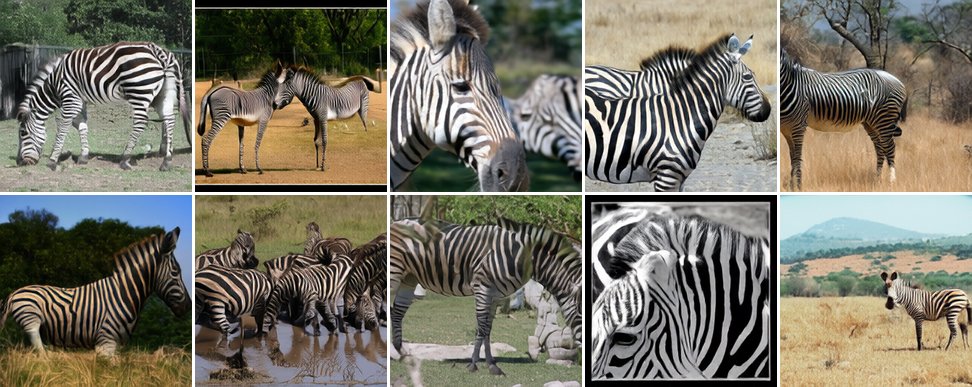}}{zebra (340)}
\vfill
\uncls{\includegraphics[width=\linewidth]{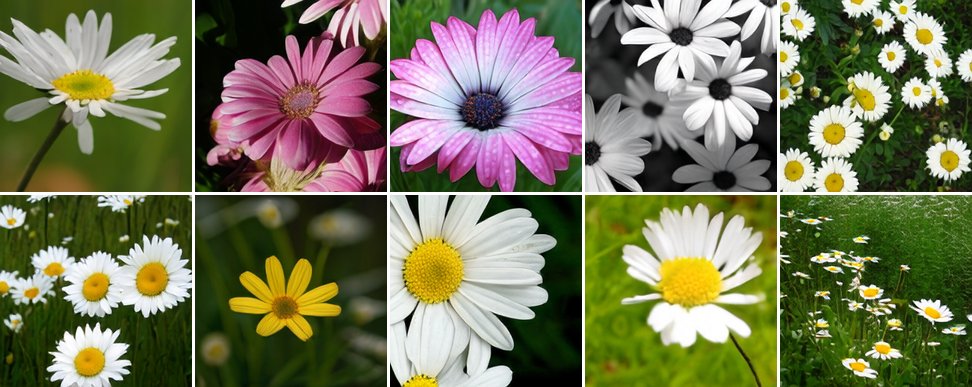}}{daisy (985)}\hfill
\uncls{\includegraphics[width=\linewidth]{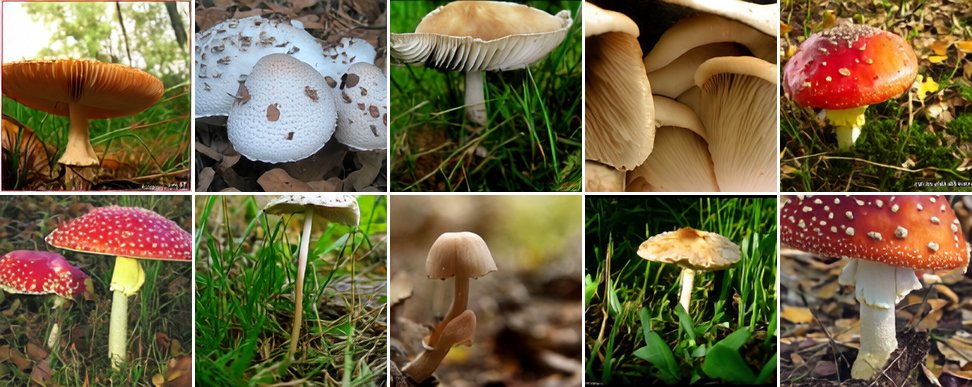}}{mushroom (947)}
}
\caption{\textbf{Uncurated samples} from UNIFIED on ImageNet-1K $256\times256$ (CFG scale $4.0$): the first $10$ samples of each class; class indices in parentheses.}
\label{fig:uncurated-1}
\end{figure}

\begin{figure}[p]
\centering
\parbox[c][\dimexpr\textheight-4\baselineskip\relax][s]{\linewidth}{%
\uncls{\includegraphics[width=\linewidth]{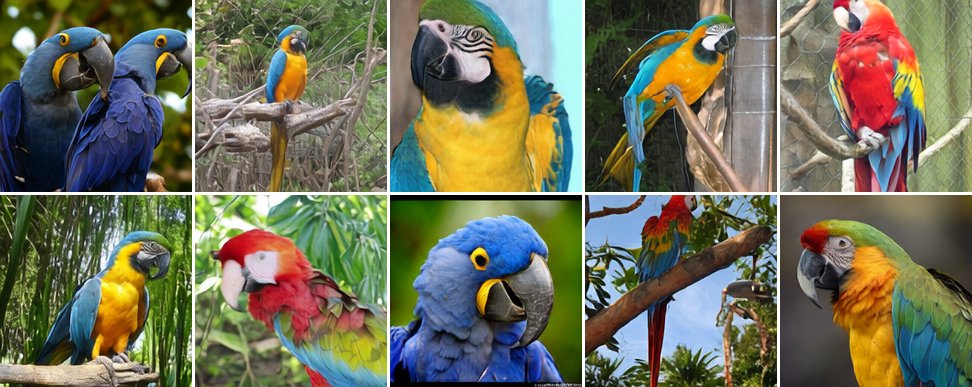}}{macaw (88)}\hfill
\uncls{\includegraphics[width=\linewidth]{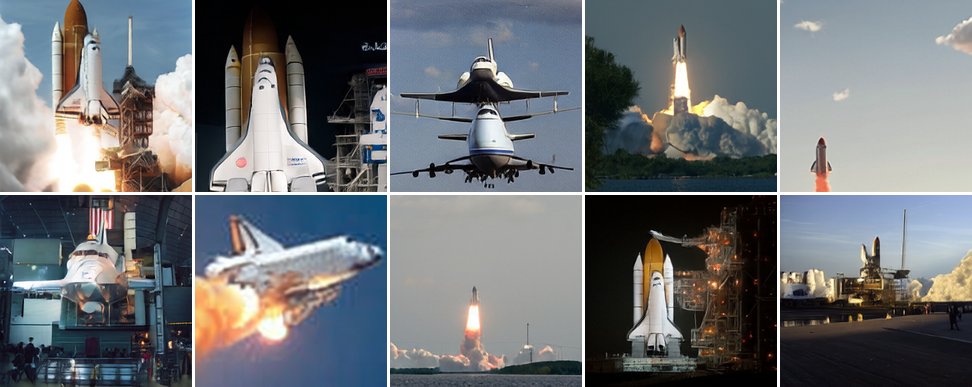}}{space shuttle (812)}
\vfill
\uncls{\includegraphics[width=\linewidth]{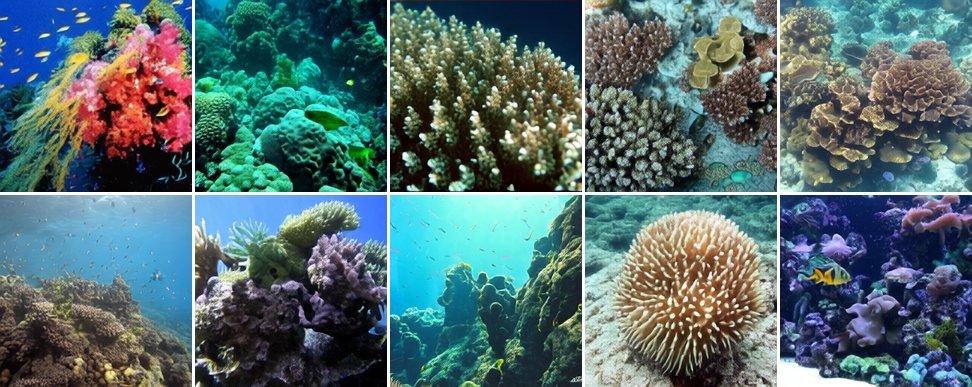}}{coral reef (973)}\hfill
\uncls{\includegraphics[width=\linewidth]{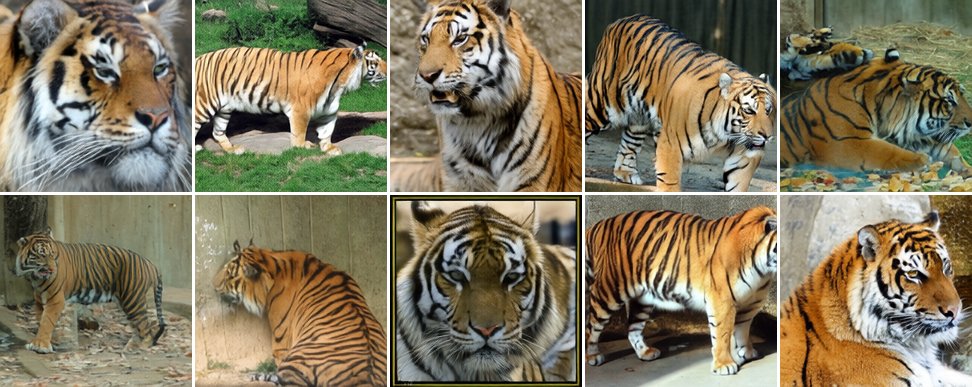}}{tiger (292)}
\vfill
\uncls{\includegraphics[width=\linewidth]{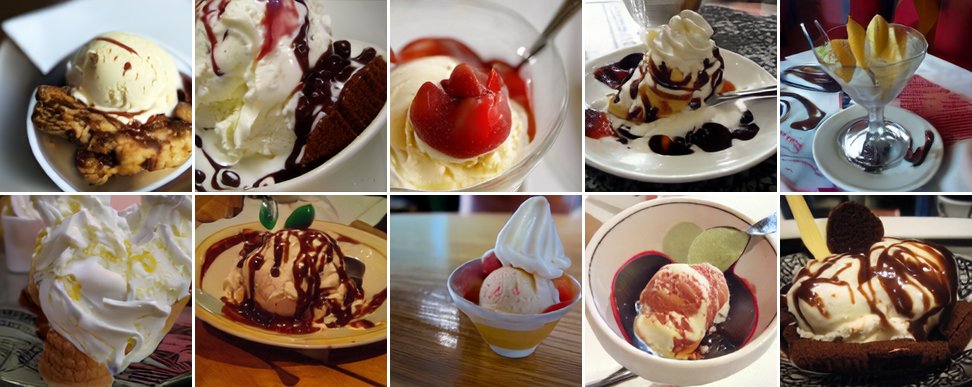}}{ice cream (928)}\hfill
\uncls{\includegraphics[width=\linewidth]{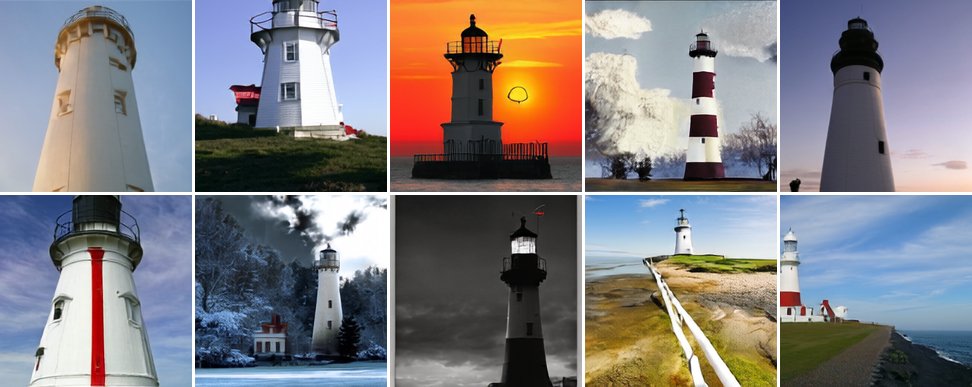}}{beacon (437)}
\vfill
\uncls{\includegraphics[width=\linewidth]{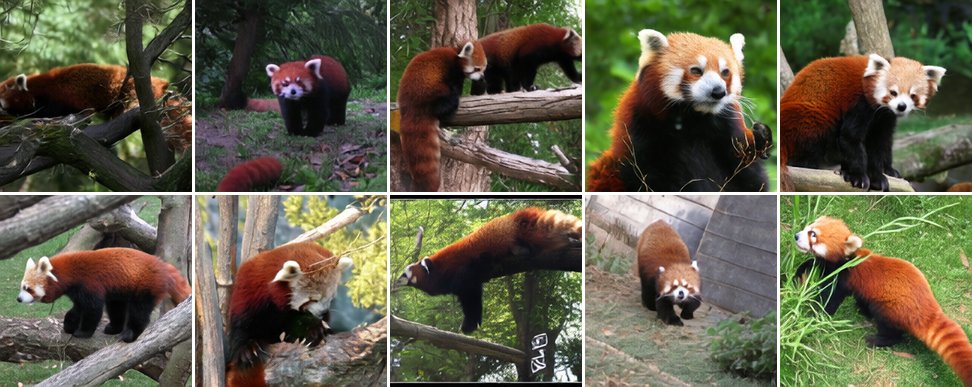}}{lesser panda (387)}\hfill
\uncls{\includegraphics[width=\linewidth]{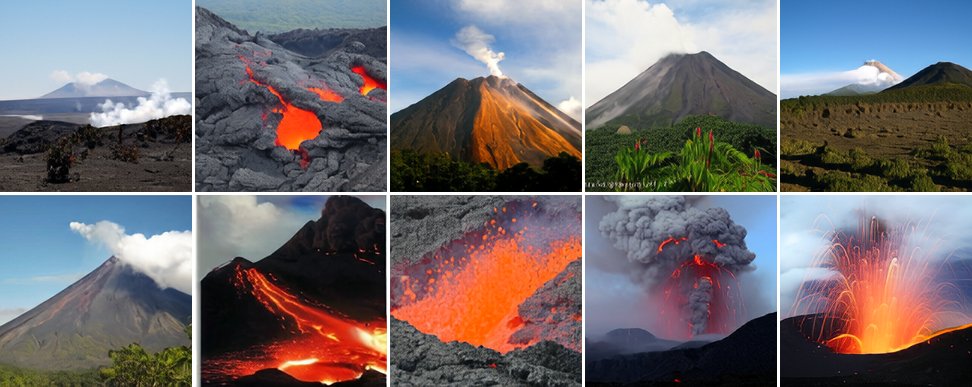}}{volcano (980)}
\vfill
\uncls{\includegraphics[width=\linewidth]{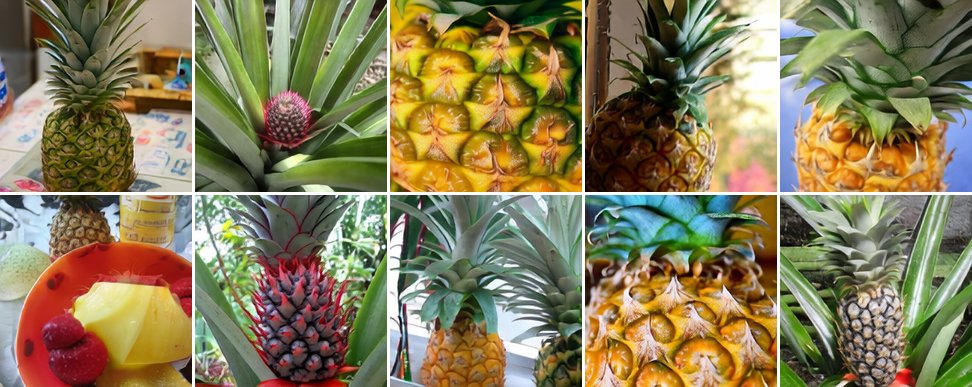}}{pineapple (953)}\hfill
\uncls{\includegraphics[width=\linewidth]{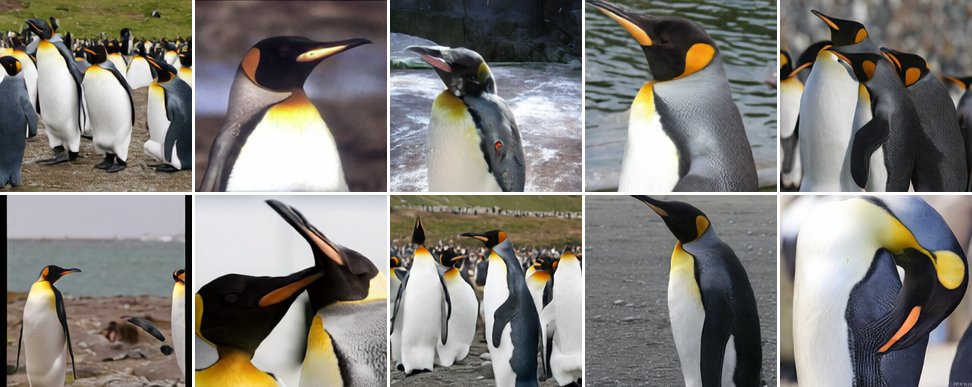}}{king penguin (145)}
\vfill
\uncls{\includegraphics[width=\linewidth]{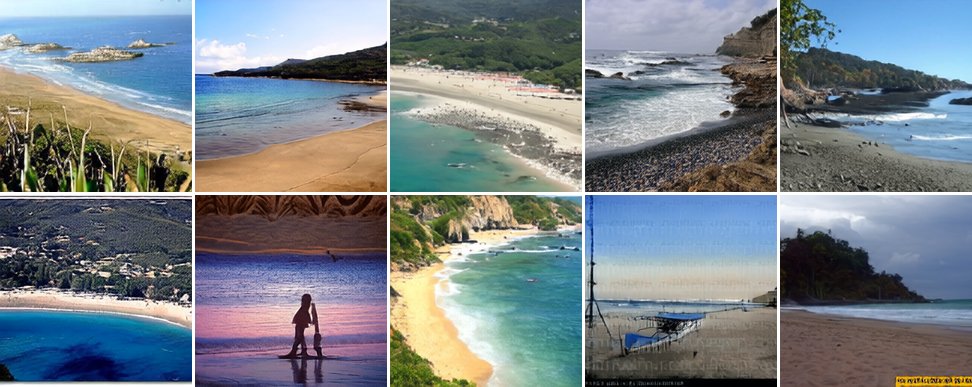}}{seashore (978)}\hfill
\uncls{\includegraphics[width=\linewidth]{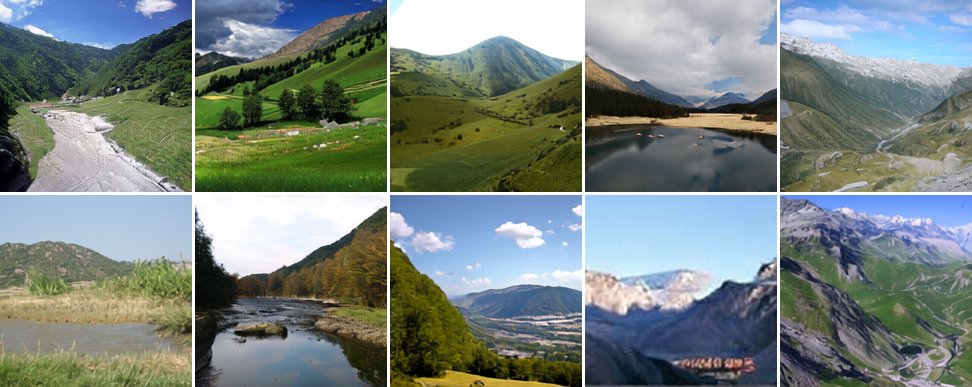}}{valley (979)}
}
\caption{\textbf{Uncurated samples} from UNIFIED on ImageNet-1K $256\times256$ (continued from Figure~\ref{fig:uncurated-1}).}
\label{fig:uncurated-2}
\end{figure}

\newcommand{\tpcell}[2]{%
  \begin{minipage}[t]{0.175\linewidth}\vspace{0pt}%
    #1\\[1pt]
    {\fontsize{5}{5.6}\selectfont\raggedright #2\par}%
  \end{minipage}}

\begin{figure}[p]
\centering
\tpcell{\includegraphics[width=\linewidth]{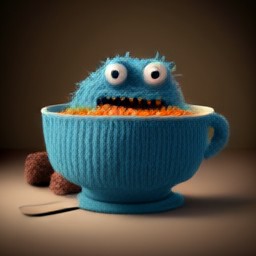}}{A bowl of soup that looks like a monster knitted out of wool}\hfill
\tpcell{\includegraphics[width=\linewidth]{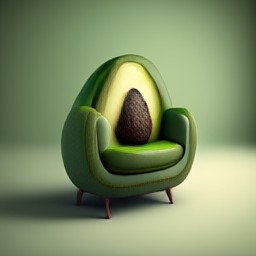}}{an armchair in the shape of an avocado}\hfill
\tpcell{\includegraphics[width=\linewidth]{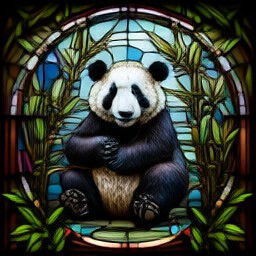}}{a stained glass window of a panda eating bamboo}\hfill
\tpcell{\includegraphics[width=\linewidth]{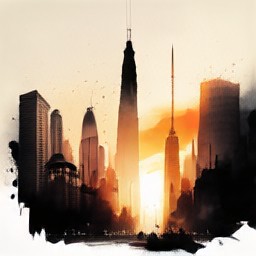}}{Downtown Shanghai at sunrise. detailed ink wash.}\hfill
\tpcell{\includegraphics[width=\linewidth]{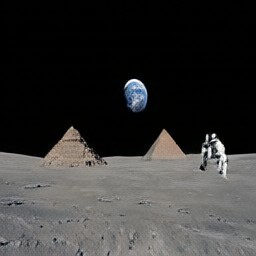}}{Ground view of the Great Pyramids and Sphinx on the moon's surface. The back of an astronaut is in the foreground. The planet Earth looms in the sky.}\par\vspace{3pt}
\tpcell{\includegraphics[width=\linewidth]{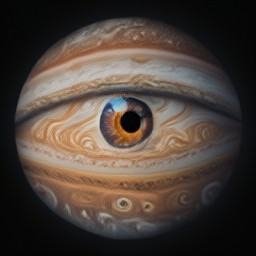}}{the eye of the planet Jupiter}\hfill
\tpcell{\includegraphics[width=\linewidth]{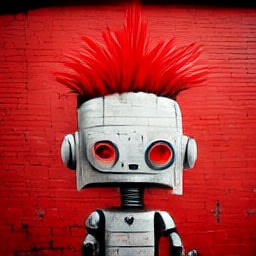}}{a white robot with a red mohawk painted as graffiti on a red brick wall}\hfill
\tpcell{\includegraphics[width=\linewidth]{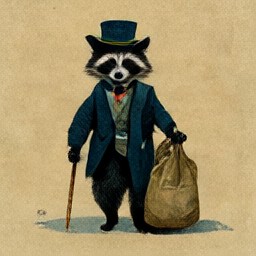}}{A raccoon wearing formal clothes, wearing a tophat and holding a cane. The raccoon is holding a garbage bag. Oil painting in the style of Hokusai.}\hfill
\tpcell{\includegraphics[width=\linewidth]{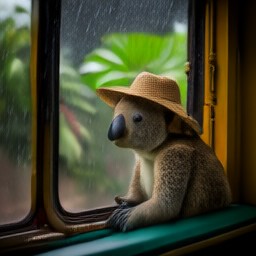}}{A train ride in the monsoon rain in Kerala. With a Koala bear wearing a hat looking out of the window. There is a lot of coconut trees out of the window.}\hfill
\tpcell{\includegraphics[width=\linewidth]{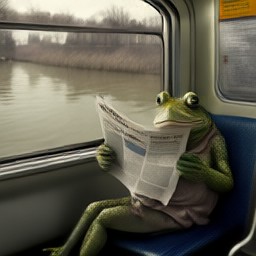}}{A photograph of the inside of a subway train. There are frogs sitting on the seats. One of them is reading a newspaper. The window shows the river in the background.}\par\vspace{3pt}
\tpcell{\includegraphics[width=\linewidth]{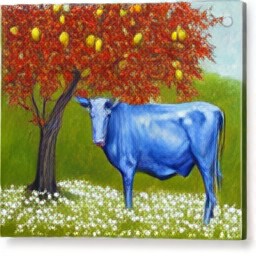}}{a blue cow is standing next to a tree with red leaves and yellow fruit. the cow is standing in a field with white flowers. impressionistic painting}\hfill
\tpcell{\includegraphics[width=\linewidth]{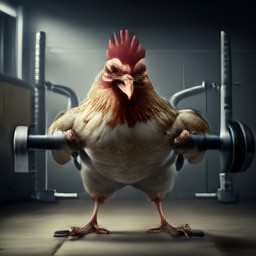}}{A high resolution photo of a chicken working out in a gym.}\hfill
\tpcell{\includegraphics[width=\linewidth]{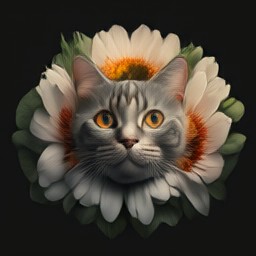}}{a flower with a cat's face in the middle}\hfill
\tpcell{\includegraphics[width=\linewidth]{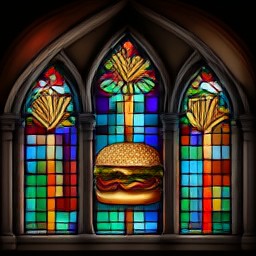}}{A church with stained glass windows depicting a hamburger and french fries.}\hfill
\tpcell{\includegraphics[width=\linewidth]{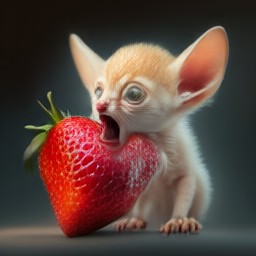}}{A baby fennec sneezing onto a strawberry, detailed, macro, studio light, droplets, backlit ears.}\par\vspace{3pt}
\tpcell{\includegraphics[width=\linewidth]{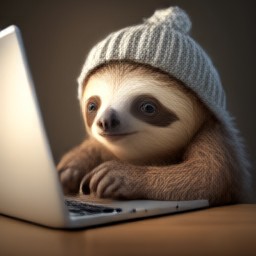}}{A fluffy baby sloth with a knitted hat trying to figure out a laptop, close up, highly detailed, studio lighting, screen reflecting in its eyes.}\hfill
\tpcell{\includegraphics[width=\linewidth]{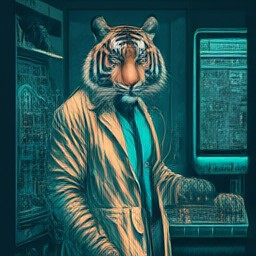}}{A tiger in a lab coat with a 1980s Miami vibe, turning a well oiled science content machine, digital art.}\hfill
\tpcell{\includegraphics[width=\linewidth]{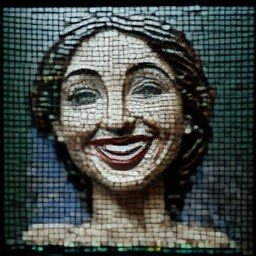}}{A ceramic glass mosaic of a smiling woman's face.}\hfill
\tpcell{\includegraphics[width=\linewidth]{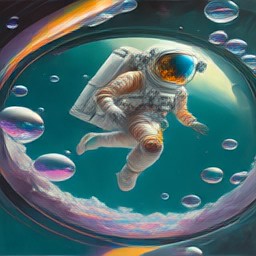}}{A highly-detailed painting of an astronaut in a jumping float pose inside a futuristic space station with an iridescent bubble skin.}\hfill
\tpcell{\includegraphics[width=\linewidth]{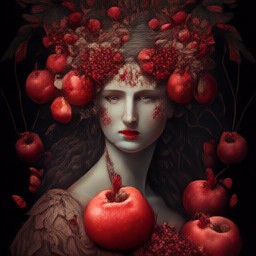}}{Persephone with pomegranates.}\par\vspace{3pt}
\tpcell{\includegraphics[width=\linewidth]{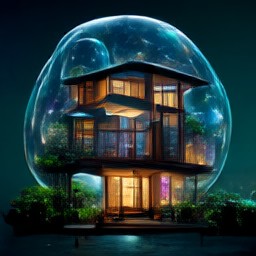}}{a home built in a huge soap bubble, windows, doors, porches, awnings, in the middle of space, cyberpunk lights.}\hfill
\tpcell{\includegraphics[width=\linewidth]{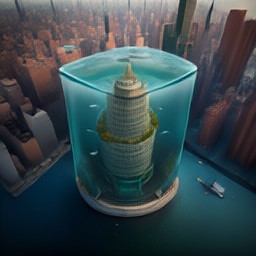}}{aerial view of a giant fish tank shaped like a tower in the middle of new york city, photorealistic.}\hfill
\tpcell{\includegraphics[width=\linewidth]{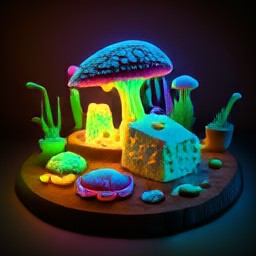}}{charcuterie board with multicolored alien cheeses, with glowing mold and fungus.}\hfill
\tpcell{\includegraphics[width=\linewidth]{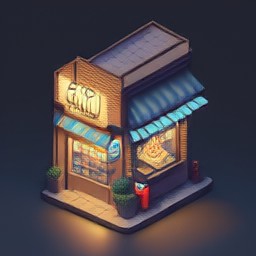}}{Isometric pixel art of a tiny ramen shop at night.}\hfill
\tpcell{\includegraphics[width=\linewidth]{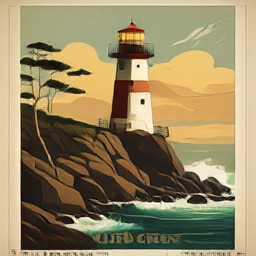}}{A 1920s travel poster of a lighthouse on a rocky coast.}\par\vspace{3pt}
\tpcell{\includegraphics[width=\linewidth]{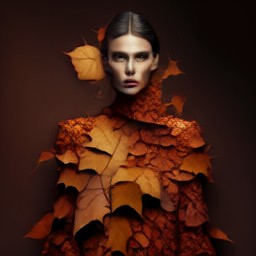}}{A studio fashion editorial photo of a model wearing a dress made of autumn leaves.}\hfill
\tpcell{\includegraphics[width=\linewidth]{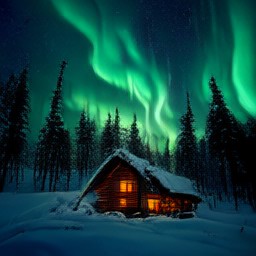}}{A snow-covered cabin in a pine forest under the northern lights.}\hfill
\tpcell{\includegraphics[width=\linewidth]{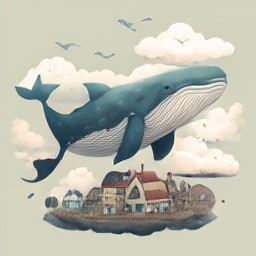}}{A whale swimming through clouds above a small village.}\hfill
\tpcell{\includegraphics[width=\linewidth]{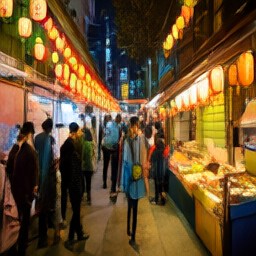}}{A bustling night market in Taipei with lanterns and food stalls.}\hfill
\tpcell{\includegraphics[width=\linewidth]{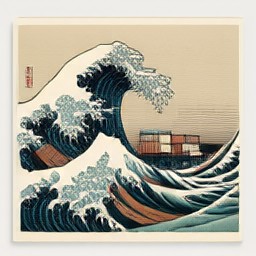}}{A ukiyo-e woodblock print of a great wave crashing over a modern container ship.}\par
\caption{\textbf{Text-to-image samples from UNIFIED-L} ($554$M, $100$K pretraining steps) at $256\times256$, with a $50$-step Euler solver and CFG scale $6.0$. Prompts are from PartiPrompts~\citep{yu2022scaling}, DrawBench~\citep{saharia2022photorealistic}, and other sources; we show the first sample for each.}
\label{fig:t2i-prompts}
\end{figure}

\end{document}